\documentclass[twoside,11pt]{article}

\usepackage[utf8]{inputenc} %
\usepackage[T1]{fontenc}    %
\usepackage{url}            %
\usepackage{booktabs}       %
\usepackage{amsfonts}       %
\usepackage{nicefrac}       %
\usepackage{microtype}      %
\usepackage{caption}
\usepackage{subcaption}
\usepackage{float}
\usepackage{mathtools}
\usepackage{multirow}
\usepackage{dsfont}
\usepackage{etoc}

\usepackage[ruled, vlined]{algorithm2e}
\usepackage[colorlinks=false,allbordercolors={1 1 1}]{hyperref}
\usepackage{amsthm}
\usepackage[nameinlink,capitalise,noabbrev]{cleveref}
\creflabelformat{equation}{#2\textup{#1}#3}

\usepackage[preprint,nohyperref]{jmlr2e}

\usepackage{amsmath}
\usepackage{mathtools}
\usepackage{bm}
\usepackage[shortlabels]{enumitem}

\usepackage[acronym]{glossaries}
\glsdisablehyper
\newacronym{gp}{GP}{Gaussian process}
\newacronym{mmd}{MMD}{maximum mean discrepancy}
\newacronym{ksd}{KSD}{kernel Stein discrepancy}
\newacronym{ipm}{IPM}{integral pseudo-probability metric}
\newacronym{rkhs}{RKHS}{reproducing kernel Hilbert space}
\newacronym{ipd}{IPD}{integral positive definite}
\newacronym{cdf}{CDF}{cumulative distribution function}
\newacronym{mle}{MLE}{maximum likelihood estimator}

\DeclareMathOperator{\MMD}{MMD}

\DeclareMathOperator{\KSD}{KSD}

\newcommand{\norm}[1]{\left\lVert#1\right\rVert}

\DeclareMathOperator*{\argmin}{arg\,min}

\DeclareMathOperator{\Normal}{\mathcal{N}}

\newcommand{\given}[1][]{\:#1\vert\:}
\DeclarePairedDelimiterX{\dvg}[2]{(}{)}{#1\;\delimsize\|\;#2}

\newcommand{\eqstop}[0]{\,.}
\newcommand{\eqcom}[0]{\,,}

\newcommand{\Reals}[0]{\mathbb{R}}

\let\P\relax
\let\U\relax
\newcommand{\P}{P}
\newcommand{\Q}{Q}
\DeclareMathOperator{\R}{\mathbb{R}}

\renewcommand{\H}{\mathcal{H}}
\newcommand{\X}{\mathcal{X}}
\DeclareMathOperator{\PX}{\mathcal{P}(\mathcal{X})}

\newcommand{\F}{\mathcal{F}}

\newcommand{\half}[0]{\frac{1}{2}}

\newcommand{\defeq}{\vcentcolon=}

\usepackage{xcolor}

\usepackage{tikz}
\usetikzlibrary{fadings}
\usetikzlibrary{patterns}
\usetikzlibrary{shadows.blur}
\usetikzlibrary{shapes}
\definecolor{blue}{HTML}{1f77b4}
\definecolor{orange}{HTML}{ff7f0e}
\definecolor{green}{HTML}{2ca02c}
\definecolor{red}{HTML}{d62728}
\definecolor{darkred}{HTML}{d62728}

\makeatletter
\newcommand{\removelatexerror}{\let\@latex@error\@gobble}
\makeatother

\DeclareMathOperator{\D}{D}

\newcommand{\thetah}{\hat{\theta}}
\newcommand{\thetahn}{\hat{\theta}_n}
\newcommand{\thetaz}{\theta_0}
\newcommand{\Prob}{\mathbb{P}}

\newcommand{\Sl}{\mathcal{S}^{l}}
\newcommand{\E}{\mathbb{E}}
\newcommand{\U}{\mathbb{U}}
\newcommand{\calU}{\mathcal{U}}
\newcommand{\nullnc}{H_0}
\newcommand{\altnc}{H_1}
\newcommand{\nullc}{H^{C}_0}
\newcommand{\altc}{H^{C}_1}

\newcommand{\tx}{\tilde{x}}

\newcommand{\lmed}{l_{\text{med}}}
\newcommand{\lkef}{l_{\text{kef}}}
\newcommand{\qkef}{q_{\text{kef}}}

\newcommand{\toprob}{\overset{\mathbb{P}}{\to}}

\newcommand{\todist}{\overset{\mathcal{D}}{\to}}

\newcommand{\iid}{\overset{iid}{\sim}}

\newcommand{\Hd}{\H_K^d}
\newcommand{\N}{\mathbb{N}}

\DeclareMathOperator{\TruncNormal}{\widetilde{\Normal}}

\newcommand{\Pts}[1][]{\P^{\textrm{ts}}_{#1}}

\newcommand{\Sb}{\boldsymbol{S}}
\newcommand{\Var}{\operatorname{\mathbb{V}ar}}

\newcommand{\g}[1]{g(\omega,#1)}
\newcommand{\Cov}{\operatorname{\mathbb{C}ov}}
\newcommand{\MMDk}{\MMD}
\newcommand{\dist}{\mathcal{D}}

\newcommand{\LH}[1]{\left\|#1\right\|_{\mathcal{H}}}
\newcommand{\InerH}[2]{\left\langle #1,#2\right\rangle_{\mathcal{H}}}
\newcommand{\Ind}{\mathds{1}}
\newcommand{\ind}{\Ind}
\newcommand{\bY}{\mathbf{Y}}

\newcommand{\Hb}{\mathbf{H}}
\newcommand{\phit}{\boldsymbol{\phi}}

\newcommand{\unitball}{\omega\in\mathcal{H}:\LH{\omega}=1}

\newcommand{\assumptions}{condition:observations,condition:domains,condition:model,condition:parameter_null,condition:mmd_kernel,condition:invertible}

\newtheorem{condition}{Assumption}
\crefname{theorem}{Theorem}{Theorems}
\crefname{condition}{Assumption}{Assumptions}
\ifdefined\proof
\renewenvironment{proof}[1]{\par\noindent{\bf Proof of #1\ }}{\hfill\BlackBox\\[2mm]}
\else
    \newenvironment{proof}[1]{\par\noindent{\bf Proof of #1\ }}{\hfill\BlackBox\\[2mm]}
\fi

\usepackage{lastpage}
\jmlrheading{26}{2025}{1-\pageref{LastPage}}{2/24; Revised 1/25}{2/25}{24-0276}{Oscar Key, Arthur Gretton, Fran\c{c}ois-Xavier Briol, and Tamara Fernandez}
\ShortHeadings{Composite Goodness-of-fit Tests with Kernels}{Key*, Gretton, Briol, and Fernandez*}
\firstpageno{1}

\begin{document}

\etocsettocdepth.toc{chapter}

\title{Composite Goodness-of-fit Tests with Kernels}

\author{%
\name Oscar Key\thanks{Equal contribution} \email oscar.key.20@ucl.ac.uk \\
\addr Centre for Artificial Intelligence, University College London \\
\AND
\name Arthur Gretton \email arthur.gretton@gmail.com \\
\addr Gatsby Computational Neuroscience Unit, University College London \\
\AND
\name Fran\c{c}ois-Xavier Briol \email f.briol@ucl.ac.uk \\
\addr Department of Statistical Science, University College London \\
\AND
\name Tamara Fernandez\footnotemark[1] \email tamara.fernandez@uai.cl \\
\addr Faculty of Engineering and Science, Adolfo Iba\~{n}ez University \\
}

\editor{Pierre Alquier}

\maketitle

\begin{abstract}%
  We propose kernel-based hypothesis tests for the challenging composite testing problem, where we are interested in whether the data comes from \emph{any} distribution in some \emph{parametric family}. Our tests make use of minimum distance estimators based on kernel-based distances such as the maximum mean discrepancy. As our main result, we show that we are able to estimate the parameter and conduct our test on the same data (without data splitting), while maintaining a correct test level. We also prove that the popular wild bootstrap will lead to an overly conservative test, and show that the parametric bootstrap is consistent and can lead to significantly improved performance in practice.
  Our approach is illustrated on a range of problems, including testing for goodness-of-fit of a non-parametric density model, and an intractable generative model of a biological cellular network.
\end{abstract}

\begin{keywords}
  bootstrap, hypothesis testing, kernel methods, minimum distance estimation, Stein's method
\end{keywords}

\section{Introduction}\label{sec:introduction}
Most statistical or machine learning algorithms are based on assumptions about the distribution of the observed data, of auxiliary variables, or of certain estimators.
Crucially, the validity of such assumptions directly impacts the performance of these algorithms.
To verify whether such assumptions are reasonable, one approach is  \emph{goodness-of-fit testing}, which considers the problem of rejecting, or not, the hypothesis that some data was generated by a \emph{fixed} distribution.
More precisely, given a distribution $\P$ and some observed data $x_1, \ldots, x_n$ from some distribution $\Q$, goodness-of-fit tests compare the null hypothesis $\nullnc: \P = \Q$ against the alternative hypothesis $\altnc: \P \neq \Q$.

There is a vast literature on goodness-of-fit testing, and the reader is referred to Chapter 14 in \citet{lehmann-testing-statistical-hypotheses} for a detailed introduction.
In the case of univariate real observations, a popular family of tests uses the \gls{cdf} as the test statistic, for example the Kolmogorov-Smirnov and Anderson-Darling tests \citep{Kolmogorov1933,Anderson1954}.
Various approaches have been taken to support multivariate observations, including
extending these \gls{cdf} tests \citep{justelMultivariateKolmogorovSmirnov}, developing a statistic based on the empirical characteristic function \citep{jimenez-gameroGoodnessoffitTestsBased}, or partitioning the observation space into bins and evaluating a test statistic on the resulting discrete empirical measures \citep{Gyorfi1991,Beirlant1994,Gyorfi2002,Beirlant2001}.
Unfortunately, the applicability of all of these tests is limited, because they rely on statistics that are often difficult to compute for complex models or high-dimensional observation spaces.

To overcome these issues, a class of kernel-based hypothesis tests has been proposed, which construct the statistic through either the \gls{mmd} \citep{lloyd-model-critcism-two-sample,Zhu2019,Balasubramanian2021,kellnerOnesampleTestNormality}, the \gls{ksd} \citep{liu-ksd-goodness-of-fit,chwialkowski-kernel-goodness-fit}, the energy distance \citep{Szekely2005,Szekely2013}, or the Hilbert-Schmidt independence criterion \citep{Sen2014}.
These tests can be applied to a wide-range of data types by choosing a test statistic based on an appropriate kernel which can be easily evaluated in the setting considered. Additionally, many of these tests are straightforward to apply even for intractable generative models for which no density exists, or for models with unnormalised likelihoods, making them particularly versatile.

In this paper, we consider the more complex question of whether our data comes from \emph{any element of some parametric family} of distributions. Let $\{\P_\theta\}_{\theta \in \Theta}$ denote a parametric family indexed by a parameter $\theta$ in some space $\Theta$.
Our test compares the null hypothesis
\begin{equation*}
    \nullc: \exists \thetaz \in \Theta \text{ such that } \P_{\thetaz} = \Q,
\end{equation*}
against the alternative hypothesis
\begin{equation*}
  \altc: \Q \not\in \{\P_\theta\}_{\theta \in \Theta}.
\end{equation*}
Such tests are known as \emph{composite goodness-of-fit tests}, and can be much more challenging to construct since $\thetaz$ is usually unknown.
They have the potential to answer important questions relating to model misspecification, allowing the user to confirm whether the parametric model they have selected is appropriate for their analysis.
Once again there is a significant literature tackling this problem, including tests which are specific to simple parametric families such as Gaussians \citep{shapiro1965analysis,lilliefors1967kolmogorov}, or composite versions of some of the non kernel-based tests mentioned above \citep{durbin1975KSTests,neyman1967chisqcomposite}.
Unfortunately, existing these tests all suffer from the same drawback as their simple goodness-of-fit counterparts, in that they are only applicable in a limited range of settings.

Our paper fills an important gap in the literature by proposing the first set of comprehensive \emph{kernel-based composite hypothesis tests}.
Given a kernel-based discrepancy $\D$, our test statistics take the form $\Delta = \min_{\theta \in \Theta} n \D(\P_\theta,\Q_n)$.
That is, we use the smallest discrepancy between $\Q_n$ and any element of the parametric family to determine whether $\nullc$ should be rejected or not.
In this work, we primarily consider the case where $\D = \MMD^2$, and include full theoretical and empirical analysis of the test's behaviour.
This analysis is challenging because, in contrast to existing kernel-based tests, the data now enters the test statistic twice: once to select the closest element of the parametric family, and a second time to estimate the discrepancy.
We also include encouraging empirical results for $\D = \KSD^2$, but leave the extension of our theoretical framework to this test for future work.

Our tests extend the advantages of kernel-based tests to the composite setting.
In particular, they can be directly applied to both generative or unnormalised models, two wide classes which cannot be tackled in full generality with classical composite tests.
Within the kernel testing literature, general composite tests have not yet been proposed.
Existing tests have either been limited to specific parametric families, such as multivariate Gaussians \citep{kellnerOnesampleTestNormality} or survival models with a specific survival function \citep{fernandezKernelizedSteinDiscrepancy}, or required splitting the data set thus reducing test power \citep{chwialkowski-kernel-goodness-fit,liu-ksd-goodness-of-fit}. \citet{Wolfer2022} also proposed a promising non-asymptotic composite test based on the MMD, but did not study it for generative models nor consider bootstrapping algorithms. We also show in our experiments that this approach leads to a more conservative test.

The remainder of the paper is as follows.
\Cref{sec:background}: we recall existing work on testing and parameter estimation with kernels.
\Cref{sec:methodology_test_statistic}: we  propose novel kernel-based composite hypothesis tests, and show that the MMD test statistic has well-behaved limiting distributions despite the reuse of the data for both estimation and testing.
\Cref{sec:methodology_bootstraps}: we compare two choices of bootstrap method that we can use to implement these tests in practice.
\Cref{sec:experiments}: we demonstrate these algorithms on a range of problems including an intractable generative model and non-parametric density estimation.
\Cref{sec:limitations}: we discuss limitations of our work.
\Cref{sec:connections}: we highlight connections with existing tests.
The code to reproduce our experiments, and implement the tests for your own models, is available at:
\url{https://github.com/oscarkey/composite-tests}

\section{Background}\label{sec:background}

We begin by reviewing the use of kernel discrepancies for testing and estimation. We  denote by $\X$ the data space and $\mathcal{P}(\X)$ the set of Borel probability distributions on $\X$.

\subsection{Kernel-based Discrepancies} \label{sec:discrepancies}

To measure the similarity of two distributions $\P,\Q \in \mathcal{P}(\mathcal{X})$, we can use a discrepancy $\D : \mathcal{P}(\X) \times \mathcal{P}(\X) \rightarrow [0,\infty)$.
We consider discrepancies related to \glspl{ipm} \citep{Muller1997}.
An \gls{ipm} indexed by the set of functions $\F$ has the form
\begin{equation} \label{eq:IPM}
  \D_\F(\P, \Q) =
  \sup_{f \in \F}
  \left| \E_{X \sim \P} [f(X)] - \E_{Y \sim \Q} [f(Y)] \right| \eqcom
\end{equation}
where $\F$ is sufficiently rich so that $\D_{\F}$ is a \emph{statistical divergence}; i.e. $\D_\F(\P, \Q) = 0 \Leftrightarrow \P = \Q$.

A first discrepancy is the \emph{maximum mean discrepancy} (MMD) \citep{grettonKernelMethodTwosampleproblem,grettonKernelTwosampleTest}.
Denote by $\H_K$ the \gls{rkhs} associated to the kernel $K : \X \times \X \rightarrow \Reals$ \citep{berlinetReproducingKernelHilbert}.
The \gls{mmd} is obtained by taking $\F_{\MMD} = \{f \in \H_K \given \norm{f}_{\H_K} \leq 1\}$,
which is a convenient choice because it allows the supremum in \Cref{eq:IPM} to be evaluated in closed form if $\E_{X\sim \P}[\sqrt{K(X,X)}]<\infty$ $\forall \P \in \mathcal{P}(\X)$:
\begin{equation} \label{eq:mmd_exact}
  \begin{split}
    &\MMD^2(\P, \Q) = \mathbb{E}_{X,X' \sim \P}[K(X, X')] - 2 \E_{\substack{X \sim \P, X' \sim \Q}} [K(X, X')]
    + \E_{X,X' \sim \Q}[K(X, X')].
  \end{split}
\end{equation}
Assuming we have access to independent and identically distributed realisations $\{\tx_i\}_{i=1}^m \iid \P$ and $\{x_j\}_{j=1}^n \iid \Q$, we can define $\P_m = \frac{1}{m} \sum_{i=1}^m \delta_{\tx_i}$ and $\Q_n = \frac{1}{n} \sum_{j=1}^n \delta_{x_j}$ (where $\delta_x$ is a Dirac measure at $x \in \X$) and we get the following estimate:
\begin{equation*}
  \MMD^2(\P_m, \Q_n)
  = \frac{1}{m^2}\sum_{i,j=1}^m K(\tx_i, \tx_j)
  - \frac{2}{n m } \sum_{i=1}^m \sum_{j=1}^n K(\tx_i, x_j)
  + \frac{1}{n^2} \sum_{i,j=1}^n K(x_i, x_j).
\end{equation*}
Furthermore, when $n=m$, this simplifies to
\begin{equation*}
  \MMD^2(\P_n, \Q_n)
  = \frac{1}{n^2} \sum_{i,j=1}^n  h_{\MMD}((x_i,\tx_i),(x_j,\tx_j)),
\end{equation*}
where
$h_{\MMD}((x_i,\tx_i),(x_j,\tx_j)) = K(\tx_i, \tx_j) + K(x_i, x_j) - K(\tx_i, x_j) - K(\tx_j, x_i)$.
This is known as a V-statistic \citep{grettonKernelTwosampleTest}, and $h_{\MMD}$ is known as the core of this statistic. The statistic is biased, but has smaller variance than alternative U-statistics for this quantity.

We also consider the \emph{kernel Stein discrepancy} (KSD) \citep{Oates2017,chwialkowski-kernel-goodness-fit,liu-ksd-goodness-of-fit}, which  is obtained by applying Stein operators to functions in some \gls{rkhs}; see \citet{Anastasiou2021} for a recent review.
Let $\X = \Reals^d$ and $\Hd = \H_K \times \ldots \times \H_K$ denote the $d$-dimensional tensor product of $\H_K$. The most common instance is the Langevin \gls{ksd} for which $\F_{\KSD} = \{ \mathcal{S}_{\P}[f] \given \norm{f}_{\Hd} \leq 1 \}$ where $\mathcal{S}_{\P}[g](x) = g(x) \cdot \nabla_x \log p(x) + \nabla_x \cdot g(x)$ is the Langevin Stein operator, $p$ is the Lebesgue density of $\P$ and $\nabla_x  = (\partial/\partial x_1,\ldots, \partial/\partial x_d)^\top$. With this choice, we obtain the discrepancy:
\begin{equation*}
  \KSD^2(\P, \Q)
  = \mathbb{E}_{X,X' \sim \Q}[h_{\KSD}(X,X')] \eqcom
\end{equation*}
\begin{equation*}
  \begin{split}
    \text{where } \qquad   h_{\KSD}(x,x')
    & = K(x, x') \nabla_{x}\log p(x) \cdot\nabla_{x'}\log p(x')
    + \nabla_{x}\log p(x) \cdot \nabla_{x'} K(x, x')        \\
    &\qquad + \nabla_{x'}\log p(x') \cdot \nabla_{x} K(x, x')
    + \nabla_{x} \cdot \nabla_{x'} K(x, x') \eqstop
  \end{split}
\end{equation*}
Given $\Q_n = \frac{1}{n} \sum_{j=1}^n \delta_{x_j}$, we obtain a V-statistic $\KSD^2(\P, \Q_n)= \frac{1}{n^2} \sum_{i,j=1}^n h_{\KSD}(x_{i},x_j)$.

We choose the \gls{mmd} and the \gls{ksd} to implement our tests because they are applicable in complementary settings.
The MMD can be estimated whenever samples from $\P_\theta$ are available.
This makes the corresponding test particularly suitable for simulator-based models (also called generative models); i.e. models which can be represented through a pair $(\U,G_\theta)$ such that sampling $x \sim \P_\theta$ involves sampling $u \sim \U$, and setting $x = G_\theta(u)$.
This class of models covers many models widely used in machine learning, including variational autoencoders \citep{kingmaAutoEncodingVariational} and generative adversarial networks \citep{liGenerativeMomentMatching,Dziugaite2015}, but also in the sciences including in synthetic biology \citep{bonassiBayesianLearningMarginal} and telecommunication engineering \citep{bhartiGeneralMethodCalibrating} to name just a few.
The interested reader is referred to \citet{Cranmer2020} for a recent review.
On the other hand, the KSD requires pointwise evaluations of $\nabla \log p$ and the KSD-based test will therefore be particularly suitable when the density of $\P_\theta$ is tractable.
This includes cases where the density can be only evaluated up to normalisation constant, such as for deep energy models or nonparametric density estimation models \citep{canuKernelMethodsExponential,Fukumizu2009,sriperumbudurDensityEstimationInfinite,Matsubara2021}, exponential random graphs \citep{Robins2007} and large spatial models based on lattice structures \citep{Besag1974}.

Throughout this paper, we will assume that our kernel-based divergences are valid statistical divergences.
This can be guaranteed for the MMD by assuming the kernel $K$ is characteristic \citep{Sriperumbudur2009}.
For the KSD  we require that $K$ is strictly integrally positive definite and
$\E_{X \sim \Q}[|\nabla_x \log p(X) - \nabla_x \log q(X)|] < \infty$ (see for example \citet[Theorem~2.1]{chwialkowski-kernel-goodness-fit} and \citet[Proposition~1]{barpMinimumSteinDiscrepancy}).
Examples of kernels on $\R^d$ which satisfy these assumptions include the Gaussian, inverse multiquadric and Mat\'ern kernels.

While these conditions ensure that the MMD and KSD can distinguish any pair of distributions when $n,m \to \infty$, for finite samples this is not necessarily the case.
The ability of the divergences to distinguish distributions is primarily dependent whether an appropriate kernel has been selected for the data in question.
Additionally, the KSD suffers from a failure mode when applied to multi-modal distributions, common for methods based on $\nabla_x \log p$:
if $\Q$ contains isolated modes with areas of near-zero density between them, the KSD cannot distinguish this from $\P$ that is identical except having a different weighting of the modes \citep{wenliangBlindnessScorebasedMethods}.
We discuss the impacts of these failure modes on ours tests in \cref{sec:limitations}.

\subsection{Goodness-of-fit Testing with Kernels}

We now recall how divergences lend themselves naturally to (standard) goodness-of-fit testing.
Let $\D:\X \times \X \rightarrow [0,\infty)$ be some statistical divergence, and recall that we are testing $\nullnc: \P = \Q$.
A natural approach is to compute $\D(\P,\Q)$ and check whether it is zero (in which case $\nullnc$ holds) or not (in which case $\altnc$ holds).
Since we only have access to independent realisations $\{x_j\}_{j=1}^n$ instead of $\Q$ itself, this idealised procedure is replaced by the evaluation of $\D(\P,\Q_n)$.
The question then becomes whether or not $\D(\P,\Q_n)$ is further away from zero than we would expect under $\nullnc$ given a data set of size $n$.

Kernel-based discrepancies can be computed in a wide range of scenarios including for intractable models. These divergences also have favourable sample complexity in that $\D(\P,\Q_n)$ converges to $\D(\P,\Q)$ at a square-root $n$ rate for both \gls{ksd} \citep[Theorem~4]{Matsubara2021} and \gls{mmd} \citep[Lemma~1]{briolStatisticalInferenceGenerative}. The \gls{mmd} was first used by \citet{grettonKernelMethodTwosampleproblem,grettonKernelTwosampleTest} for two-sample testing, and then studied for goodness-of-fit testing by \citet{lloyd-model-critcism-two-sample}. \gls{mmd} tests are closely related to many classical tests; see \citet{Sejdinovic2013}. \citet{chwialkowski-kernel-goodness-fit,liu-ksd-goodness-of-fit} introduced an alternate test based on $\KSD^2(\P,\Q_n)$.

To determine when $\nullnc$ should be rejected, we select an appropriate threshold $c_\alpha \in \R$, which will depend on the level of the test $\alpha \in [0,1]$. More precisely, $c_\alpha$ should be the $(1-\alpha)$-quantile of the distribution of the test statistic under $\nullnc$. This distribution will usually be unknown, but can be approximated using a bootstrap method. A common example is the wild bootstrap \citep{shao-dependent-wild-bootstrap,leucht-dep-wild-bootstrap}, which was specialised for kernel tests by \citet{chwialkowski-wild-bootstrap-kernel-tests}. We will return to this point in Section \ref{sec:methodology_bootstraps}.

\subsection{Minimum Distance Estimation with Kernels}

A statistical divergence $\D$ can also be used for parameter estimation through \emph{minimum distance estimation} \citep{Wolfowitz1957,Parr1980}.
Given a parametric family $\{\P_\theta\}_{\theta \in \Theta} \subset \PX$ indexed by $\Theta \subseteq \R^p$ and some data $\{x_i\}_{i=1}^n \iid \Q$, a natural estimator is
\begin{equation*}
  \hat{\theta}_n := \argmin_{\theta \in \Theta} \D(\P_\theta,\Q_n),
\end{equation*}
In practice, the minimum is often obtained using numerical optimisation algorithms. Under regularity conditions on $\D$ and $\P_\theta$, the estimator approaches $\theta^* := \argmin_{\theta \in \Theta} \D(\P_\theta,\Q)$ as $n \rightarrow \infty$. In particular, when the model is well-specified (i.e. $\nullc$ holds), then $\theta^* = \thetaz$.

The \gls{mmd} was first used for parameter estimation for neural networks by \citet{Dziugaite2015,liGenerativeMomentMatching,sutherlandGenerativeModelsModel,liMMDGANDeeper,Binkowski2018}, then studied more broadly as a statistical estimator by \citet{briolStatisticalInferenceGenerative,cherief-abdellatifFiniteSampleProperties,Cherief-Abdellatif2019-MMDBayes,Niu2021,Dellaporta2022,Bharti2023}. They closely relate to minimum scoring rule estimators based on the kernel-scoring rule \citep{Dawid2007}.
The use of \gls{ksd} for parameter estimation was first proposed by \citet{barpMinimumSteinDiscrepancy}, and later extended by \citet{Betsch2020,Grathwohl2020,Gong2020,Matsubara2021}. These estimators are also closely related to score-matching estimators \citep{Hyvarinen2006}; see \citet[Theorem~2]{barpMinimumSteinDiscrepancy} for details.

\section{Methodology: The Test Statistic}
\label{sec:methodology_test_statistic}

We are now ready to present our composite goodness-of-fit tests. The first section describes our general framework and studies the asymptotic distribution of the MMD test statistic. In the next section, we then describe how to implement the tests through bootstrapping.

\subsection{Composite Goodness-of-fit Testing with Kernels}

The high-level idea behind our test-statistics is to use the smallest kernel-based discrepancy value between the data $\Q_n$ and any element of the parametric family:
\begin{align*}
  \Delta & := \min_{\theta\in \Theta} n \D(\P_\theta,\Q_n)
  = \min_{\P \in \{\P_\theta\}_{\theta \in \Theta}} n \D(\P,\Q_n).
\end{align*}
To implement the tests, we propose a two-stage approach:
\begin{description}
  \item[Stage 1 (Estimate):] Compute $\thetah_n = \argmin_{\theta \in \Theta} \D(\P_\theta, \Q_n)$, so that $\Delta = n \D(\P_{\thetahn},\Q_n)$.
  \item[Stage 2 (Test):] Compute or estimate a test threshold $c_\alpha$ from the distribution of test statistic under $\nullc$. If $\Delta > c_\alpha$ reject $\nullc$, else do not reject.
\end{description}
In this paper, we will study two other instances of this framework. Our primary focus will be on the \emph{\gls{mmd} composite goodness-of-fit test}, where $\D = \MMD^2$. The next section provides regularity conditions so that as $n \to \infty$, $\Delta$ for this test converges to an infinite sum of $\chi^2$ random variables under $\nullc$ but diverges under $\altc$, thus allowing us to distinguish between the two hypotheses. The \gls{mmd} test is widely applicable since it only requires that we can sample from $\P_\theta$.

The second instance will be the \emph{\gls{ksd} composite goodness-of-fit test}, where $\D = \KSD^2$. Although we do not provide theoretical guarantees for this case, we do conjecture that similar guarantees exist. This test is particularly interesting since it only requires a tractable (possibly unnormalised) density function for $\P_\theta$.

In both tests, stage 1 requires computing the solution to the optimisation problem $\thetah_n = \argmin_{\theta \in \Theta} \D(\P_\theta, \Q_n)$.
The best method for this depends on the model and should be selected alongside it.
In this paper, for the MMD test, we use gradient-based stochastic optimisation and automatic differentiation.
For the KSD test, we use the closed-form expression for the minimiser, which is available when $\P_\theta$ is a member of the exponential family of distributions (see \cref{app:closed-form-ksd-estimator}).
However, stochastic optimisation could also be used in order to work with more general families of models.

\subsection{Theoretical Analysis of the MMD Composite Goodness-of-fit fest}
To present our analysis of the MMD test, we first define some additional notation.
Since we will be interested in asymptotic distributions, we will now consider the data to be random variables $X_1,\ldots,X_n \overset{i.i.d.}{\sim} \Q$, rather than the realisations $x_1,\ldots,x_n$ introduced so far.
We then overload $\Q_n = \frac{1}{n} \sum_{i=1}^n \delta_{X_i}$ to define a random measure. This is done in order to simplify notation, and whether we refer to the random or realised $\Q_n$ can be inferred from the context.

The results we present in this paper are based on the asymptotic distribution of $n\MMD^2(\P_{\thetah_n},\Q_n)$ as $n \rightarrow \infty$.
This is similar to standard kernel-based goodness-of-fit tests which use $n\MMD^2(\P,\Q_n)$, the main difference being that our test statistic depends on the data through both $\thetah_n$ and $\Q_n$.
In other words, we are now using the data twice, once for estimation and once for testing.
We note that this creates some non-trivial dependence and makes the derivation of theoretical guarantees significantly more challenging. We first state and discuss our assumptions, then present all results in the following subsection.

\begin{figure}
  \centering
  \begin{subfigure}[b]{0.49\linewidth}
    \centering
    \includegraphics{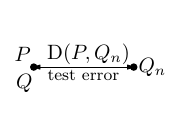}
    \caption{Non-composite test}
  \end{subfigure}
  \begin{subfigure}[b]{0.49\linewidth}
    \centering
    \includegraphics{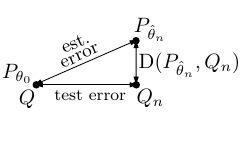}
    \caption{Composite test}
  \end{subfigure}
  \caption{
    Illustration of the sources of error when estimating the test statistic under the null hypothesis.
    Existing non-composite tests use the test statistic $n\D(\P,\Q_n)$, where $\D$ is a statistical divergence, and thus encounter only ``test'' error.
    The composite test we introduce uses the statistic $n\D(\P_{\thetahn},\Q_n)$, and thus encounters both ``estimation'' and ``test'' error.
  }
  \label{fig:error_diagrams}
\end{figure}

\subsubsection{Assumptions}
\label{sec:mmd_assumptions}

We will write $\mathcal{C}^m(\X)$ for the set of $m$-times continuously differentiable functions on $\X$, and $\mathcal{C}^m_b(\X)$ for the subset of $\mathcal{C}^m(\X)$ where all of the $m$ derivatives are bounded. For bivariate functions, we will denote by $\mathcal{C}^{m,m}_b(\X \times \X)$ the set of function which are $\mathcal{C}^{m}_b(\X \times \X)$ with respect to each input. We can now present our first assumption.

\begin{condition} \label{condition:observations}
  The observations $\{x_i\}_{i=1}^n$ are independent and identically distributed realisations from $\Q$.
\end{condition}
This assumption is common, and covers a very large class of applications, including all our examples in \Cref{sec:experiments}. It could be relaxed to dependent data but this would require the development of specialised tools and notions of dependence which are beyond the scope of this paper; see for example \citet{cherief-abdellatifFiniteSampleProperties}.

\begin{condition} \label{condition:domains}
  $\X \subseteq \R^d$, $\X$ is separable, and $\Theta$ is a compact and convex subset of $\R^p$.
\end{condition}
The assumptions are relatively mild and needed for the consistency of our estimators. Even when working with models where the assumptions on $\Theta$ are not satisfied, it is often possible to reparameterise the model so that the assumption holds.

Our next assumption defines regularity assumptions on the model considered by the MMD test, which will have to be verified on a per-model basis.

\begin{condition}\label{condition:model}
  $\{\P_\theta\}_{\theta \in \Theta}$ is an identifiable parametric family of models.
  Each element $\P_\theta \in \mathcal{P}$ has density $p_\theta$ with respect to a common measure $\lambda$ which satisfies:
  \begin{itemize}
    \item For each fixed $x \in \mathcal{X}$, $p_\cdot(x) \in \mathcal{C}^3(\Theta)$ and $p_{\theta}:\X\to\mathbb{R}$ is measurable for almost all $\theta\in\Theta$.

    \item For any collection $\ell,i,j \in \{1,\ldots,p\}$, the following holds true:
    \begin{align*}
      & \int_{\X}\sup_{\theta\in\Theta}\big|\frac{\partial}{\partial\theta_\ell}p_{\theta}(x)\big|\lambda(dx)<\infty, \\
      & \int_{\X}\sup_{\theta\in\Theta}\big|\frac{\partial^2}{\partial\theta_\ell\partial\theta_i}p_{\theta}(x)\big|\lambda(dx)<\infty, \\
      \text{and }
      & \int_{\X}\sup_{\theta\in\Theta}\big|\frac{\partial^3}{\partial\theta_\ell\partial\theta_i\partial\theta_j}p_{\theta}(x)\big|\lambda(dx)<\infty.
    \end{align*}
  \end{itemize}
\end{condition}
The smoothness assumptions with respect to the parameter are required to ensure convergence of $\hat{\theta}_n$ to $\thetaz$ in an appropriate sense under $\nullc$. Note that this assumption, together with the fact that $\Theta$ is compact, allows us to deduce $\int_{\X} \sup_{\theta\in\Theta}p_{\theta}(x)\lambda(dx)<\infty$. On the other hand, the smoothness assumptions with respect to $\X$ are needed for the divergences to be well defined.

In many cases, when applying the MMD test, we will not have access to a computable density function and will only be able to sample from the model.
For example, many models are defined in terms of a distribution $\U$ on some latent space, and a generator function $G_\theta:\calU \rightarrow \X$, where a sample from the model $\tx = G_\theta(u)$ with $u \sim \U$.
In this case, the results could be updated to instead place assumptions on $\U$ and $G_\theta$ rather than on the density, as was done in \citet{briolStatisticalInferenceGenerative}.

We make an additional assumption about the model family under the null hypothesis.
\begin{condition} \label{condition:parameter_null}
  Under $\nullc$, we have that $\thetaz$, the parameter value under the null hypothesis, belongs to the interior of $\Theta$.
\end{condition}
For most models this assumption will be met, or could be met by reparameterising the model.
Our next assumption relates to the kernel underlying the MMD.
\begin{condition} \label{condition:mmd_kernel}
  The kernel $K:\X\times\X\to\R$ is bounded, continuous and characteristic.
\end{condition}
This assumption is mild and will be satisfied by kernels commonly used in the literature on minimum MMD estimators, such as the Gaussian and Mat\'ern kernels of sufficient smoothness.

Our final assumption concerns the regularity of the divergences around the minimisers. This is necessary to guarantee convergence of our minimum distance estimators under $\nullc$, and is a mild assumption which is common in the literature (see e.g. Theorem 2 in \citet{briolStatisticalInferenceGenerative} and Theorem 4 in \citet{barpMinimumSteinDiscrepancy}).
\begin{condition} \label{condition:invertible}
  The Hessian matrix $\Hb\in\Reals^{p\times p}$ given by $\mathbf{H}_{ij}=\frac{\partial^2}{\partial\theta_i\partial\theta_j}\MMDk^2(P_\theta,P_{\theta_0})\big|_{\theta=\theta_0}$ for any $i,j\in \{1,\ldots,p\}$ is positive definite.
\end{condition}

\begin{remark}
  We could extend the results to the KSD test by noting that the KSD is equivalent to an MMD with a specific kernel \citep{barpMinimumSteinDiscrepancy}.
  However, this kernel is often unbounded and depends on $p_\theta$.
  Thus, additional assumptions on $p_\theta$ would be required.
  We leave this extension to future work.
  In our experiments we assume that the KSD is a valid statistical divergence, which holds under the conditions discussed in \Cref{sec:discrepancies}.
\end{remark}

\subsubsection[Behaviour under Null and Alternative]{Behaviour of the Test Statistic Under $\nullc$ and $\altc$}
We now present results on the asymptotic distribution of our test statistic, and the power (i.e. probability of correctly rejecting $\nullc$ when it does not hold) of the associated tests, as $n \to \infty$.
\begin{theorem}[Convergence under the null hypothesis] \label{Thm:2}\label{theorem:mmd_null_convergence}
  Under $\nullc$ and \Cref{\assumptions}:
  \begin{align}
    n\MMD^2(\P_{\thetahn},\Q_n)
    \todist \sum_{i=1}^\infty\lambda_iZ_i^2 \qquad \text{as } n\rightarrow \infty, \label{eqn:8}
  \end{align}
  where $(Z_i)_{i\geq 1}$ are a collection of i.i.d. standard normal random variables and
  $(\lambda_i)_{i \geq 1}$ are constants, which depend on the choice of kernel, where $\sum_{i=1}^\infty \lambda_i < \infty$.
\end{theorem}

\Cref{theorem:mmd_null_convergence} guarantees that our test statistic has a well-behaved limiting distribution under the null hypothesis $\nullc$.
Unfortunately, this distribution is not tractable, but knowing that it exists and is well-behaved is still necessary for guaranteeing that we can approximate quantiles with a bootstrapping method, as will be discussed in \Cref{sec:methodology_bootstraps}. The results are analogous to Theorem 12 in \citet{grettonKernelTwosampleTest},
though this result is only valid for testing $\nullnc \text{ vs. } \altnc$ and not $\nullc \text{ vs. } \altc$. Our second result concerns the power of the test: when $\mathcal{H}_1^{\text{C}}$ holds, we show that our test is able to reject $\mathcal{H}_0^\text{C}$ when $n$ is large enough.
\begin{theorem}[Consistency under the alternative hypothesis]\label{Thm:alte}\label{theorem:mmd_consistency}
  Under \Cref{\assumptions}, we have that $\altc$ holds if and only if $\liminf_{n\to\infty} \MMDk^2(\P_{\thetahn},\Q_n)>0$.
\end{theorem}
As our test statistic, $n \MMD^2(\P_{\thetahn},\Q_n)$, is scaled by $n$, the above result implies that the test statistic diverges as $n \to \infty$.

\section{Methodology: Practical Implementation via Bootstraps}
\label{sec:methodology_bootstraps}

To implement the tests, we require an estimate of the rejection threshold $c_\alpha$, which should be set to the $(1-\alpha)$-quantile of the distribution of $\Delta = n \D(\P_{\thetah_n},\Q_n)$ under $\nullc$.
Since this distribution is unknown, we estimate the threshold using bootstrap algorithms.
In following section we present two possible choices: the wild bootstrap and the parametric bootstrap, and analyse their behaviour when applied to the MMD test.

\subsection{The Wild Bootstrap} \label{sec:wild_bootstrap}

A natural first approach is to use the wild bootstrap, since that is the current state-of-the-art for non-composite kernel goodness-of-fit tests \citep{chwialkowski-kernel-goodness-fit,liu-ksd-goodness-of-fit,schrabKSDAggregatedGoodnessoffit}.
The algorithm is based on bootstrapped versions of the test statistics, defined for non-composite tests as
\begin{align*}
  n \MMD^2_W(\P_n, \Q_n)
   & \defeq \frac{1}{n} \sum_{i,j=1}^n W_i W_j
  h_{\MMD}((x_i,\tx_i), (x_j,\tx_j)) ,         \\
  n \KSD^2_W(\P, \Q_n)
   & \defeq \frac{1}{n} \sum_{i,j=1}^n W_i W_j
  h_{\KSD,\P}(x_i, x_j) ,
\end{align*}
where $W_1,\ldots,W_n$ are i.i.d. Rademacher random variables (i.e. random variables taking value $-1$ and $1$ with probability $1/2$ each). Note that the bootstrapped version of the MMD test requires realisations of both $P$ and $Q$ (i.e. $P_n$ and $Q_n$) whereas the bootstrapped version of the KSD test requires evaluations of the score of the distribution $P$ and samples from $Q$ (denoted through the empirical measure $Q_n$).

Given fixed observations $\Q_n$, it is possible to draw samples $\Delta^{(1)},\ldots,\Delta^{(b)}$ of the bootstrapped test statistics by taking fresh realisations of $W_1,\ldots,W_n$ for each sample.
If we consider an MMD test, under $\nullnc$, \citet[Theorem~1]{chwialkowski-wild-bootstrap-kernel-tests} show that the distributions of $n \MMD^2(\P,\Q_n)$ and $n \MMD_W^2(\P,\Q_n)$ converge to the same distribution as $n \to \infty$.
Thus, we can estimate the threshold $c_\alpha$ during a non-composite test by computing the $(1-\alpha)$-quantile of $\{\Delta^{(k)}\}_{k=1}^b$.
Under $\altnc$, \citet[Proposition 3.2]{chwialkowski-wild-bootstrap-kernel-tests} show that $n \MMD_W^2(\P_n,\Q_n)$  will converge to a finite quantity, while \Cref{theorem:mmd_consistency} shows that $n \MMD^2(\P_n,\Q_n) \to \infty$. This means that $\Prob(\MMD^2(\P_n,\Q_n) > c_{\alpha}) \to 1$ as $n\rightarrow \infty$ and the test rejects $\nullnc$ correctly.

We can extend the wild bootstrap to the composite case by first estimating the parameter and then applying the wild bootstrap to the estimated distribution.
Thus, the bootstrapped test statistics are $n \MMD_W^2(\P_{\thetahn, n},\Q_n)$ and $n \KSD_W^2(\P_{\thetahn},\Q_n)$, where $\P_{\thetahn, n} = \frac{1}{n} \sum_{i=1}^n \delta_{\bar{X}_i}$ with $\bar{X}_i \sim \P_{\thetahn}$.
\Cref{alg:wild_bootstrap_mmd} describes this approach for the MMD, and \Cref{alg:wild_bootstrap_ksd} for the KSD. Note that the KSD version of the test does not require an additional sampling step since the KSD can be estimated with a single sample, whereas two different samples are needed to estimate the MMD.
\begin{figure}
  \setlength{\algomargin}{0.2em}
  \begin{minipage}[t]{0.52\textwidth}
    \begin{algorithm}[H]
      \KwIn{$\P_{\thetahn}$, $\Q_n$, $\alpha$, $b$}
      $P_{\hat{\theta}_n,n} = \frac{1}{n} \sum_{i=1}^n \delta_{\tx_i}$ where $ \tx_1, \ldots, \tx_n \iid \P_{\thetahn}$\;
      \For{$k \in \{1, \ldots, b\}$}{
      $w^{(k)} = (w_1^{(k)},\ldots,w_n^{(k)}) \iid \operatorname{Rademacher}$\;
      $\Delta^{(k)} = n \MMD^2_{w^{(k)}}(P_{\hat{\theta}_n,n}, \Q_n)$\;
      }
      $c_\alpha = \operatorname{quantile}(\{\Delta^{(1)}, \ldots, \Delta^{(b)}\}, 1 - \alpha)$\;

      \caption{Wild bootstrap (MMD)}
      \label{alg:wild_bootstrap_mmd}
    \end{algorithm}
  \end{minipage}
  \hfill
  \begin{minipage}[t]{0.47\textwidth}
    \begin{algorithm}[H]
      \KwIn{$\P_{\thetahn}$, $\Q_n$, $\alpha$, $b$}
      \For{$k \in \{1, \ldots, b\}$}{
      $w_1^{(k)},\ldots,w_n^{(k)} \iid \operatorname{Rademacher}$\;
      $\Delta^{(k)} = n \KSD^2_{w^{(k)}}(\P_{\thetahn}, \Q_n)$\;
      }
      $c_\alpha = \operatorname{quantile}(\{\Delta^{(1)}, \ldots, \Delta^{(b)}\}, 1 - \alpha)$\;

      \caption{Wild bootstrap (KSD)}
      \label{alg:wild_bootstrap_ksd}
    \end{algorithm}
  \end{minipage}
\end{figure}

It is not clear from the existing results in the literature if applying the wild bootstrap to the composite test in this fashion will result in a test with a type I error rate (i.e. probability of wrongly rejecting $\nullc$ when it holds) smaller than or equal to $\alpha$.
This is because the composite test statistic contains an additional source of error, in comparison to the non-composite statistic.
$\MMD^2(\P_n,\Q_n)$ contains one source of error, due to estimating $\MMD^2(\P,\Q_n)$ through samples.
However, $\MMD^2(\P_{\thetahn,n},\Q_n)$ contains a second source of error because $\P_{\thetaz}$, the true data generating distribution under $\nullc$, is estimated with $\P_{\thetah_n}$.
\Cref{fig:error_diagrams} illustrates these sources of error in the non-composite and composite test statistics under the null hypothesis.
If we fix an estimate of $\thetahn$ and then apply the wild bootstrap, we fail to take into account this additional source of error.

\begin{figure}[t]
  \centering
  \includegraphics{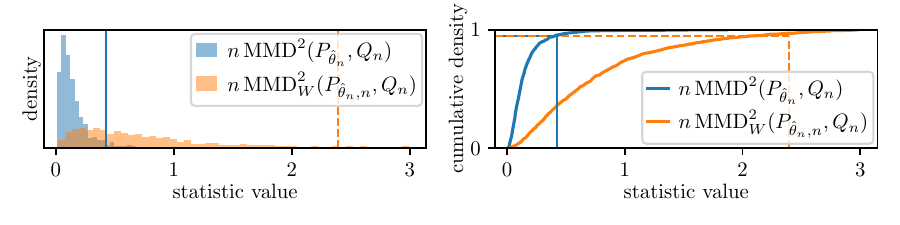}
  \caption[Distribution of the wild bootstrap statistic]{
    Distribution of $\MMD^2(\P_{\thetahn}, \Q_n)$ and $\MMD^2_W(\P_{\thetahn,n}, \Q_n)$ under $\nullc$, obtained by simulation.
    \tikz[baseline=0ex]{\draw [line width=0.5mm, solid, blue] (0,0.0ex) -- (0,1.6ex);}
    and
    \tikz[baseline=0ex]{\draw [line width=0.5mm, dashed, orange] (0,0.0ex) -- (0,1.6ex);}
    show the respective $(1-\alpha)$ quantiles, demonstrating that the wild bootstrap estimates a conservative threshold.
  }
  \label{fig:wild_bootstrap_distribution}
\end{figure}

Our main result for this section shows that applying the wild bootstrap does lead to a type I error rate smaller than or equal to $\alpha$ in the MMD test. However, our result shows the (asymptotic) type I error rate is strictly below $\alpha$, which is surprising as usually resampling strategies yield asymptotically-exact type-I error rates. This result suggests that our test is conservative and could potentially be improved further.
\begin{theorem}\label{thm:WBOscar}
  Let $\alpha\in(0,1)$, and define
  \begin{equation*}
    q_\alpha=\inf\left\{\gamma:\limsup_{n\to\infty}\Prob(n \MMD^2_W(\P_{\thetahn,n}, \Q_n)>\gamma)\leq \alpha\right\}.
  \end{equation*}
  Under the null hypothesis and \Cref{\assumptions}, it holds
  \begin{equation*}
    \limsup_{n\to\infty}\Prob\left(n\MMDk^2(P_{\thetahn},Q_n)\geq q_\alpha\right)<\alpha.
  \end{equation*}
\end{theorem}
The root cause of the above result is that, under both $\nullc$ and $\altc$, $
  \MMDk^2(P_{\thetahn},Q_n) \leq \MMDk^2(P_{\theta_0},Q_n)$,
because $\thetahn$ is chosen as the minimiser of $\theta\to\MMDk^2(P_\theta,Q_n)$.
The impact of this on the wild bootstrap is that, for all $x \in \Reals$,
\begin{equation*}
  \limsup_{n\to\infty}\Prob\left(n \MMD^2(\P_{\thetahn}, \Q_n)\geq x\right)
  <
  \limsup_{n\to\infty}\Prob\left(n \MMD^2_W(\P_{\thetahn,n}, \Q_n)\geq x\right)
\end{equation*}
as we show in the proof of \Cref{thm:WBOscar}.
This means that every quantile of the asymptotic distribution of the wild bootstrap statistic is shifted away from zero compared to the test statistic, as demonstrated in \cref{fig:wild_bootstrap_distribution}.
Thus, the threshold computed from the samples of the bootstrapped test statistic is too large, and the resulting test is conservative.
We conjecture that similar behaviour occurs under $\altc$, resulting in the loss of power we observe in the experiments (see e.g. \Cref{fig:gpc_increasing_n_alt}).
To achieve better performance, we can use an alternative bootstrap which does take account of estimation error.

\subsection{The Parametric Bootstrap}
We now consider the parametric bootstrap \citep{stuteBootstrapBasedGoodnessoffittests} presented in \Cref{alg:parametric_bootstrap}, which is commonly used in the composite testing literature (for example \citet{kellnerOnesampleTestNormality}).
To approximate the distribution of $n \D(\P_{\thetah_n},\Q_n)$ under $\nullc$, this approach first fits the parameter to the observations.
It then repeatedly resamples the observations, re-estimates the parameter and recomputes the test statistic.
By repeatedly re-estimating the parameter, the parametric bootstrap takes account of the estimation error.

\begin{algorithm}
  \KwIn{$\D$, $\P_{\theta}$, $\Q_n$, $\alpha$, $b$ (num bootstrap samples)}
  $\thetah_n = \argmin_\theta \D(\P_\theta, \Q_n)$\;
  \For{$k \in \{1, \ldots, b\}$}{
  $\Q_n^{(k)} = \frac{1}{n} \sum_{i=1}^n \delta_{x_i^{(k)}}$, $\big\{x_i^{(k)}\big\}_{i=1}^n \iid \P_{\thetah_n}$\;
  $\thetah^{(k)}_n = \argmin_{\theta \in \Theta} \D(\P_{\theta}, \Q_n^{(k)})$\;
  $\Delta^{(k)} = n \D(\P_{\thetah^{(k)}_n}, \Q_n^{(k)})$\;
  }
  $c_\alpha = \operatorname{quantile}(\{\Delta^{(1)}, \ldots, \Delta^{(b)}\}, 1 - \alpha)$\;

  \caption{Parametric bootstrap}
  \label{alg:parametric_bootstrap}
\end{algorithm}

\Cref{alg:parametric_bootstrap} generates samples $\Delta^{(k)}$ of the bootstrapped test statistic $n \D(\P_{\thetahn^\star}, \Q_n^\star)$,
where $\thetahn^\star = \argmin_{\theta \in \Theta} \D(\P_\theta, \Q^\star_n)$, and
$\Q^\star_n = \frac{1}{n} \sum_{i=1}^n \delta_{X^\star_i}$ with $X^\star_i \sim \P_{\thetahn}$.
We show that the parametric bootstrap is valid for the \gls{mmd} test, which once again requires that the distribution of the test statistic $n \MMD^2(\P_{\thetahn}, \Q_n)$ and the distribution of the bootstrapped test statistic $n \MMD^2(\P_{\thetahn^\star},\Q^\star_n)$ converge to the same distribution as $n \to \infty$.
To present the result we require some additional notation, which is defined formally in \Cref{app:parametric_bootstrap_definitions}: considering random variables $A_n$ and $A$, we write   $A_n \overset{\dist_D}{\to} A$ to indicate that $A_n$ converges to $A$ in distribution conditioned on the data, $\{X_i\}_{i=1}^n$.
\begin{theorem}[Parametric bootstrap under the null]\label{Thm:2B}\label{theorem:mmd_parametric_bootstrap_null}
  Under $\nullc$ and \Cref{\assumptions}, we have that $n\MMDk^2(P_{\theta_n^\star},\Q_n^\star)\overset{\dist_D}{\to}\Gamma_{\infty}$, where $\Gamma_\infty$ is a random variable with distribution given by \cref{eqn:8}, i.e. the distribution of the test statistic under $\nullc$.
\end{theorem}
Note that the above convergence holds conditioned on any sequence of observations, thus holds in general.
In \Cref{sec:experiments} we find empirically that applying the parametric bootstrap to our tests results in a good type I error rate and a better performance than the wild bootstrap when $n$ is small.
However, the parametric bootstrap is substantially more computationally intensive because it requires repeatedly computing a minimum distance estimator (and hence the kernel matrix) on fresh data, whereas the wild bootstrap only uses a single minimum distance estimator. Assuming we are computing the minimum distance estimators through $T$ steps of a numerical optimiser based on function evaluations or gradients, the computational complexity of the wild bootstrap algorithms is $O((T+b)n^2)$, whilst the parametric bootstrap is $O(T b n^2)$.
We therefore expect the parametric bootstrap to be significantly more expensive than the wild bootstrap.
However, this extra computation is likely an issue only for large $n$, and in this high data regime we find in our experiments that the wild bootstrap achieves similar power to the parametric bootstrap.
Thus, in a high data regime there is little penalty to using the cheaper wild bootstrap. Note that the cost could be further reduced to be linear in $n$ through alternative estimates of the MMD (see e.g. Lemma 14 of \cite{grettonKernelTwosampleTest}), but this would likely lead to substantially lower test power.

\section{Empirical Results}\label{sec:experiments}

In the following section, we now study the performance of our tests in a range of settings. We consider three examples: (i) a multivariate Gaussian location and scale model, (ii) a generative model of the interaction of genes through time called a toggle-switch model, and (iii) an unnormalised nonparametric density model. For all experiments we set the test level $\alpha = 0.05$, and give full implementation details in \Cref{appendix:experiment_details}.

\subsection{Gaussian Model}
We begin by exploring the different configurations of our test, in regard to the underlying discrepancy and the bootstrap method, for various Gaussian models.
Unless otherwise stated, we use a Gaussian kernel with lengthscale computed using the median heuristic \citep{grettonKernelMethodTwosampleproblem}.

\begin{figure}
  \centering
  \begin{subfigure}[t]{0.5\textwidth - 0.08in}
    \centering
    \includegraphics{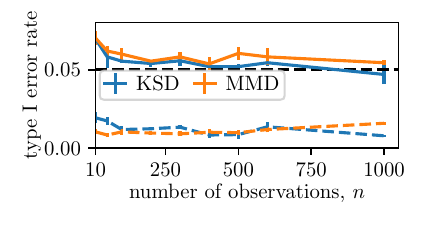}
    \caption{Type I error rate under $\nullc$, where $\Q = \Normal(\mu_0, 1^2)$.}
  \end{subfigure}
  \hfill
  \begin{subfigure}[t]{0.5\textwidth - 0.08in}
    \centering
    \includegraphics{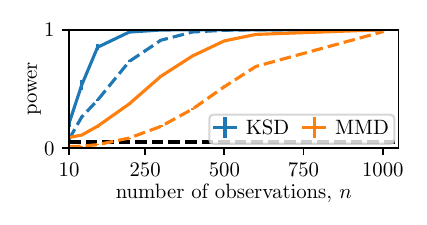}
    \caption{Power under $\altc$, where $\Q = \operatorname{Students-t}$ with $2$ degrees of freedom.}
    \label{fig:gpc_increasing_n_alt}
  \end{subfigure}
  \caption[Performance of the tests as n increases]{
    Performance of the MMD and KSD tests as $n$ increases, where $\nullc : \P = \Normal(\mu, 1)$.
    \tikz[baseline=-0.5ex]{
      \draw [line width=0.5mm, blue] (0,-0.3ex) -- (2ex,-0.3ex);
      \draw [line width=0.5mm, orange] (0,0.3ex) -- (2ex,0.3ex);
    } show the parametric bootstrap,
    \tikz[baseline=-0.5ex]{
      \draw [line width=0.5mm, dashed, blue] (0,-0.3ex) -- (2ex,-0.3ex);
      \draw [line width=0.5mm, dashed, orange] (0,0.3ex) -- (2ex,0.3ex);
    } show the wild bootstrap,
    \tikz[baseline=-0.5ex]{\draw [line width=0.5mm, dashed] (0,0) -- (2ex,0);} shows the level.
    The error bars show one standard error across $4$ random seeds.
  }
  \label{fig:gpc_increasing_n}
\end{figure}
We test Gaussianity by considering $\nullc : \P_{\theta} = \mathcal{N}(\mu, \Sigma)$ with unknown parameters $\theta=\{\mu, \Sigma\}$, for $\mu \in \R^d$ and $\Sigma \in (\R^+)^{d \times d}$.
We start with $d=1$, and test against a particular alternative where $\Q=$ Students-t with 2 degrees of freedom.
\Cref{fig:gpc_increasing_n} shows the type I error rate and power of the MMD and KSD tests using both bootstraps as $n$ increases.
Considering the MMD test under $\nullc$, the test using the wild bootstrap converges to a conservative type I error rate as $n$ becomes large, while the test using the parametric has the correct rate.
This is consistent with our analysis of the bootstraps in \cref{sec:methodology_bootstraps}.
Under $\altc$, the parametric bootstrap has higher power, once again due to the wild bootstrap being conservative, but both bootstraps achieve a power of one once $n$ is large enough.
For these results, $m$, the number of samples from $\P_\theta$, was set to $m=n$ as a balance between computational cost and power.
Better power could be achieved by increasing $m$, although this would increase the computational cost.

The figure also shows the same set of results for the KSD test.
It has similar convergence behaviour, suggesting that our theoretical results may also hold for the KSD.
Comparing the MMD and KSD tests under $\altc$ we note that the KSD test has higher power for any given $n$.
This is both because the KSD does not require sampling from $\P_\theta$, and because the MMD estimation process requires stochastic optimisation, whereas for the KSD we use a closed-form estimator.

\begin{figure}
  \centering
  \begin{subfigure}[t]{0.5\textwidth - 0.08in}
    \centering
    \includegraphics{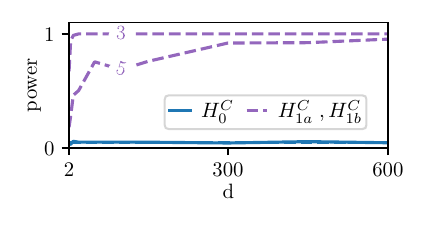}
    \caption{MMD test}
  \end{subfigure}
  \hfill
  \begin{subfigure}[t]{0.5\textwidth - 0.08in}
    \centering
    \includegraphics{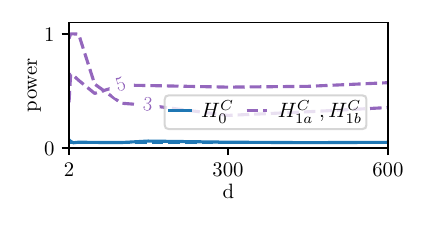}
    \caption{KSD test}
  \end{subfigure}
  \caption[Power of the tests as d increases]{
    Power of the parametric bootstrap tests as $d$ increases.
    $\nullc : \P = \Normal(\mu, \sigma^2)$, $H^C_{1a}$ and $H^C_{1b}$ indicate $\Q = \operatorname{Students-t}$, with $3$ and $5$ degrees of freedom respectively.
    \tikz[baseline=-0.5ex]{\draw [line width=0.5mm, dashed] (0,0) -- (2ex,0);} shows the level.
  }
  \label{fig:gpc_increasing_d}
\end{figure}
Second, we examine the performance of the parametric bootstrap tests as $d$, the dimensionality of the observations, increases.
We test against the family of multivariate Gaussians with known covariance, $\nullc: \P = \Normal(\mu_d, I_{d \times d})$ with unknown parameter $\theta = \mu_d \in \Reals^d$.
We consider two alternatives, $H^C_{1a}$ and $H^C_{1b}$, where we generate data from multivariate t-distributions with $\nu=3$ and $\nu=5$ degrees of freedom respectively.
As $\nu$ becomes large, the observed data become indistinguishable from a Gaussian, hence we expect the tests to have lower power against larger $\nu$.
We set the scale parameter, $\Sigma = I_{d \times d} (\nu - 2) / \nu$, so that the samples have covariance $I$ and the distance between the samples does not grow more rapidly than under $\nullc$ as $d$ increases.
\Cref{fig:gpc_increasing_d} shows that both tests maintain the type I error rate of $0.05$ under $\nullc$.
Under $H^C_{1a}$ and $H^C_{1b}$ we find that the power of the MMD test increases with $d$, while the power of the KSD test decreases with $d$.

\begin{figure}
  \centering
  \includegraphics{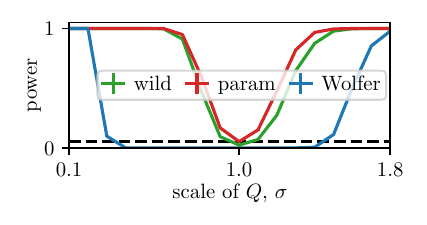}
  \caption[Comparison of MMD tests]{
    Comparison of the MMD tests against \citet{Wolfer2022}.
    $\nullc : \P = \Normal(\mu, 1)$ and $\Q = \Normal(\mu_0, \sigma^2)$.
    $n=200$.
    \tikz[baseline=-0.5ex]{\draw [line width=0.5mm, dashed] (0,0) -- (2ex,0);} shows the level.
    The error bars show one standard error across $4$ random seeds.
  }
  \label{fig:gpc_ours_vs_variance_aware}
\end{figure}
Next, we further examine the power of the MMD tests when using the wild and parametric bootstraps.
We also include the non-asymptotic MMD test introduced by \citet{Wolfer2022}, for comparison.
We take $\nullc$ to be a family of one dimensional Gaussians with known variance, specifically $\P = \Normal(\mu, \sigma^2)$, where $\theta = \mu \in \Reals$ and $\sigma^2 = 1$.
We generate data from $\Normal(\mu_0, \sigma^2)$, where $\mu_0 \in \Reals$ is an arbitrary mean, and compute the power of the tests for different values of $\sigma^2$.
An ideal test would have power close to $\alpha = 0.05$ when $\sigma^2 = 1$, and power close to one otherwise.
\Cref{fig:gpc_ours_vs_variance_aware} shows the results, with the parametric bootstrap having higher power than the wild bootstrap once again.
The test introduced by \citet{Wolfer2022} achieves lower power than both our tests with either bootstrap.

We also consider the time taken by each bootstrap method.
For $n=100$ and $d=1$, we found that each wild bootstrap test took approximately $1$ms, while each parametric bootstrap test approximately $5$ms.
These figures depend heavily on the implementation and hardware details (as given in \Cref{appendix:experiment_details}).
In particular, we execute all bootstrap samples in parallel on a GPU, which is particularly beneficial for the parametric bootstrap where each sample has a high computational cost.
However, it will not be possible for larger models because it requires keeping $b$ copies of the model parameters in memory.
Additionally, it would be possible to further optimise the bootstrap implementations.
Despite this, these approximate numbers illustrate that the parametric bootstrap test is much more expensive.
Thus, for large $n$ where the two bootstraps achieve similar power (see \cref{fig:gpc_increasing_n_alt}) it may be better to use the wild bootstrap.

\begin{figure}
  \centering
  \includegraphics{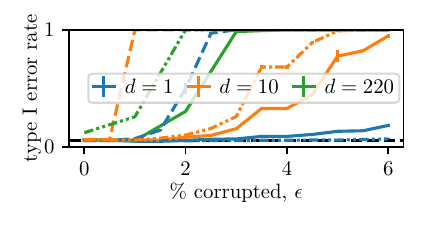}
  \caption[Robustness of the tests]{
    Type I error rate of the tests under corrupted observations.
    $\nullc : \P = \Normal(\mu_d, \Sigma_{d \times d})$ and
    $\Q = (1-\epsilon) \P + \epsilon \Normal(10_d, 0.2 I_{d \times d})$.
    $n=200$.
    \tikz[baseline=-0.5ex]{
      \draw [line width=0.5mm, solid, blue] (0,-0.5ex) -- (2ex,-0.5ex);
      \draw [line width=0.5mm, solid, orange] (0,0.0ex) -- (2ex,0.0ex);
      \draw [line width=0.5mm, solid, green] (0,0.5ex) -- (2ex,0.5ex);
    } show the MMD,
    \tikz[baseline=-0.5ex]{
      \draw [line width=0.5mm, dashed, blue] (0,-0.5ex) -- (2ex,-0.5ex);
      \draw [line width=0.5mm, dashed, orange] (0,0.0ex) -- (2ex,0.0ex);
      \draw [line width=0.5mm, dashed, green] (0,0.5ex) -- (2ex,0.5ex);
    } the KSD,
    \tikz[baseline=-0.5ex]{
      \draw [line width=0.5mm, dash dot, blue] (0,-0.5ex) -- (2ex,-0.5ex);
      \draw [line width=0.5mm, dash dot, orange] (0,0.0ex) -- (2ex,0.0ex);
      \draw [line width=0.5mm, dash dot, green] (0,0.5ex) -- (2ex,0.5ex);
    } the KSD with robust kernel,
    and \tikz[baseline=-0.5ex]{\draw [line width=0.5mm, dashed, black] (0,0) -- (2ex,0);} the level.
    The error bars show one standard error across $4$ random seeds.
  }
  \label{fig:robustness}
\end{figure}
\Cref{fig:robustness} explores the robustness of the test to corrupted observations under Huber's contamination model \citep{huber-book}, as previously explored in the testing setting by \citet{Liu2024,Schrab2024}.
Once again we consider $\nullc: \P = \Normal(\mu_d, I_{d \times d})$, for $\mu_d \in \Reals^d$.
We assume that $\nullc$ holds, but some observations have been corrupted with a large amount of noise.
Thus, we generate data from
$\Q = (1-\epsilon) \Normal(\bm{0}_d, I_{d \times d}) + \epsilon \Normal(\bm{10}_d, 0.2 I_{d \times d})$, where $0 \leq \epsilon \leq 1$ controls the percentage of observations that are corrupted.
A robust test would maintain the correct type I error rate as $\epsilon$ grows.
We evaluate the MMD test and KSD test with a Gaussian kernel (as defined in Appendix C.2).
We also include a robust KSD test using the tilted kernel
$K'(x, x') = w(x) K(x, x') w(x')$, where $K$ is the Gaussian kernel,
$w(x) = (1 + a^2 \norm{x}_2^2)^{-b}$,
and we choose $a=1$, $b=\half$
\citep{barpMinimumSteinDiscrepancy,Liu2024}.
The results on one-dimensional observations, $d=1$, show that the MMD test is robust up to 2-3\% corruption, the KSD test with Gaussian kernel is not robust, and the KSD test with tilted kernel is robust to $>6$\% corruption.
All the tests become less robust as $d$ increases.

\begin{figure}
  \centering
  \includegraphics{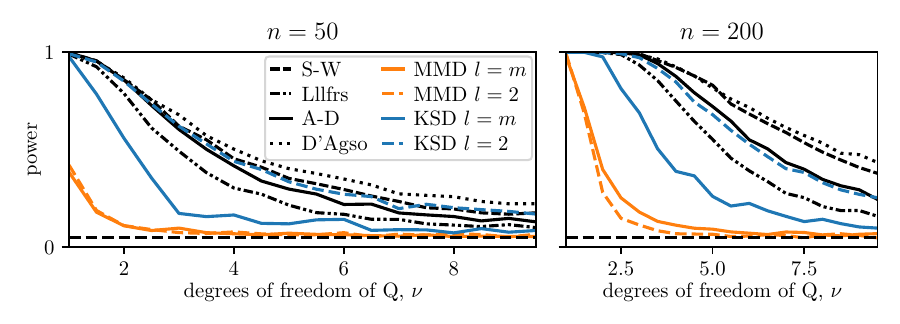}
  \caption[Comparison to specialised Gaussianity tests]{
    Comparison of the \gls{mmd} and \gls{ksd} tests (parametric bootstrap), and the Shapiro-Wilks (S-W), Lilliefors (Lllfrs), Anderson-Darling (A-D) and D'Agostino's (D'Agso) tests.
    $\nullnc : \P = \Normal(\mu, \sigma^2)$, and $\Q = $ a series of Student's t-distributions.
    Two kernel lengthscales are shown for our tests $l=m$ (the median heuristic) and $l=2$ (selected via grid search).
    \tikz[baseline=-0.5ex]{\draw [line width=0.5mm, dashed] (0,0) -- (2ex,0);} shows the level.
  }
  \label{fig:gpc_other_tests}
\end{figure}
To complete the experiments on Gaussian models, we compare the performance of our tests to existing specialised tests of Gaussianity.
We consider three nonparametric tests: the Anderson-Darling, Lilliefors (the Kolomogorov-Smirnov test specialised to Gaussians) and Shapiro-Wilks tests.
We also consider D'Agostino's $K^2$ parametric test \citep{dagostinoTestsDepartureNormality}.
For this comparison we return to testing $\nullc : \P_{\theta} = \Normal(\mu, \sigma^2)$, for unknown parameters $\theta = \{\mu, \sigma^2\}$, in one dimension.
We perform the test against data generated from a Student's t-distribution, and vary the degree of freedom $\nu$.
Once again, we expect the test power to fall to $\alpha = 0.05$, the specified level of the test, as $\nu$ becomes large.
\Cref{fig:gpc_other_tests} shows the results of this experiment, for both tests using the parametric bootstrap and for two choices of lengthscale.
The MMD test has substantially lower power than the other tests for both lengthscales.
The KSD test is an improvement, having performance closer to the specialised tests when setting the lengthscale with the median heuristic.
Increasing $n$ improves the power of each test, but does not change the ordering.
These results reveal that we should prefer tests specifically designed for the model in question where possible, as opposed to our generally applicable tests.
The advantage of our tests is that they are applicable beyond Gaussianity, including for unnormalised or generative models. They are also applicable to multivariate data, which is not the case for the Anderson-Darling or Lilliefors tests.
In this plot we also demonstrate the importance of choosing the kernel.
We include the power of our test for a more optimal choice of $l=2$, which we found by sampling additional observations and performing a grid search.
For this kernel, the KSD test has comparable power to the specialised tests.
However, in practice we cannot use this method to select the kernel, as we are not able to sample additional observations.
Future work could look at better strategies than the median heuristic for selecting the bandwidth parameter in composite tests.

\subsection{Toggle-Switch Model}
\begin{figure}
  \centering
  \includegraphics{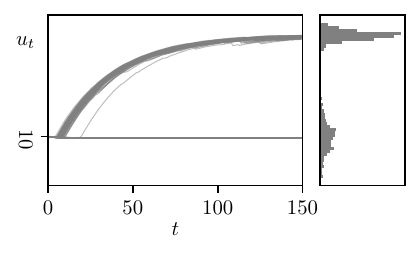}
  \caption{
    \emph{Left:} Evolution of the $u_t$ variable of the toggle switch model, starting with initial state $u_t=10$.
    Each line shows a different random evolution of the model.
    \emph{Right:} Histogram of the resulting values of $y$.
  }
  \label{fig:toggle_switch_time_evo}
\end{figure}
\begin{figure}
  \centering
  \includegraphics{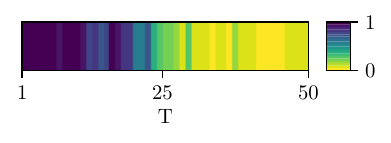}
  \caption{Fraction out of $20$ repeats for which the MMD test (parametric bootstrap) rejects $\nullc : \P = \Pts[\theta,T]$, comparing against data generated from $\Q = \Pts[\theta_0,300]$.}
  \label{fig:toggle_switch_increasing_T}
\end{figure}
\begin{figure}
  \centering
  \includegraphics{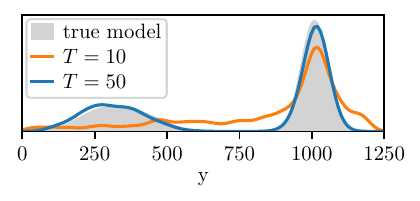}
  \caption{
    Fit of the toggle switch model for $T=10$ (for which $\nullc$ is rejected) and $T=50$ (for which it is not).
    The densities are generated using kernel density estimation on $500$ samples.
  }
  \label{fig:toggle_switch_fits}
\end{figure}

We now consider a `toggle switch' model, which is a generative model with unknown likelihood and hence suitable for our composite \gls{mmd} tests. The model describes how the expression level of two genes $u$ and $v$ interact in a sample of cells  \citep{bonassiBayesianLearningMarginal,bonassiSequentialMonteCarlo}. Sampling from the model involves two coupled discretised-time equations.
Denote by $\Pts[\theta, T]$ a model with parameters $\theta = (\alpha_1, \alpha_2, \beta_1, \beta_2, \mu, \sigma, \gamma)$ with a discretisation consisting of $T$ steps.
To sample $y \sim \Pts[\theta, T]$, we sample
$y \sim \TruncNormal(\mu + u_T,  \mu \sigma/u_T^\gamma)$,
where
\begin{align*}
  \qquad   u_{t+1} & \sim \TruncNormal(\mu_{u,t}, 0.5),
  \qquad \mu_{u,t} = u_t + \frac{\alpha_1}{\left(1 + v_t ^ {\beta_1}\right)} - (1 + 0.03 u_t), \\
  \qquad v_{t+1}   & \sim \TruncNormal(\mu_{v,t}, 0.5),
  \qquad \mu_{v,t} = v_t + \frac{\alpha_2}{\left(1 + u_t ^ {\beta_2}\right)} - (1 + 0.03 v_t),
\end{align*}
and $u_0 = v_0 = 10$.
In the above $\TruncNormal$ indicates a Gaussian distribution truncated to give non-negative realisations.
\Cref{fig:toggle_switch_time_evo} demonstrates how this sampling process works, showing the evolution of $u_t$ as $t$ increases.
To fit the model, we use stochastic optimisation; see \Cref{app:toggle_switch_details}.

Existing work considers the model to have converged for $T = 300$, but using such a large value can be computationally expensive, thus a practitioner may wonder if it is possible to use a smaller value.
We apply our parametric bootstrap test to this scenario by generating $n=500$ observations from $\Pts[\theta_0, 300]$, and then testing the $\nullc: \exists \tilde{\theta}_0 \text{ such that } \Pts[\tilde{\theta}_0, T] = \Pts[\theta_0, 300]$ for values of $1 \leq T \leq 50$.
To generate the observations we follow \citet{berntonApproximateBayesianComputation} and use $\theta_0 = (22, 12, 4, 4.5, 325, 0.25, 0.15)$.
\Cref{fig:toggle_switch_increasing_T} shows the result of the test, revealing that the test is unlikely to reject $\nullc$ for $T \geq 40$.
In \Cref{fig:toggle_switch_fits} we give examples of the fit of the model for small and large $T$.

\subsection{Density Estimation with a Kernel Exponential Family Model}
\begin{figure}
  \centering
  \includegraphics{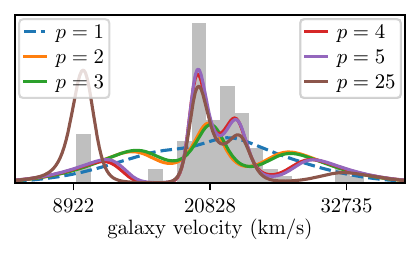}
  \caption{
    Fit of the kernel exponential family model on the galaxies data set.
    The grey histogram shows the data set, while the coloured lines show the density of the fitted model with varying $p$.
    The dashed lines indicate that the wild bootstrap test rejected $\nullc$, while the solid lines indicate that it failed to reject.
  }
  \label{fig:kvf_galaxies}
\end{figure}
We now consider the kernel exponential family $p_\theta(x) \propto \qkef(x) \exp(f(x))$ where $f$ is an element of some RKHS $\H_\kappa$ with Gaussian kernel $\kappa : \X \times \X \rightarrow \R$  with lengthscale $\lkef$, and $\qkef$ is a reference density \citep{canuKernelMethodsExponential}.
Following \citet{Matsubara2021}, we will work with a finite-rank approximation
$f(x) = \sum_{i=1}^p \theta_i \phi_i(x)$ where $\theta \in \Theta = \R^p$ and
\begin{equation*}
  \phi_i = \sqrt{\frac{2^i x^{2i}}{\lkef^{2i} i!}} \exp \left( - \frac{x^2}{\lkef^2} \right), \text{ for all } i \eqcom
\end{equation*}
are basis functions.
\citet{Matsubara2021} use this model for a density estimation task on a data set of velocities of $82$ galaxies \citep{postmanProbesLargescaleStructure, roederDensityEstimationConfidence}, and use the approximation with $p=25$ for computational convenience. However, an open question is whether $p=25$ is sufficient to achieve a good fit to the data.

We propose to use our composite \gls{ksd} test to answer this question.
\Cref{fig:kvf_galaxies} shows the fit of the model for increasing values of $p$, and whether the test rejected $\nullc$.
We use a sum of IMQ kernels; see \Cref{app:kernels} for a discussion of this choice.
We find that the test rejects $\nullc$ for $p=1,2,3$, but does not reject for $p=4,5,25$, with the wild and parametric bootstraps producing similar results.
This suggests that $p=25$ is a suitable choice, though it would also be reasonable to use a smaller value of $p$ which would further decrease the computational cost of inference.

\section{Limitations}
\label{sec:limitations}

While we find that our composite tests allow us to answer much more complex questions than existing kernel tests, we did also observe some limitations that we discuss below.

The performance of any kernel-based test depends heavily on the choice of kernel, as this determines the ability of the discrepancy to distinguish between distributions when observations are finite.
In our composite tests, this issue has a larger effect on the power of the test since we use the discrepancy not only as a test statistic but also for parameter estimation.
For example, in Appendix B, we demonstrate how a very poor choice of kernel can result in biased estimates of the parameter and thus a type I error rate larger than the level.
In practice, we should therefore consider our method as a test of both the model and the performance of the estimator, rather than of the model alone.
We consider this to be a desirable behaviour, since, in practice, a parametric model is only ever as useful as our ability to estimate its parameters.

In the non-composite case, several techniques have been developed to address kernel selection,
including the power-maximizing kernel \citep{sutherlandGenerativeModelsModel},
Sup-MMD \citep{fukumizuKernelChoiceClassifiability},
and the aggregated procedure of \citet{schrabMMDAggregatedTwoSample,schrabKSDAggregatedGoodnessoffit}.
Future work could investigate how these procedures could be adapted for composite tests.
As an initial trial, in Appendix C we apply the aggregated procedure to our composite KSD test, finding that it sometimes achieves a small increase in power, and sometimes a small decrease.

Future work could also consider kernels which are specialised to structured data such as time-series, graphs, and images.

A further issue that can arise is the numerical optimiser implementing the estimator failing to converge to a global optimum (for example, because the objective is non-convex and the optimiser only finds local minimums).
This failure mode is also illustrated in \cref{app:limitations}.
Once again, this means we should consider our method as a test of the fitting method, in addition to the model itself.

All tests based on the KSD can suffer from type I errors due to its inability to distinguish between multi-modal distributions that only differ by the weight of each mode, and this limitation also applies to our test.
The limitation can be observed in \cref{fig:kvf_galaxies} where, for $p=25$, the model allocates too much weight to the left mode, yet the test still fails to reject $\nullc$.
However, as $n \to \infty$ this error will no longer occur.

A final limitation of the tests is due to the difficulty of approximating $c_\alpha$ for models which are expensive to fit due to a large number of parameters and/or the size of the associated data sets. In those cases, the wild bootstrap might not provide a good enough approximation of $c_\alpha$, whilst the high computational cost of the parametric bootstrap (which requires fitting the model many times) may be prohibitive. Alternative bootstrap algorithms could be developed for this task; see for example the work of \citet{Kojadinovic2012}.

\section{Connections with Existing Tests} \label{sec:connections}

Connections between (non-composite) kernel-based and classical tests are well-studied; see \citet{Sejdinovic2013} for an extensive discussion showing that MMD tests are equivalent to tests with the energy distance when using an appropriate kernel.
Naturally, similar results also hold for composite tests, and we sketch out some of these connections below.

\subsection{Likelihood-ratio Tests}

The first connection is with likelihood-ratio tests \citep[Section 12.4.4]{lehmann-testing-statistical-hypotheses}, which compare a null model $p_{\theta_0}$ and an alternative parametric model $q_{\gamma}$. Interestingly, we will show in this section that our tests are intimately linked to likelihood ratio tests where the alternative model is parameterised in some reproducing kernel Hilbert space.

Likelihood-ratio are based on the test statistic $S_n(\gamma;f) =\frac{\partial}{\partial \gamma} \frac{1}{n}\sum_{i=1}^n \log q_{\gamma}(x_i)$  which is optimised over $\gamma$ via the \gls{mle}. Now consider the alternative model given by $q_{f,\gamma}(x) \propto \exp(\gamma f(x)) p_{\theta_0}(x)$ for some $\gamma\in\mathbb{R}$.
This model is indexed by the function $f$, and consists of a perturbation of $p_{\theta_0}$ of size $\gamma$ in the sense that $q_{f,\gamma} = p_{\theta_0}$ when $\gamma = 0$. In this case,
\begin{align*}
  S_n(\gamma;f) & = \frac{1}{n} \sum_{i=1}^n f(x_i)-\frac{\int_{\X} f(x) \exp(\gamma f(x)) p_{\theta_0}(x)dx}{\int_{\X} \exp(\gamma f(x))p_{\theta_0}(x)dx},
\end{align*}
and
$S_n(0;f) = \frac{1}{n} \sum_{i=1}^n f(x_i) - \int_{\X} f(x)p_{\theta_0}(x)dx$. If $\gamma=0$, the \gls{mle} of $\gamma$ will be close to zero, and thus  $S_n(0)\approx 0$. Note that $|S_n(\gamma;f)| > 0$ implies that for some value $\gamma>0$, the perturbation model is a better fit for $\Q_n$ than $\P_{\theta_0}$. This choice of alternative model allows us to recover the \gls{mmd} and \gls{ksd} by considering suitably defined perturbations in some \gls{rkhs}:
\begin{align*}
  \sup_{f \in \F_{\text{MMD}}}S_n(0;f)=\text{MMD}(\P_{\theta_0},\Q_n), \qquad
  \sup_{f \in \F_{\text{KSD}}}S_n(0;f)=\text{KSD}(\P_{\theta_0},\Q_n).
\end{align*}
where $\F_{\text{MMD}}$ and $\F_{\text{KSD}}$ were defined in \Cref{sec:background}. Therefore, MMD and KSD tests are likelihood-ratio tests with an alternative model parametrised in some RKHS. Of course, the argument above also holds if $\thetaz$ is replaced by $\thetahn$. As a result, our composite tests can be thought of as a composite version of likelihood-ratio tests where $\thetahn$ is a minimum distance estimator based on the MMD and KSD.

\subsection{Tests Based on Characteristic Functions}

The composite goodness-of-fit tests in this paper are also closely connected with hypothesis tests based on distances between characteristic functions. Recall that the characteristic function of some distribution $\Q$ is defined as $\phi_{\Q}(\omega) = \E_{X \sim \Q} [\exp(i X^\top \omega)]$. The following results shows that using the \gls{mmd} is equivalent to comparing the (weighted) $L^2$ distance between characteristic functions, where the weight depends on the choice of kernel.
\begin{theorem}[Theorem 3 and Corollary 4, \citet{Sriperumbudur2009}]\label{thm:Sriperumbudur}
  Let $\X = \R^d$ and suppose $K(x,y)=\psi(x-y)$ where $\psi:\R^d \rightarrow \R$ is a bounded and continuous positive function. Following Bochner's theorem, $\psi$ is the Fourier transform of a finite nonnegative Borel measure $\Lambda$: $\psi(x) = \int_{\R^d} \exp(-i x^\top \omega) \Lambda(d\omega)$ and hence the \gls{mmd} can be expressed as
  \begin{align*}
    \MMD^2(\P,\Q) = \int_{\R^d} |\phi_{\P}(\omega) - \phi_{\Q}(\omega)|^2 \Lambda(d\omega).
  \end{align*}
\end{theorem}
There are a number of composite tests based on weighted distances between characteristic functions. One example is the work of \citet{Henze1990}, which consider testing for multivariate Gaussians with unknown mean and covariance. This test therefore uses a distance equivalent to the MMD with a specific choice of kernel. However, this paper makes use of maximum likelihood estimation rather than of a minimum MMD estimator. Interestingly, this means their test is equivalent to that in \citet{kellnerOnesampleTestNormality}, but the latter paper does not make the connection and propose to use the test by \citet{Henze1990} as a competitor in their numerical experiments. Note that this correspondence extends to the functional data case; see \citet{Wynne2021} for more details.

Another related work is that of \citet{Koutrouvelis1981}, who focus on (possibly non-Gaussian) univariate distributions and consider an approximation of the weighted $L^2$ distance between empirical characteristic functions. Here, the approximation consists of discretising the integral in the definition of $L^2$ distance, and this discrepancy can therefore be thought of as an approximation of the MMD with a specific choice of kernel. Their approach does however have an additional level of approximation due to the discretisation of the integral which is unnecessary when noting the connection with the MMD. Note that \citet{Koutrouvelis1981} used the same for both discrepancy for estimation and testing.

The connection between tests based on kernels and characteristic functions can also be extended to the case of the \gls{ksd}, albeit through modified characteristic functions. To state the result, define $\Sl_{\P,x} g = g(x) (\partial \log p(x)/\partial x_l) + (\partial g(x)/\partial x_l)$ for some sufficiently regular scalar-valued test function $g$ and $l \in \{1,\ldots,d\}$. For some fixed $l$, the modified characteristic function of some distribution $\Q$ will take the form
\begin{align*}
  \tilde{\phi}^l_{\Q}(\omega) = \E_{X\sim\Q} \left[\Sl_{\P,x}\left[ \exp(i X^\top \omega)\right]\right].
\end{align*}

\begin{theorem}\label{thm:KSD_characteristic}[Simplified version of Theorem 4.1, \citet{Wynne2022}]
  Let $\X = \R^d$ and suppose $K(x,y)=\psi(x-y)$ where $\psi:\R^d \rightarrow \R$ is a bounded and continuous positive function so that $\psi(x) = \int_{\R^d} \exp(-i x^\top \omega) \Lambda(d\omega)$. Then, the KSD can be expressed as
  \begin{align*}
    \KSD^2(\P,\Q) =  \int_{\R^d} \sum_{l=1}^d   \left| \tilde{\phi}^l_{\Q}(\omega) \right|^2 \Lambda(d \omega) =  \int_{\R^d} \sum_{l=1}^d   \left| \tilde{\phi}^l_{\Q}(\omega)- \tilde{\phi}^l_{\P}(\omega) \right|^2 \Lambda(d \omega).
  \end{align*}
\end{theorem}
\citet{ebnerCombiningZeroBias} proposed a test of normality which uses a test statistic exactly of the form in \Cref{thm:KSD_characteristic}. Their test implicitly assumes that $\Lambda$ has a density, and they study the impact of the choice of density on the performance of the test. Through the theorem above, we can see that this is equivalent to varying the choice of kernel $K$ indexing the KSD.

\section{Conclusion}\label{sec:conclusions}

This paper proposes and studies two kernel-based tests to verify whether a given data set is a realisation from any element of some parametric family of distributions. This was achieved by combining existing kernel hypothesis tests with recently developed minimum distance estimators, using either the MMD or KSD. We also studied two bootstrap algorithms to implement these tests: a wild bootstrap algorithm which is suitable in large-data regimes, and a parametric bootstrap which is suitable in smaller data regimes.

A number of limitations were mentioned in \Cref{sec:limitations}, and these could each be addressed in future work. For example, the issue that the tests perform poorly when the minimum distance estimators provide a poor parameter estimate could be mitigated by allowing for the use of alternative estimators which are specialised for models at hand. This is straightforward to do in practice, but would require an extension of our theoretical results. Relatedly, the performance of the composite tests is dependent on the choice of an appropriate divergence, usually through a choice of kernel. This issue could be mitigated by using the aggregate approach of \citet{schrabMMDAggregatedTwoSample,schrabKSDAggregatedGoodnessoffit}, which removes the need to carefully select a kernel, within our composite tests.

Additionally, our paper focused on Euclidean data, our MMD and KSD tests could straightforwardly be generalised to more complex settings. For example, the cases of discrete  \citep{Yang2018}, manifold \citep{Xu2020directional}, censored \citep{Fernandez2019} or time-to-event \citep{fernandezKernelizedSteinDiscrepancy} data could all be covered through our general methodology. Furthermore, models such as point processes  \citep{Yang2019}, latent variable models \citep{Kanagawa2019} or exponential random graph models \citep{Xu2021} could also be tackled in this way.

Finally, we also envisage that our approach could be extended to more complex testing questions. For example, the framework could be extended to relative tests, where instead of testing whether data comes from a fixed parametric model, the relative fit of several parametric models would be compared \citep{Bounliphone2015,Jitkrittum2018,Lim2019}. Alternatively, we could extend our methodology to construct composite tests for conditional distributions \citep{jitkrittumTestingGoodnessFit}.

\acks
We thank Antonin Schrab for his helpful instructions on implementing the aggregated version of test, as presented in \cref{app:aggregated_test}.

OK and FXB acknowledge support from the Engineering and Physical Sciences Research Council with grant numbers EP/S021566/1 and EP/Y011805/1 respectively.
TF was supported by the Data Observatory Foundation - ANID Technology Center No. DO210001 and ANID FONDECYT grant No. 11221143.
Arthur Gretton acknowledges support from the Gatsby Charitable Foundation.

\newpage
\appendix
{
  \begin{center}
    \LARGE
    \vspace{5mm}
    \textbf{Appendix}
    \vspace{5mm}
  \end{center}
}

\etocsettocdepth.toc{subsection}
\tableofcontents

\section{Theoretical Results for MMD}
\subsection{Definitions}
\subsubsection{General Notation}\label{subsec:gnotation}
In this section we make some additional definitions which will be used in the proofs. We denote by $\Q_n$ the empirical measure based on the sample $X_1,\ldots,X_n\overset{i.i.d.}{\sim}\Q$, and we denote by  $\Q_n^\star$
the empirical measure based on the parametric bootstrap samples $X_1^\star,\ldots, X_n^\star\overset{i.i.d}{\sim}\P_{\thetahn}$ given $\thetahn$.

\begin{enumerate}
  \item We define the functions $L_n:\Theta\to\Reals$ and $L:\Theta\to\Reals$ by
        \begin{align}
        L_n(\theta)=\MMDk^2(\P_\theta,\Q_n)\qquad\text{and}\qquad L(\theta)=\MMDk^2(\P_\theta,\Q).\label{Def:L_n}
        \end{align}
        Similarly, for the parametric Bootstrap, we define $L_n^\star:\Theta\to\Reals$ by
        \begin{align}
          L_n^\star(\theta)=\MMDk^2(\P_\theta,\Q_n^\star). \label{Def:L_nstar}
        \end{align}
        Note then that $\thetahn=\argmin_{\theta\in\Theta} L_n(\theta)$ and $\theta_n^\star=\argmin_{\theta\in\Theta} L_n^\star(\theta)$.

  \item We denote by $\Hb\in\Reals^{p\times p}$ the Hessian matrix given by $\mathbf{H}_{ij}=\frac{\partial^2}{\partial\theta_i\partial\theta_j}L(\theta_0)$ for any $i,j\in \{1,\ldots,p\}$.

  \item We define the function $g:\mathcal{H}\times\Theta\to\Reals$ by $\g{\theta}=\E_{X\sim \P_\theta}(\omega(X))$.

  \item We denote by $\mu_\theta,\mu_n$ and $\mu_n^\star\in\mathcal{H}$, respectively, the kernel mean embeddings of the distribution function $\P_\theta$ and the empirical distribution functions $\Q_n$ and $\Q_n^\star$, given by
        \begin{equation*}
          \mu_{\theta}(x)=\int_{\X} K(x,y)p_{\theta}(y)\lambda(dy),\quad \mu_n(x)=\frac{1}{n}\sum_{i=1}^n K(x,X_i), \quad
          \mu_n^\star(x)=\frac{1}{n}\sum_{i=1}^n K(x,X_i^\star).
        \end{equation*}

  \item We define $\phit:\X\times\Theta\to\Reals^p$ by
        \begin{align}
          \phit(x,\theta) & =\nabla_\theta \mu_\theta(x)-\int_\X (\nabla_\theta \mu_\theta(y))p_\theta(y)\lambda(dy),\label{eqn:phi}
        \end{align}
        where $\nabla_\theta \mu_\theta:\mathcal{X} \rightarrow \mathbb{R}^p$ is such that $(\nabla_\theta \mu_\theta (x))_i = \frac{\partial}{\partial \theta_i} \mu_\theta(x)$ for $i \in \{1,\ldots,p\}$.

  \item We define by $\eta:\mathcal{H}\times \X\times\Theta\to\Reals$  by
        \begin{align}
          \eta(\omega,x,\theta)={\omega}(x)-\g{\theta}-2\left\langle\nabla_{\theta}\g{\theta_0},\Hb^{-1}\phit(x,\theta)\right\rangle,\label{def:eta1}
        \end{align}
        where $\nabla_\theta g:\mathcal{H} \times \Theta \rightarrow \mathbb{R}^p$ is such that $(\nabla_\theta g(\omega,\theta))_i = \frac{\partial}{\partial \theta_i} g(\omega,\theta)$ for $i \in \{1,\ldots,p\}$.

  \item We define the functionals $S_n:\mathcal{H}\to\Reals$ and $S_n^\star:\mathcal{H}\to\Reals$ by
        \begin{align}
          S_n(\omega) & =\frac{1}{\sqrt{n}}\sum_{i=1}^n\eta(\omega,X_i,\theta_0),\qquad\text{and}\qquad
          S_n^\star(\omega)=\frac{1}{\sqrt{n}}\sum_{i=1}^n\eta(\omega,X_i^\star,\thetahn),\label{eqn:Snfunction}
        \end{align}
        where $\eta:\mathcal{H}\times\X\times\Theta\to\Reals$ is defined in \cref{def:eta1}.
\end{enumerate}

\begin{remark}[\textit{Bound for RKHS functions}]\label{remakH}
Observe that for any $\omega\in\mathcal{H}$ and $x\in\X$,
\begin{align*}
|\omega(x)|=|\InerH{\omega}{K_x}|\leq \LH{\omega}\LH{K_x}=\LH{\omega}\sqrt{K(x,x)},
\end{align*}
where the inequality is due to the Cauchy-Schwarz's inequality. By \Cref{\assumptions}, we have that the kernel is bounded by some constant $C>0$. Thus $\sup_{x\in\X}|\omega(x)|\leq \LH{\omega}\sqrt{C}$.
\end{remark}

\begin{remark}[\textit{Interchange of integration and differentiation}]\label{remark:lebesgue}
\;
Observe that for any fixed $\omega\in\mathcal{H}$, $g(\omega,\cdot)\in\mathcal{C}^3(\Theta)$ and, moreover
  \begin{align*}
  \mathcal{D}_\theta g(\omega,\theta)=\int_{\X} \omega(x)\mathcal{D}_\theta p_\theta(x)\lambda(dx),
  \end{align*}
  where $\mathcal{D}_\theta$ is any derivative up to order three. The previous result holds by an application of Lebesgue's dominated convergence theorem since  $p_\theta(x)\in\mathcal{C}^3(\Theta)$ for all $x\in\X$, and by the integrability conditions of \Cref{condition:model}. Similarly $\mu_\theta(x)\in\mathcal{C}^3(\Theta)$ for all $x\in\X$, and $\mathcal{D}_\theta \mu_\theta(x)=\int_{\mathcal{X}} K(x,y)\mathcal{D}_\theta p_\theta(y)\lambda(dy)$.

Moreover, by the integrability conditions of \Cref{condition:model} and since the kernel is bounded by a constant $C>0$ (\Cref{condition:mmd_kernel}), it holds
 \begin{align}
     \sup_{\theta\in\Theta}\sup_{x\in\X}\left|\int_\X K(x,y)\mathcal{D}_\theta p_\theta(y)\lambda(dy)\right|\leq  C\int_\X \sup_{\theta\in\Theta}\left|\mathcal{D}_\theta p_\theta(y)\right|\lambda(dy)<\infty.\label{eqn:remarkboundedmu}
 \end{align}

\end{remark}

\subsubsection{Test Statistic}
In our proofs we consider an alternative form of the definition of the MMD, which is equivalent to that in \Cref{eq:mmd_exact}:
\begin{equation*}
  n\MMDk^2(\P_\theta,\Q_n) = \sup_{\unitball} \left(
    \frac{1}{\sqrt{n}} \sum_{i=1}^n \omega(X_i) - g(\omega, \theta)
  \right)^2 .
\end{equation*}

\subsubsection{Wild Bootstrap}
In order to analyse the asymptotic behaviour of the wild bootstrap test-statistic, we write it in terms of a supremum as follows
\begin{align*}
     n \MMD^2_W(\P_{\thetahn,n}, \Q_n)
   & \defeq \frac{1}{n} \sum_{i,j=1}^n W_i W_j
  h_{\MMD}((X_i,\tilde{X}_i), (X_j,\tilde{X}_j))\\
  &=\sup_{\unitball} \left(\frac{1}{\sqrt{n}}\sum_{i=1}^n W_i(\omega(X_i)-\omega(\tilde{X}_i))\right)^2,
\end{align*}
where recall that $X_1,\ldots,X_n\overset{i.i.d.}{\sim}Q$, $\tilde X_1,\ldots,\tilde X_n\overset{i.i.d.}{\sim}P_{\thetahn}$ given $\thetahn$, and $W_1,\ldots,W_n\overset{i.i.d.}{\sim}$ Rademacher are independent of everything.

\subsubsection{Parametric bootstrap} \label{app:parametric_bootstrap_definitions}
To obtain the results for the parametric bootstrap, we condition on the data $X_1, \ldots, X_n$.
To this end, we use the following notation:
\begin{itemize}
  \item Define
  \begin{align*}
    &\Prob_D(\cdot) = \Prob(\cdot|X_1,\ldots,X_n),
    \quad \E_D(\cdot) = \E(\cdot|X_1,\ldots,X_n),  \\
    &\Var_D(\cdot) = \Var(\cdot | X_1,\ldots,X_n), \text{ and }
    \Cov_D(\cdot,\cdot) = \Cov(\cdot,\cdot|X_1,\ldots,X_n).
  \end{align*}
  \item Define $a_n=o_{p_D}(1)$ if, for each $\varepsilon>0$, it holds $\Prob_D(|a_n|\geq \varepsilon)\overset{\Prob}{\to} 0$ when $n \to \infty$.

  \item Define $a_n=O_{p_D}(1)$ if for any $\varepsilon>0$ there exists $M>0$ such that
  \begin{equation*}
    \Prob\left(\{\Prob_D(|a_n|>M)\leq \varepsilon\}\right) \to 1.
  \end{equation*}

  \item Given a random variable $a$, define $a_n\overset{\dist_D}{\to}a$ (i.e. $a_n$ converges in distribution to $a$ given the data points $X_1,\ldots,X_n$) if and only if $|\E_D(f(a_n)-f(a))| \overset{\Prob}{\to} 0$, for any bounded, uniformly continuous real-valued $f$.
\end{itemize}

\subsection{Preliminary Results}\label{sec:preliminaries}
We state some preliminary results that will be used in our proofs. The first corresponds to Theorem 1 of \citet{GFKT}, and it will be used to prove \Cref{Thm:2,Thm:2B}.
\begin{theorem}\label{thm:suplinear}
Let $(S_n)_{n\geq 1}$ be a sequence of bounded linear test-statistics satisfying conditions $G_0$-$G_2$ (stated below). Then
\begin{align*}
  \sup_{\unitball}S_n^2(\omega) \todist \sum_{i=1}^\infty \lambda_i Z_i^2,
\end{align*}
where $Z_1,Z_2,\ldots$ are i.i.d. standard normal random variables, and $\lambda_1,\lambda_2,\ldots$ are the eigenvalues of the operator $T_\sigma:\mathcal{H}\to\mathcal{H}$ defined by $(T_\sigma\omega)(x)=\sigma(\omega,K_x)$, where  $\sigma:\mathcal{H}\times\mathcal{H}\to\Reals$ is the bilinear form of Condition $G_0$.
\end{theorem}

We state conditions $G_0$-$G_2$.
\begin{description}
\item[Condition $G_0$] \textit{There exists a continuous bilinear form $\sigma:\mathcal{H}\times\mathcal{H}\to\Reals$ such that for any $m\in\mathbb{N}$, the bounded linear test-statistic $S_n(w_1+\cdots+w_m)$ converges in distribution to a normal random variable with
mean 0 and variance given by $\sum_{i=1}^m\sum_{j=1}^m\sigma(w_i,w_j)$.}
\item[Condition $G_1$] \textit{For some orthonormal basis $(\psi_i)_{i\geq1}$ of $\mathcal{H}$ we have $\sum_{i\geq 1}\sigma(\psi_i,\psi_i)<\infty$.}
\item[Condition $G_2$] \textit{For some orthonormal basis $(\psi_i)_{i\geq1}$ of $\mathcal{H}$, and for any $\varepsilon>0$, we have that
\begin{align*}
    \lim_{i\to\infty}\limsup_{n\to\infty}\Prob\left(\sum_{k=i+1}^\infty S_n^2(\psi_k)\geq\varepsilon\right)=0.
\end{align*}
}
\end{description}

The following result appears in \cite{nickl2012statistical}.
\begin{theorem}[Theorem 1, \cite{nickl2012statistical}]\label{Thm:preliminary1}
Suppose that $\Theta\subset \Reals^p$ is compact (i.e. bounded and closed). Assume that $L:\Theta\to\Reals$ is a deterministic function with $L\in\mathcal{C}^0(\Theta)$, and that $\theta_{opt}$ is the unique minimiser of $L$. If
\begin{align*}
    \sup_{\theta\in\Theta}|L_n(\theta)-L(\theta)|\overset{\Prob}{\to}0
\end{align*}
as $n$ grows to infinity, then any solution $\thetahn$ of $\min_{\theta\in\Theta}L_n(\theta)$ satisfies
\begin{align*}
    \thetahn\overset{\Prob}{\to}\theta_{opt}\quad\text{as}\quad n\to\infty.
\end{align*}
\end{theorem}

\subsection{Auxiliary Results} \label{appendix:aux_results}
The following collection of lemmas and propositions will be used in the proofs of our main results. Their proofs are provided in \Cref{sec:proofauxiliar}.

\begin{lemma}\label{lemma:repre}
Under \Cref{\assumptions} and $\nullc$ it holds that
\begin{align*}
n\MMDk^2(\P_{\thetahn},\Q_n) = n\MMDk^2(\P_{\theta_0},\Q_n)-\frac{2}{n}\sum_{i=1}^n\sum_{j=1}^n\phit(X_j,\theta_0)^\intercal\Hb^{-1}\phit(X_i,\theta_0)+o_p(1),
\end{align*}
where $\phit:\X\times\Theta\to\Reals^p$ is defined in \cref{eqn:phi}, and $\Hb$ is the Hessian matrix defined in \cref{subsec:gnotation}.
\end{lemma}

\begin{proposition}[Normal distribution for the Wild Bootstrap statistic]\label{prop:normalWBOscar}
\quad \\
Define $Z_n:\H\to\R$ by
\begin{equation*}
  Z_n(\omega)=\frac{1}{\sqrt{n}}\sum_{i=1}^n W_i(\omega(X_i)-\omega(\tilde{X}_i)),
\end{equation*}
where $X_1,\ldots,X_n\overset{i.i.d.}{\sim}Q$ and $\tilde X_1,\ldots,\tilde X_n\overset{i.i.d.}{\sim}P_{\thetahn}$ given $\thetahn$, and $W_1,\ldots,W_n\overset{i.i.d.}{\sim}\text{Rademacher}$ that are independent of everything else.

Then, $Z_n(\omega)\overset{\dist_D}{\to} N(0,\nu(\omega,\omega))$, where
\begin{align}
    \nu(\omega,\omega')=2\int (\omega(x)-g(\omega,\theta_0))(\omega'(x)-g(\omega',\theta_0)) p_{\theta_0}(x)\lambda(dx).\label{eqn:nufunction}
\end{align}
\end{proposition}

\begin{proposition}\label{Prop:convergence}
Under \Cref{\assumptions} and the null hypothesis it holds that $\thetahn\overset{\Prob}{\to}\theta_0$.
\end{proposition}

\begin{proposition}\label{prop:Normal}
\begin{align*}
 \nabla_{\theta}L_n(\theta)=-\frac{2}{n}\sum_{i=1}^n\phit(X_i,\theta),\qquad \text{and}\qquad  \nabla_{\theta}L_n^\star(\theta)=-\frac{2}{n}\sum_{i=1}^n\phit(X_i^\star,\theta).
\end{align*}
Moreover, under \Cref{\assumptions} the following items hold.
\begin{enumerate}
    \item[i)]  $\E_{X\sim \P_{\theta}}(\phit(X,\theta))=\mathbf{0}$, and $\E_{D}(\phit( X^\star,\thetahn))=\mathbf{0}$, where $X^\star\sim \P_{\thetahn}$ given $\thetahn$,
    \item[ii)]$\sup_{\theta\in\Theta}\sup_{x\in\Reals^d}\|\phit(x,\theta)\|<\infty$, and
    \item[iii)] $\sqrt{n}\|\nabla_{\theta}L_n(\theta_0)\|=O_p(1)$ holds under the null hypothesis, and $\sqrt{n}\|\nabla_{\theta}L_n^\star(\thetahn)\|=O_p(1)$.
\end{enumerate}
\end{proposition}

\begin{proposition}\label{prop:eta}
Under \Cref{\assumptions} the following items hold.
\begin{itemize}
    \item[i)]  For any fixed $\theta\in\Theta$ and $\omega\in\mathcal{H}$, $\E_{X\sim \P_{\theta}}(\eta(\omega,X,\theta))=0$,
    \item[ii)]  there exists constants $C_1>0$ and $C_2>0$ such that for any fixed $\theta\in\Theta$ and $\omega\in\mathcal{H}$, $$\E_{X\sim \P_{\theta}}(\eta^2(\omega,X,\theta))\leq C_1\E_{X\sim \P_\theta}\left({\omega}^2(X)\right)+C_2\left\|\nabla_{\theta}\g{\theta_0}\right\|^2,$$
    \item[iii)] there exists constants $C_1>0$ and $C_2>0$ that do not depend on $\omega$, $x$ nor $\theta$, such that for all $\omega\in\mathcal{H}$, it holds $\sup_{\theta\in\Theta}\sup_{x\in\X}|\eta(\omega,x,\theta)|< C_1\LH{\omega}+  C_2\LH{\omega}^2 <\infty$.
\end{itemize}

\end{proposition}

\begin{lemma}\label{prop:uniformconvergence}
 Under \Cref{\assumptions} and the null hypothesis, it holds that for any $\ell,j\in \{1,\ldots,p\}$ and $\varepsilon>0$
\begin{align}
   \sup_{\theta\in\Theta}\left|\frac{1}{n}\sum_{i=1}^n\left(\frac{\partial^2}{\partial\theta_\ell\partial\theta_j}\mu_{\theta}(X_i)-\E_{X\sim \P_{\theta_0}}\left(\frac{\partial^2}{\partial\theta_\ell\partial\theta_j}\mu_{\theta}(X)\right)\right)\right|\overset{\Prob}{\to}0,\label{eqn:unifconv1}
\end{align}
and,
\begin{align}
   \Prob_D\left(\sup_{\theta\in\Theta}\left|\frac{1}{n}\sum_{i=1}^n\left(\frac{\partial^2}{\partial\theta_\ell\partial\theta_j}\mu_{\theta}(X_i^\star)-\E_{X\sim \P_{\theta_0}}\left(\frac{\partial^2}{\partial\theta_\ell\partial\theta_j}\mu_{\theta}(X)\right)\right)\right|\geq \varepsilon\right)\overset{\Prob}{\to}0,\label{eqn:unifconv2}
\end{align}
as $n$ grows to infinity.

\end{lemma}
\begin{proposition}\label{prop:convergence of Hessian}
Define $\mathbf{H}_n$ and $\mathbf{H}_n^\star$ by
\begin{align*}
 (\mathbf{H}_n)_{ij}=\frac{\partial^2}{\partial\theta_i\partial\theta_j}L_n(\tilde{\theta}^j)\qquad\text{and}\qquad (\mathbf{H}_n^\star)_{ij}=\frac{\partial^2}{\partial\theta_i\partial\theta_j}L_n^\star(\tilde{\theta}^j_\star)
\end{align*}
for any $i,j\in \{1,\ldots,p\}$, where $L_n$ and $L_n^\star$ are  defined in \cref{Def:L_n,Def:L_nstar}, respectively, $\tilde\theta^1,\ldots,\tilde\theta^p$ lie between $\theta_0$ and $\thetahn$, and $\tilde\theta^1_\star,\ldots,\tilde\theta^p_\star$ lie between $\thetahn$ and $\thetahn^\star$.

Then, under \Cref{\assumptions}  and the null hypothesis it holds
\begin{align*}
\|\mathbf{H}_n-\Hb\|=o_p(1)\qquad\text{and}\qquad  \|\mathbf{H}_n^\star -\mathbf{H}\|=o_{p_D}(1)
\end{align*}
as $n$ grows to infinity.
\end{proposition}

\begin{proposition}\label{prop:normallimittheta} Under \Cref{\assumptions} and the null hypothesis, the following holds.
\begin{itemize}
    \item[i)] $\sqrt{n}(\theta_0-\thetahn)=\sqrt{n}\Hb^{-1} \nabla_\theta L_n(\theta_0)+o_p(1),$ and
    \item[ii)] $\sqrt{n}(\thetahn-\theta_n^\star)=\sqrt{n}\Hb^{-1} \nabla_\theta L_n^\star(\thetahn)+o_p(1).$
\end{itemize}
\end{proposition}

\begin{lemma}\label{Lemma:convergencePB1}  Under \Cref{\assumptions}, it holds
\begin{itemize}
    \item[i)] $\sqrt{n}(\thetahn-\theta_n^\star)=O_p(1)$, and
    \item[ii)] $\theta_n^\star-\theta_0=o_p(1)$ holds under the null hypothesis.
\end{itemize}
\end{lemma}

\begin{proposition}\label{Lemma:1}
Under the null hypothesis and \Cref{\assumptions}, we have that
\begin{equation*}
  \frac{1}{\sqrt{n}}\sum_{i=1}^n\omega(X_i)-g(\omega,\thetahn)=S_n(\omega)+o_p(1),
\end{equation*}
holds for every $\omega\in\mathcal{H}$, where $S_n:\mathcal{H}\to\Reals$ is defined in \cref{eqn:Snfunction} and the error term does not depend on $\omega$.
\end{proposition}

\begin{proposition}\label{Prop:Normal1} Consider $\Sb_n=(S_n(\omega_1),\ldots,S_n(\omega_m))$, where $\omega_\ell\in\mathcal{H}$ for each $\ell\in[m]$ and fixed $m\in\mathbb{N}$.  Then, under the null hypothesis, and \Cref{\assumptions}, we have $\Sb_n \todist N_m(0,\boldsymbol{\Sigma})$ as $n$ grows to infinity, where for any $i,j\in[m]$ we have
\begin{align}
\boldsymbol{\Sigma}_{ij}=\sigma(\omega_i,\omega_j)&:=%
\E_{X\sim \P_{\theta_0}}\left(\eta(\omega_i,X,\theta_0)\eta(\omega_j,X,\theta_0)\right),\label{eq:bilinear}
\end{align}
and the function $\eta$ is defined in \cref{def:eta1}.
\end{proposition}

The next two results are analogous to \Cref{Lemma:1} and \Cref{Prop:Normal1} but for the parametric bootstrap test-statistic.
Note that these results are obtained conditioned on the data $X_1,\ldots,X_n$.

\begin{proposition}\label{Parametric approx}
Under the null hypothesis and \Cref{\assumptions}, it holds that
\begin{equation*}
  \frac{1}{\sqrt{n}}\sum_{i=1}^n\omega( X_i^\star)-g(\omega,\theta_n^\star)=S_n^\star(\omega)+o_{p}(1),
\end{equation*}
where $S_n^\star:\mathcal{H}\to\Reals$ is defined in \cref{eqn:Snfunction}.
\end{proposition}

\begin{proposition}\label{Prop:Normal2} Consider $\Sb_n^\star=(S_n^\star(\omega_1),\ldots,S_n^\star(\omega_m))$, where $\omega_\ell\in\mathcal{H}$ for each $\ell\in[m]$ and  $m\in\mathbb{N}$ is fixed.  Then, under the null hypothesis, and \Cref{\assumptions}, we have $\Sb_n^\star \todist N_m(0,\boldsymbol{\Sigma})$ as $n$ grows to infinity, where ${\boldsymbol\Sigma}$ is defined in \cref{eq:bilinear}
\end{proposition}

\begin{proposition}\label{prop:secondmomentbound}
Let $(\psi_i)_{i\geq 1}$ be any orthonormal basis of $\mathcal{H}$. Then under \Cref{\assumptions} it holds that $\sum_{i\geq 1}\|\nabla_{\theta}g(\psi_i,\theta_0)\|^2<\infty$.
\end{proposition}

\subsection{Proofs of Main Results}
This section contains proofs of results given in the main paper.

\begin{proof}{\Cref{Thm:2}}
By the definition of the maximum mean discrepancy, it holds
\begin{align*}
    n\MMDk^2(\P_{\thetahn},\Q_n)=\sup_{\unitball}\gamma_n^2(\omega),
    \quad\text{where}\quad \gamma_n(\omega)=\frac{1}{\sqrt{n}}\sum_{i=1}^n\omega(X_i)-g(\omega,\thetahn).
\end{align*}
\Cref{Lemma:1} yields $\gamma_n(\omega)=S_n(\omega)+o_p(1)$, where $S_n:\mathcal{H}\to\Reals$ is the functional defined in \cref{eqn:Snfunction}, and the error term does not depend on $\omega$. Then, by using the fact that $\sup_{\unitball}{S}^2_n(\omega)$ converges in distribution (which will be proved next) we obtain
\begin{align*}
n\MMDk^2(\P_{\thetahn},\Q_n)=\sup_{\unitball}S^2_n(\omega)+o_p(1).
\end{align*}

By Slutsky's theorem, we deduce that $n\MMDk^2(\P_{\thetahn},\Q_n)$ and $\sup_{\unitball}{S}_n^2(\omega)$ have the same asymptotic null distribution.

It is not difficult to verify that $S_n(\omega)=\frac{1}{\sqrt{n}}\sum_{i=1}^n\eta(\omega,X_i,\theta_0)$ is bounded and linear in $\mathcal{H}$. The linearity follows from the fact that for any fixed $x\in\X$ and $\theta\in\Theta$, $\eta(\omega,x,\theta)$ is linear on its argument $\omega$. Boundedness follows from item iii of \Cref{prop:eta} since $\sup_{\theta\in\Theta}\sup_{x\in\X}|\eta(\omega,x,\theta)|< C_1\LH{\omega}+  C_2\LH{\omega}^2$ where $C_1>0$ and $C_2>0$ are constants that do not depend on $\omega$.
Then, we can use \Cref{thm:suplinear} to study the asymptotic null distribution of $\sup_{\unitball}{S}_n(\omega)^2$ and thus obtain the result.
To apply \Cref{thm:suplinear} we need to verify  conditions $G_0, G_1$ and $G_2$, which are stated in \Cref{sec:preliminaries}.

Condition $G_0$ follows directly from \Cref{Prop:Normal1}, where $\sigma:\mathcal{H}\times\mathcal{H}\to\Reals$ is given by $\sigma(\omega,\omega'):=\E_{X\sim \P_{\theta_0}}(\eta(\omega,X,\theta_0)\eta(\omega',X,\theta_0))$ for any $\omega,\omega'\in\mathcal{H}$. To verify condition $G_1$ consider an orthonormal basis $(\psi_i)_{i\geq 1}$ of $\mathcal{H}$ and  observe that by \Cref{prop:eta}.ii. there exists constants $C_1>0$ and $C_2>0$ such that for all $i\geq 1$ it holds
\begin{align*}
    \sigma(\psi_i,\psi_i)=\E_{X\sim \P_{\theta_0}}(\eta^2(\psi_i,X,\theta_0))<C_1\E_{X\sim \P_{\theta_0}}\left({\psi_i}^2(X)\right)+C_2\left\|\nabla_{\theta}g(\psi_i,\theta_0)\right\|^2.
\end{align*}
 Then,
\begin{align}
    \sum_{i\geq 1}\sigma(\psi_i,\psi_i)\leq C_1\sum_{i\geq 1}\E_{X\sim \P_{\theta_0}}\left({\psi_i}^2(X)\right)+C_2\sum_{i\geq 1}\left\|\nabla_{\theta}g(\psi_i,\theta_0)\right\|^2<\infty.\label{eq:FiniteSigma}
\end{align}
where the last inequality holds since $K(x,x)=\sum_{i\geq 1}\psi_i^2(x)$, and the kernel is bounded by our assumptions, and by \Cref{prop:secondmomentbound}. Finally, we proceed to verify condition $G_2$. For this, let $\varepsilon>0$, and note
\begin{align*}
    \Prob\left(\sum_{k\geq i}{S}_n^2(\psi_k)\geq\varepsilon\right)\leq \sum_{k\geq i} \varepsilon^{-1}\E\left({S}_n^2(\psi_k)\right)
    &=\varepsilon^{-1}\sum_{k\geq i}\E_{X\sim \P_{\theta_0}}(\eta^2(\psi_k,X,\theta_0))\\
    &=\varepsilon^{-1}\sum_{k\geq i} \sigma(\psi_k,\psi_k),
\end{align*}
where first inequality is due to Markov's inequality, the subsequent equality follows from the definition of ${S}_n(\omega)=\frac{1}{\sqrt{n}}\sum_{i=1}^n\eta(\omega,X_i,\theta_0)$ given in \cref{eqn:Snfunction}, and the last equality is due to the definition of $\sigma$. Finally, notice that
\begin{align*}
  \lim_{i\to\infty}\limsup_{n\to\infty}\Prob\left(\sum_{k\geq i}{S}_n(\psi_k)^2\geq\varepsilon\right)\leq \varepsilon^{-1}\lim_{i\to\infty}\sum_{k\geq i}\sigma(\psi_k,\psi_k)=0,
\end{align*}
where  the limit is due to \cref{eq:FiniteSigma}.
\end{proof}

\begin{proof}{\Cref{Thm:alte}}
Observe that
\begin{align*}
 \operatorname{MMD}(\P_{\thetahn},\Q_n)=\LH{\mu_{\thetahn}-\mu_n}\geq \LH{\mu_{\thetahn}-\mu_{\Q}}-\LH{\mu_{\Q}-\mu_n},
\end{align*}
where $\mu_Q$ is the kernel mean embedding of $Q$ given by $\mu_Q(x)=\int K(x,y)q(y)\lambda(dy)$. Note that $\lim_{n\to\infty}\LH{\mu_{\Q}-\mu_n}=0$ a.s. by the standard law of large numbers for
RKHS in \citet[Theorem~1]{chen2011note}. Next, we will prove that $$\liminf_{n\to\infty}\LH{\mu_{\thetahn}-\mu_{\Q}}>0\quad  a.s.$$

We proceed by contradiction. Suppose that $\liminf_{n\to\infty}\LH{\mu_{\thetahn}-\mu_{\Q}}=0$ a.s., then there exists a collection of indices $(a(n))_{n\geq 1}$ such that the subsequence $\mu_{\widehat{\theta}_{a(n)}}$ satisfies
\begin{align}
\lim_{n\to\infty}\LH{\mu_{\widehat{\theta}_{a(n)}}-\mu_{\Q}}^2=0\quad a.s.\label{eq:alte1}
\end{align}
Additionally, note that since the set $\Theta$ is compact, there exists a subsequence $(\widehat{\theta}_{a(b(n))})_{n\geq 1}$ and $\theta^\star\in\Theta$  such that $\lim_{n\to\infty}\|\widehat{\theta}_{a(b(n))}-\theta^\star\|=0,  a.s.$ By \Cref{condition:model,condition:mmd_kernel}, the kernel $K$ is bounded,  $p_{\theta}(x)\in\mathcal{C}^3(\Theta)$ for each fixed $x\in\X$, and $\int_\X\sup_{\theta\in\Theta}p_{\theta}(x)\lambda(x)dx<\infty$.
Therefore, by the Lebesgue's dominated convergence theorem we deduce
\begin{align}
    &\LH{\mu_{\widehat\theta_{a(b(n))}}-\mu_{\theta^\star}}^2\nonumber\\
    &\qquad=\int_\X\int_\X K(x,y)\left(p_{\widehat\theta_{a(b(n))}}(x)-p_{\theta^\star}(x)\right)\left(p_{\widehat\theta_{a(b(n))}}(y)-p_{\theta^\star}(y)\right)\lambda(x)\lambda(y)dxdy\nonumber\\
    &\qquad\to0\label{eq:alte2}
\end{align}
when $n$ grows to infinity. Then, by the triangle inequality and \cref{eq:alte1,eq:alte2} we have
\begin{align*}
  \LH{\mu_{\theta^\star}-\mu_{\Q}}\leq\LH{\mu_{\theta^\star}-\mu_{\widehat\theta_{a(b(n))}}}+\LH{\mu_{\widehat\theta_{a(b(n))}}-\mu_{\Q}}\to0\quad a.s.
\end{align*}
Thus $\LH{\mu_{\theta^\star}-\mu_{\Q}}=\MMDk(\Q,\P_{\theta^\star})=0$, and since the kernel is characteristic by \Cref{condition:mmd_kernel}, we conclude $\Q=\P_{\theta^\star}$, but this is a contradiction since under the alternative hypothesis $\Q\not\in\mathcal{P}_{\theta}$.
\end{proof}

\begin{proof}{\Cref{thm:WBOscar}}
Observe that
\begin{align}
    n\MMD^2_W(\P_{\thetahn,n},\Q_n)=\sup_{\unitball}Z^2_n(\omega),\label{eqn:WZrepre}
\end{align}
where $Z_n:\H\to\R$ is defined $Z_n(\omega)=\frac{1}{\sqrt{n}}\sum_{i=1}^n W_i(\omega(X_i)-\omega(\tilde{X}_i))$. It is not difficult to see that $Z_n$ is linear on its argument $\omega$, and that $Z_n$ is bounded since \Cref{\assumptions} yield $\sup_{x\in\X} |\omega(x)|\leq \LH{\omega}\sup_{x\in\X}\sqrt{K(x,x)}<\infty$ . Then, the asymptotic distribution of the wild bootstrap test-statistic $n\MMD^2_W(\P_{\thetahn,n},\Q_n)$ can be obtained from \Cref{thm:suplinear} and \cref{eqn:WZrepre}, by  verifying that $Z_n$ satisfies conditions $G_0$-$G_2$, which we proceed to verify.

Observe that Condition $G_0$ follows directly from \Cref{prop:normalWBOscar}, where we obtain that
\begin{align*}
    Z_n(\omega_1+\ldots+\omega_m)\overset{\dist_D}{\to}N\left(0,\sum_{i=1}^m\sum_{j=1}^m\nu(\omega_i,\omega_j)\right),
\end{align*}
for any $m\in \N$ and $\omega_1,\ldots,\omega_m\in\H$, where $\nu:\H\times\H\to\R$ is given by
\begin{align*}
    \nu(\omega,\omega')=2\int(\omega(x)-g(\omega,\theta_0))(\omega'(x)-g(\omega',\theta_0))p_{\theta_0}(x)\lambda(dx).
\end{align*}
To verify Condition $G_1$, we consider some orthonormal basis  $(\psi_i)_{i\geq 1}$ of $\H$, and verify that  $\sum_{i\geq 1}\nu(\psi_i,\psi_i)<\infty$ holds. Note that
\begin{align*}
    \sum_{i\geq 1}\nu(\psi_i,\psi_i)&=\sum_{i\geq 1}2\int(\psi_i(x)-g(\psi_i,\theta_0))^2 p_{\theta_0}(x)\lambda(dx)\\
    &\leq 2\int\sum_{i\geq 1}\psi_i(x)^2p_{\theta_0}(x)\lambda(dx)=2\int K(x,x)p_{\theta_0}(x)\lambda(dx)<\infty,
\end{align*}
where the first inequality follows from the fact that the variance of a random variable is upper bounded by its second moment, and the last inequality holds from the fact that $\sum_{i\geq 1}\psi_i^2(x)=K(x,x)$, and that the kernel is bounded by \Cref{condition:mmd_kernel}.

Finally, note that $\lim_{i\to\infty}\limsup_{n\to\infty}\Prob_D\left(\sum_{k\geq i}Z_n^2(\psi_k)\geq \varepsilon\right)=0$ verifies Condition $G_2$.  By Markov's inequality
\begin{align*}
    \Prob_D\left(\sum_{k\geq i}Z_n^2(\psi_k)
\geq \varepsilon\right)
&\leq \frac{1}{\varepsilon}\sum_{k\geq i}\E_D(Z_n^2(\psi_k)),
\end{align*}
and note that
\begin{align*}
\frac{1}{\varepsilon}\sum_{k\geq i}\E_D(Z_n^2(\psi_k))&=\frac{1}{\varepsilon}\sum_{k\geq i}\E_D\left(\left(\frac{1}{\sqrt{n}}\sum_{j=1}^n W_j(\psi_k(X_j)-\psi_k(\tilde{X}_j))\right)^2\right)\\
&=\frac{1}{\varepsilon}\sum_{k\geq i}\E_D\left(\E\left(\left.\left(\frac{1}{\sqrt{n}}\sum_{j=1}^n W_j(\psi_k(X_j)-\psi_k(\tilde{X}_j))\right)^2\right|(\tilde X_\ell)_{\ell=1}^n\right)\right)\\
&=\frac{1}{\varepsilon}\sum_{k\geq i}\E_D\left(\frac{1}{n}\sum_{j=1}^n(\psi_k(X_j)-\psi_k(\tilde{X}_j))^2\right)\\
&\leq\frac{2}{\varepsilon}\sum_{k\geq i} \left(\frac{1}{n}\sum_{j=1}^n\psi_k(X_j)\right)^2+\frac{2}{\varepsilon}\sum_{k\geq i}\E_D\left(\psi_k^2(\tilde{X}_1)\right)
\end{align*}
where the first equality follows from replacing $Z_n$ by its definition, the third equality follows from noticing that conditioned on the data $(X_i,\tilde X_i)_{i=1}^n$, $Z_n$ is the sum of independent random variables, $\E(W_i|(X_i,\tilde X_i)_{i=1}^n)=0$ and $W_i^2=1$. Finally, note that the last inequality holds from noticing that $(a-b)^2\leq 2(a^2+b^2)$ for any $a,b\in\R$, and $(\tilde X_i)_{i=1}^n$ are i.i.d. give the data $(X_i)_{i=1}^n$.

By using the previous set of equations, we obtain
\begin{align*}
&\lim_{i\to\infty}\limsup_{n\to\infty}\Prob_D\left(\sum_{k\leq i}Z_n^2(\psi_k)
\geq \varepsilon\right)\\
&\qquad\leq \lim_{i\to\infty}\limsup_{n\to\infty}\frac{2}{\varepsilon}\sum_{k\geq i} \left(\frac{1}{n}\sum_{j=1}^n\psi_k(X_j)\right)^2+\lim_{i\to\infty}\limsup_{n\to\infty}\frac{2}{\varepsilon}\int\sum_{k\geq i}\psi^2_k(x)p_{\thetahn}(x)\lambda(dx)\\
\end{align*}
We shall prove the previous equation is zero. For the first term observe that
\begin{align*}
\lim_{i\to\infty}\limsup_{n\to\infty}\frac{2}{\varepsilon}\sum_{k\geq i} \left(\frac{1}{n}\sum_{j=1}^n\psi_k(X_j)\right)^2&\leq \lim_{i\to\infty}\limsup_{n\to\infty}\frac{2}{n\varepsilon} \sum_{j=1}^n\sum_{k\geq i}\psi^2_k(X_j)\\
&=\lim_{i\to\infty}\frac{2}{\varepsilon} \int\sum_{k\geq i}\psi^2_k(x)p_{\theta_0}(x)\lambda(dx)\\
&=0,
\end{align*}
where the first inequality follows by Jensen's inequality, the first equality follows from the law of large numbers since $\sum_{k\geq i}\psi^2_k(x)\leq \sum_{k\geq 1}\psi_k^2(x)=K(x,x)$ and the kernel is bounded by \Cref{condition:mmd_kernel}. The same argument, combined with Lebesgue's dominated convergence theorem, explains the last equality.

For the second term, the Lebesgue's dominated convergence theorem yields
\begin{align*}
  \lim_{i\to\infty}\limsup_{n\to\infty}\frac{2}{\varepsilon}\int\sum_{k\geq i}\psi_k^2(x)p_{\thetahn}(x)\lambda(dx)=\lim_{i\to\infty}\frac{2}{\varepsilon}\int\sum_{k\geq i}\psi_k^2(x)p_{\theta_0}(x)\lambda(dx)=0.
\end{align*}

Since conditions $G_0$-$G_2$ are satisfied, then \Cref{thm:suplinear} yields that
\begin{align}
    n\MMD^2_W(\P_{\thetahn,n},\Q_n)\overset{\dist_D}{\to}\sum_{i=1}^\infty\lambda_i'Z_i'^2,\label{eqn:dist1}
\end{align}
where $Z_1',\ldots,Z_n'$ are standard normal i.i.d. random variables, and $\lambda_1',\lambda_2',\ldots,$ are the eigenvalues of the operator $T_\nu:\H\to\H$ given by $(T_\nu\omega)(x)=\nu(\omega,K_x)$.

On the other hand, observe that the wild bootstrap for the goodness-of-fit problem satisfies that
\begin{align}
    n\MMD^2_W(P_{\theta_0},\Q_n)\overset{\dist_D}{\to}\sum_{i=1}^\infty\lambda_iZ_i^2,\label{eqn:dist2}
\end{align}
where $Z_1,\ldots,Z_n$ are standard normal i.i.d. random variables, and $\lambda_1,\lambda_2,\ldots,$ are the eigenvalues of the operator $T_\sigma:\H\to\H$ given by $(T_\sigma\omega)(x)=\sigma(\omega,K_x)$ where $\sigma=\frac{1}{2}\nu$ (hence the eigenvalues are half of those in \cref{eqn:dist1}). Therefore, we deduce that for every $x
\in\R$,
\begin{align}
    \limsup_{n\to\infty}\Prob(n \MMD^2_W(\P_{\thetahn,n}, \Q_n)\geq x)&=\limsup_{n\to\infty}\Prob\left(n\MMD^2_W(\P_{\theta_0}, \Q_n)\geq \frac{x}{2}\right)\nonumber\\
    &>\limsup_{n\to\infty}\Prob\left(n\MMD^2_W(\P_{\theta_0}, \Q_n)\geq x\right)\label{eqn:inequaliy}
\end{align}

Replacing $x=q_\alpha$ in the previous equation yields
\begin{align}
    \alpha=\limsup_{n\to\infty}\Prob(n \MMD^2_W(\P_{\thetahn,n}, \Q_n)\geq q_\alpha)%
    &> \limsup_{n\to\infty}\Prob\left(n \MMD^2_W(\P_{\theta_0}, \Q_n)\geq q_\alpha\right)\nonumber\\
    &= \limsup_{n\to\infty}\Prob\left(n \MMD^2(\P_{\theta_0}, \Q_n)\geq q_\alpha\right),\label{eqn:alphaeq}
\end{align}
where the first equality is due to the definition of $q_\alpha$, the first inequality holds by \cref{eqn:inequaliy} and the last equality holds since $n \MMD^2(\P_{\theta_0},\Q_n)$ and $n \MMD^2_W(\P_{\theta_0},\Q_n)$ have the same asymptotic null distribution \citep[Theorem~1]{chwialkowski-wild-bootstrap-kernel-tests}.

Finally note that
\begin{align*}
&\limsup_{n\to\infty}\Prob\left(n \MMD^2(\P_{\thetahn},\Q_n)\geq q_\alpha\right)\\
&\qquad=\limsup_{n\to\infty}\Prob\left(
      n\MMD^2(\P_{\theta_0}, \Q_n)
      - \frac{2}{n}\sum_{i=1}^n \sum_{j=1}^n \phit(X_j,\theta_0)^\intercal \Hb^{-1}\phit(X_i,\theta_0)
      \geq q_\alpha
    \right)\\
    &\qquad <\limsup_{n\to\infty}\Prob\left(
      n\MMD^2(\P_{\theta_0}, \Q_n)\geq q_\alpha
    \right)<\alpha,
\end{align*}
where the first equality holds due to \Cref{lemma:repre}, the first inequality  holds since both $\Hb$ and $\Hb^{-1}$ are positive definite matrices, and thus $\sum_{i=1}^n \sum_{j=1}^n \phit(X_j,\theta_0)^\intercal \Hb^{-1}\phit(X_i,\theta_0)\geq 0$, and the last inequality holds by \cref{eqn:alphaeq}.
\end{proof}

\begin{proof}{\Cref{Thm:2B}}
By the definition of the maximum mean discrepancy, we have
\begin{align*}
    n\MMDk^2(\P_{\theta_n^\star},\Q_n^\star)=\sup_{\unitball}(\gamma_n^\star(\omega))^2,
    \quad\text{where}\quad \gamma_n^\star(\omega)=\frac{1}{\sqrt{n}}\sum_{i=1}^n\omega(X_i^\star)-g(\omega,\theta_n^\star).
\end{align*}
\Cref{Parametric approx} yields $\gamma_n^\star(\omega)=S_n^\star(\omega)+o_p(1)$, where $S_n^\star:\mathcal{H}\to\Reals$ is the functional defined in \cref{eqn:Snfunction}, and the error term does not depend on $\omega$. Then, by using the fact that $\sup_{\unitball}({S}_n^\star(\omega))^2$ converges in distribution (which is proved next) we obtain
\begin{align*}
    n\MMDk^2(\P_{\theta_n^\star},\Q_n^\star)=\sup_{\unitball}(S_n^\star(\omega))^2+o_p(1).
\end{align*}
By Slutsky's theorem, we conclude that $n\MMDk^2(\P_{\theta_n^\star},\Q_n^\star)$ and $\sup_{\unitball}({S}_n^\star(\omega))^2$ have the same asymptotic null distribution.

We use \Cref{thm:suplinear} to obtain the asymptotic null distribution of $\sup_{\unitball}({S}_n^\star(\omega))^2$. Observe however that to apply this theorem, we need to check that
\begin{equation*}
  {S}_n^\star(\omega)=\frac{1}{\sqrt{n}}\sum_{i=1}^n \eta(\omega,X_i^\star,\thetahn)
\end{equation*}
is a bounded linear functional. The linearity follows from the fact that for any fixed $x\in\X$ and $\omega\in\Theta$, $\eta(\omega,x,\theta)$ is a linear function of $\omega$. Boundedness follows from item iii of \Cref{prop:eta} since $\sup_{\theta\in\Theta}\sup_{x\in\X}|\eta(\omega,x,\theta)|< C_1\LH{\omega}+  C_2\LH{\omega}^2$ where $C_1>0$ and $C_2>0$ are constants that do not depend on $\omega$.

We proceed to check Conditions $G_0$, $G_1$ and $G_2$ of \Cref{thm:suplinear}. Condition $G_0$ follows directly from \Cref{Prop:Normal2}, where we identify  $\sigma(\omega,\omega')=\E_{X\sim \P_{\theta_0}}(\eta(\omega,X,\theta_0)\eta(\omega',X,\theta_0))$ for any $\omega,\omega'\in\mathcal{H}$. Indeed, note that it is the same bilinear form of \cref{eq:bilinear}, and thus Condition $G_1$ was already proved in \cref{eq:FiniteSigma} in the proof of \Cref{Thm:2}.

Finally, to check condition $G_2$ observe that
\begin{align}
    \Prob_D\left(\sum_{k\geq i} ({S}_n^\star(\psi_k))^2\geq\varepsilon\right)
    \leq \frac{1}{\varepsilon}\sum_{k\geq i}\E_D\left(({S}_n^\star(\psi_k))^2\right)
    &=\varepsilon^{-1}\sum_{k\geq i}\E_D(\eta^2(\psi_k,X^\star,\thetahn))\nonumber\\
&=\varepsilon^{-1}\sum_{k\geq i}\int_\X\eta^2(\psi_k,z,\thetahn)p_{\thetahn}(z)\lambda(dz)\label{eqn:cmb4}
\end{align}
where $X^\star\sim \P_{\thetahn}|\thetahn$.  The first inequality follows from Markov's inequality, and the first equality follows since ${S}_n^\star(\omega)=\frac{1}{\sqrt{n}}\sum_{i=1}^n\eta(\omega,X_i^\star,\thetahn)$, and recall that $(X_i^\star)_{i=1}^n$ are generated i.i.d. from $\P_{\thetahn}|\thetahn$.

From \Cref{prop:eta}.ii., there exists constants $C_1>0$ and $C_2>0$ such that for all $k\geq 1$ it holds
\begin{align}
   \int_\X\eta^2(\psi_k,z,\thetahn)p_{\thetahn}(z)\lambda(dz)\leq C_1\int_\X \psi^2_k(z)p_{\thetahn}(z)\lambda(dz)+C_2\left\|\nabla_{\theta}g(\psi_k,\theta_0)\right\|^2.\label{eqn:cmb3}
\end{align}
Finally, by combining \cref{eqn:cmb4,eqn:cmb3}, we get
\begin{align*}
&\lim_{i\to\infty}\limsup_{n\to\infty}\Prob_D\left(\sum_{k\geq i}({S}^\star_n(\psi_k))^2\geq\varepsilon\right)\\
  &\leq \lim_{i\to\infty}\limsup_{n\to\infty} \sum_{k\geq i}\varepsilon^{-1} C_1 \int_\X \psi^2_k(z) p_{\thetahn}(z)\lambda(dz) + \varepsilon^{-1} C_2 \lim_{i\to\infty} \sum_{k\geq i} \left\| \nabla_{\theta} g(\psi_k,\theta_0) \right\|^2.
\end{align*}
Note that the second limit of the previous equation converges to zero by  \Cref{prop:secondmomentbound}. For the first limit, observe
\begin{align*}
\lim_{i\to\infty}\limsup_{n\to\infty}\sum_{k\geq i}\int_\X \psi^2_k(z)p_{\thetahn}(z)\lambda(dz) &\leq  \lim_{i\to\infty}\sum_{k\geq i}^\infty\int_\X \psi^2_k(z)\sup_{\theta\in\Theta}p_{\theta}(z)\lambda(dz)= 0
\end{align*}
since $\sum_{k=1}^\infty\psi^2_k(y)=K(y,y)$, and $\int_\X \sup_{\theta\in\Theta}p_{\theta}(z)\lambda(dz)<\infty$  under \Cref{\assumptions}.
\end{proof}

\subsection{Proof of Auxiliary Results}\label{sec:proofauxiliar}
We proceed to prove the auxiliary results.

\begin{proof}{\Cref{lemma:repre}}
By the definition of the MMD, it holds that
\begin{align*}
    n\MMDk^2(\P_{\thetahn},\Q_n)=\sup_{\unitball}\gamma^2_n(\omega),
    \quad\text{where}\quad \gamma_n(\omega)=\frac{1}{\sqrt{n}}\sum_{i=1}^n\omega(X_i)-g(\omega,\thetahn).
\end{align*}
\Cref{Lemma:1} yields $\gamma_n(\omega)=S_n(\omega)+o_p(1)$, where $S_n:\mathcal{H}\to\Reals$ is the functional defined in \cref{eqn:Snfunction}, and the error term does not depend on $\omega$. Then, by using the fact that $\sup_{\unitball}{S}^2_n(\omega)$ converges in distribution- which is proved in \Cref{Thm:2}-
we obtain
\begin{align*}
    n\MMDk^2(\P_{\thetahn},\Q_n)=\sup_{\unitball} S_n^2(\omega) + o_p(1).
\end{align*}

Note that $S_n(\omega)=A_n(\omega)+B_n(\omega),$ where  $A_n:\mathcal{H}\to\Reals$ and $B_n:\mathcal{H}\to\Reals$ are given by
\begin{align*}
A_n(\omega)
=\frac{1}{\sqrt{n}} \sum_{i=1}^n (\omega(X_i)-g(\omega,\theta_0))
\quad\text{and}\quad
B_n(\omega)=-\frac{2}{\sqrt{n}}\sum_{i=1}^n\langle\nabla_\theta g(\omega,\theta_0),\Hb^{-1}\phit(X_i,\theta_0)\rangle.
\end{align*}
Thus
\begin{align}
    n\MMDk^2(\P_{\thetahn},\Q_n)&=\sup_{\unitball}(A_n(\omega)+B_n(\omega))^2+o_p(1)\nonumber\\
    &=(A_n^1+B_n^1)(A_n^2+B_n^2)K+o_p(1)\nonumber\\
    &=n\MMDk^2(\P_{\theta_0},\Q_n)+A_n^1B_n^2K+B_n^1A_n^2K+B_n^1B_n^2K+o_p(1),\label{eqn:ABresult}
\end{align}
where $A_n^iK$ with $i\in\{1,2\}$ denotes the application of $A_n$ to the $i$-the coordinate of the kernel, and the same notation is used for $B_n^i$. The second equality is due to \cite{GFKT}, and the third equality holds since  $n\MMDk^2(\P_{\theta_0},\Q_n)=A_n^1A_n^2K$.

We proceed to analyse $A_n^1B_n^2K$, $B_n^1A_n^2K$ and $B_n^1B_n^2$. Observe that
\begin{align*}
B_n\omega
&=-\frac{2}{\sqrt{n}}\sum_{i=1}^n\left\langle\int_\X\omega(x)\nabla_\theta p_{\theta_0}(x)\lambda(dx),\Hb^{-1}\phit(X_i,\theta_0)\right\rangle
\end{align*}
where the equality follows by Lebesgue's dominated convergence theorem (see \Cref{remark:lebesgue}). For all $x\in\X$, define $K_x:\X\to\Reals$ by $K_x(\cdot)=K(x,\cdot)$. Then, $B_n^2 K\in\mathcal{H}$ is given by
\begin{align}
    (B_n^2K)(x)= B_n K_x&=-\frac{2}{\sqrt{n}}\sum_{i=1}^n\left\langle\int_\X K_x(y)\nabla_\theta p_{\theta_0}(y)\lambda(dy),\Hb^{-1}\phit(X_i,\theta_0)\right\rangle.\label{eqn:WBoot}
\end{align}
By  linearity,  we obtain
\begin{align}
    A_n^1 B_n^2K
    &=-\frac{2}{\sqrt{n}}\sum_{i=1}^n\left\langle\int_\X (A_n^1 K)(y)\nabla_\theta p_{\theta_0}(y)\lambda(dy),\Hb^{-1}\phit(X_i,\theta_0)\right\rangle\label{eqn:ABK}
\end{align}
Note that for any fixed $y\in\X$, it holds
\begin{equation*}
  (A_n^1 K)(y)=\frac{1}{\sqrt{n}}\sum_{j=1}^n \left(K(X_j,y)-\int_\X K(x,y)p_{\theta_0}(x)\lambda(dx)\right).
\end{equation*}
Then, a simple computation shows that
\begin{align*}
\int_\X  (A_n^1 K)(x)  \nabla_\theta p_{\theta_0}(x)\lambda(dx)&=\frac{1}{\sqrt{n}}\sum_{j=1}^n\left(\nabla_\theta\mu_{\theta
_0}(X_j)-\int_\X(\nabla_\theta\mu_{\theta
_0}(x))p_{\theta_0}(x)\lambda(dx)\right)\\
&=\frac{1}{\sqrt{n}}\sum_{j=1}^n \phit(X_j,\theta_0).
\end{align*}
By substituting the previous equation into \cref{eqn:ABK}, we obtain
\begin{align}
    A_n^1B_n^2K
    &=-\frac{2}{n}\sum_{i=1}^n\sum_{j=1}^n\left\langle\phit(X_j,\theta_0),\Hb^{-1}\phit(X_i,\theta_0)\right\rangle
       =-\frac{2}{n}\sum_{i=1}^n\sum_{j=1}^n\phit(X_j,\theta_0)^\intercal\Hb^{-1}\phi(X_i,\theta_0).\label{eqn:ABK2}
\end{align}
A similar computation shows that $B_n^1A_n^2K=A_n^1B_n^2K$. For the term $B_n^1B_n^2K$,  we have
\begin{align}
    &B_n^1B_n^2K\nonumber\\
    &=-\frac{2}{\sqrt{n}}\sum_{i=1}^n\left\langle \int_\X (B_n^1K){(x)}\nabla_\theta p_{\theta_0}(x)\lambda(dx),\Hb^{-1}\phit(X_i,\theta_0)\right\rangle\nonumber\\
    &=\frac{4}{n}\sum_{i=1}^n\sum_{j=1}^n\left\langle \int_\X \left( \left\langle\int_\X K_x(y)\nabla_\theta p_{\theta_0}(y)\lambda(dy), \Hb^{-1}\phit(X_j,\theta_0)\right\rangle \right) \nabla_\theta p_{\theta_0}(x)\lambda(dx),
    \right. \nonumber \\
    &\hspace{7em} \left. \Hb^{-1}\phit(X_i,\theta_0)\right\rangle \nonumber \\
    &=\frac{4}{n}\sum_{i=1}^n\sum_{j=1}^n\int_\X\int_\X K(x,y)\left\langle\nabla_\theta p_{\theta_0}(y),\Hb^{-1}\phit(X_j,\theta_0)\right\rangle\left\langle\nabla_\theta p_{\theta_0}(x),\Hb^{-1}\phit(X_i,\theta_0)\right\rangle\lambda(dy)\lambda(dx)\nonumber\\
    &=\frac{2}{n}\sum_{i=1}^n\sum_{j=1}^n\sum_{\ell=1}^p\sum_{m=1}^p(\Hb^{-1}\phit(X_i,\theta_0))_\ell(\Hb^{-1}\phit(X_j,\theta_0))_m\Hb_{m,\ell}\nonumber\\
    &=\frac{2}{n}\sum_{i=1}^n\sum_{j=1}^n\phit(X_j,\theta_0)^\intercal\Hb^{-1}\phit(X_i,\theta_0),\label{eqn:BBK}
\end{align}
where the second equality holds from replacing $(B_n^2K)(x)$ by the expression given in  \cref{eqn:WBoot}, and the fourth equality holds since
\begin{align*}
    \Hb_{m,\ell}=2\int_\X \int_\X K(x,y)\left(\frac{\partial}{\partial\theta_m}p_{\theta_0}(y)\right)\left(\frac{\partial}{\partial\theta_\ell}p_{\theta_0}(x)\right)\lambda(dy)\lambda(dx).
\end{align*}

By substituting \cref{eqn:ABK2} and \cref{eqn:BBK} into \cref{eqn:ABresult}, we obtain
\begin{align*}
n\MMDk^2(\P_{\thetahn},\Q_n)=n\MMDk^2(\P_{\theta_0},\Q_n)-\frac{2}{n}\sum_{i=1}^n\sum_{j=1}^n\phit(X_j,\theta_0)^\intercal\Hb^{-1}\phit(X_i,\theta_0)+o_p(1).
\end{align*}
\end{proof}

\begin{proof}{\Cref{prop:normalWBOscar}}
Note that conditioned on the data $\widetilde D=(X_i,\tilde X_i)_{i=1}^n$, $Z_n(\omega)$ is a sum of i.i.d. random variables with mean given by
\begin{equation*}
  \E_{\widetilde D}\left(Z_n(\omega)\right)
  =\E_{ {\widetilde D}}\left(\frac{1}{\sqrt{n}}\sum_{i=1}^n W_i(\omega(X_i)-\omega(\tilde{X}_i))\right)
  =0,
\end{equation*}
and variance given by $ \Var_{\widetilde D}(Z_n(\omega))=\frac{1}{n}\sum_{i=1}^n(\omega(X_i)-\omega(\tilde{X}_i))^2$. We claim that
\begin{align*}
   \frac{1}{n}\sum_{i=1}^n(\omega(X_i)-\omega(\tilde{X}_i))^2&\overset{\Prob}{\to}2\int(\omega(x)-g(\omega,\theta_0))^2p_{\theta_0}(x)\lambda(dx)=\nu(\omega,\omega).
\end{align*}
Thus, by Lyapunov’s Central limit theorem, we obtain $Z_n(\omega)\overset{\dist_D}{\to} N(0,\nu(\omega,\omega))$ which yields the desired result.

To prove the claim, consider $\xi_n=\frac{1}{n}\sum_{i=1}^n(\omega(X_i)-\omega(\tilde{X}_i))^2$. We will show that $\E(\xi_n)=\nu(\omega,\omega)$ and that $\Var(\xi_n)\to0$. Therefore the convergence in probability follows from Markov's inequality.

Consider the data $D=(X_i)_{i\geq 1}$, and define $\E_{D}(\cdot)=\E(\cdot|(X_i)_{i=1}^n)$. Then,  observe that
\begin{align}
    \E\left(\xi_n\right)
    & =\E\left(\left(\omega(X_1)-\omega(\tilde{X}_1)\right)^2\right)
    = \E\left(\E_{D}\left(\left(\omega(X_1)-\omega(\tilde{X}_1)\right)^2\right)\right)\nonumber\\
    &=\E\left(\omega^2(X_1)-2\omega(X_1)\E_{D}(\omega(\tilde{X}_1))+\E_{D}(\omega^2(\tilde{X}_1))\right),\label{eqn:variance}
\end{align}
where
\begin{align*}
 \E_{D}\left(\omega(\tilde{X}_1)\right)=\int \omega(z)p_{\thetahn}(z)\lambda(dz)\quad\text{and}\quad \E_{D}\left(\omega^2(\tilde{X}_1)\right)
 =\int \omega^2(z)p_{\thetahn}(z)\lambda(dz).
\end{align*}
Recall that $\int \sup_{\theta\in\Theta}p_{\theta}(x)\lambda(dx)<\infty$ by \Cref{condition:model}, and that $\omega\in\mathcal{H}$ is bounded by \cref{condition:mmd_kernel}. Then, under the null hypothesis, and by applying Lebesgue's dominated convergence theorem twice, it holds
\begin{align*}
\lim_{n\to\infty} \E \left( \E_{D} \left(\omega(\tilde{X}_1)\right)\right)
= \E\left(\lim_{n\to\infty}\int \omega(z)p_{\thetahn}(z)\lambda(dz)\right)=\int \omega(z)p_{\theta_0}(z)\lambda(dz).
\end{align*}
By using the same arguments we can show that
$\lim_{n\to\infty}\E(\E_{D}(\omega^2(\tilde{X}_1)))=  \int \omega^2(z)p_{\theta_0}(z)\lambda(dz)$.

The previous equations together with \cref{eqn:variance}, yield
\begin{align*}
\lim_{n\to\infty} \E\left(\xi_n\right)
& = 2\left(\int \omega^2(z) p_{\theta_0}(z)-\left(\int \omega(z)p_{\theta_0}(z)\right)^2\right)=\nu(\omega,\omega).
\end{align*}

Finally, we proceed to verify that $\lim_{n\to\infty}\Var(\xi_n)=0$.
Note that
\begin{equation*}
  \Var\left(\xi_n\right)
  = \Var\left(\E_{D}(\xi_n)\right)+\E(\Var_{D}\left(\xi_n\right)),
\end{equation*}
and by the previous arguments $\lim_{n\to\infty}\Var\left(\E_{D}(\xi_n)\right)=0$ since $\E_{D}(\xi_n)$ converges to a constant. Note that conditioned on the data $D$, the random variables $(\omega(X_i)-\omega(\tilde X_i))_{i=1}^n$ are independent. Thus
\begin{align*}
\Var_{D} \left(\xi_n\right)
= \frac{1}{n^2} \sum_{i=1}^n\Var_{D}\left(\left(\omega(X_i)-\omega(\tilde{X}_i)\right)^2\right)\to 0,
\end{align*}
where the limit holds since the $\omega$ is bounded. Hence we conclude that $\lim_{n\to\infty}\Var(\xi_n)=0$.

\end{proof}

\begin{proof}{\Cref{Prop:convergence}}
To prove this result we use \Cref{Thm:preliminary1}. Note that by \Cref{condition:domains} the set $\Theta$ is bounded and closed, by \Cref{condition:model} we have $L(\theta)\in\mathcal{C}^0(\Theta)$, and by \Cref{condition:mmd_kernel} the kernel $K$ is characteristic, which means that the maximum mean discrepancy is a distance between probability measures. Thus, since the parametric family $\{ \P_\theta \}_{\theta \in \Theta}$ is identifiable (\Cref{condition:model}), we deduce that $L(\theta)=\MMDk^2(\P_{\theta},\P_{\theta_0})$  has a unique minimiser under the null hypothesis.

Observe that
\begin{align*}
    \sup_{\theta\in\Theta}|L_n(\theta)-L(\theta)|
    & =\sup_{\theta\in\Theta} \left|\LH{\mu_\theta-\mu_n}^2-\LH{\mu_\theta-\mu_{\theta_0}}^2 \right| \\
    &\leq \sup_{\theta\in\Theta} \left|2\InerH{\mu_\theta}{\mu_{\theta_0}-\mu_n}\right|+\left|\LH{\mu_n}^2-\LH{\mu_{\theta_0}}^2\right|\\
    &\leq2\sup_{\theta\in\Theta}\LH{\mu_\theta}\LH{\mu_{\theta_0}-\mu_n}+\left|\LH{\mu_n}^2-\LH{\mu_{\theta_0}}^2\right|\\
    &\leq C\LH{\mu_{\theta_0}-\mu_n}+\left|\LH{\mu_n}^2-\LH{\mu_{\theta_0}}^2\right|\to0
\end{align*}
in probability, where the last result holds  by the standard law of large numbers for
RKHS \citet[Section 9.1.1-9.1.2]{berlinetReproducingKernelHilbert}. Then, by \Cref{Thm:preliminary1}, we deduce that $\thetahn\to\theta_0$ in probability.
\end{proof}

\begin{proof}{\Cref{prop:Normal}}
Recall the definition $\phit$ in \cref{eqn:phi}, then
\begin{align*}
    \nabla_\theta L_n(\theta)= \nabla_\theta\LH{\mu_n-\mu_{\theta}}^2
    &=-2\InerH{ \nabla_\theta\mu_{\theta}}{\mu_n-\mu_{\theta}}\\
    &=-\frac{2}{n}\sum_{i=1}^n\left(\nabla_\theta\mu_{\theta}(X_i)-\int_\X\left( \nabla_\theta\mu_{\theta}(x)\right)p_{\theta}(x)\lambda(dx)\right)\\
    &=-\frac{2}{n}\sum_{i=1}^n\phit(X_i,\theta).
\end{align*}
By doing the same computations, we obtain
$\nabla_\theta L_n^\star(\theta)=\nabla_\theta\LH{\mu_n^\star-\mu_{\theta}}^2=-\frac{2}{n}\sum_{i=1}^n\phit(X_i^\star,\theta),$
where recall that $\mu_n^\star$ is the mean kernel embedding of the empirical distribution $\Q_n^\star$ obtained from the parametric bootstrap samples $X_1^\star,\ldots,X_n^\star$.

We proceed to prove item i). Observe that clearly,
\begin{align*}
\E_{X\sim \P_\theta}(\phit(X,\theta))&=\E_{X\sim \P_\theta}\left(\nabla_\theta \mu_\theta(X)-\int_\X (\nabla_\theta \mu_\theta(y))p_\theta(y)\lambda(dy)\right)=0,
\end{align*}
and similarly $\E_D(\phit(X_i^\star,\thetahn))=0$. We continue by checking ii). For any fixed $x\in\X$ and $\theta\in\Theta$, we have
\begin{align}
\left\|\phit(x,\theta)\right\|^2&=\sum_{i=1}^p\left(\frac{\partial}{\partial\theta_i}\mu_\theta(x)-\int_\X \left(\frac{\partial}{\partial\theta_i}\mu_\theta(y)\right)p_\theta(y)\lambda(dy)\right)^2
 \leq 4\sum_{i=1}^p\left(\sup_{x\in\X}\left|\frac{\partial}{\partial\theta_i}\mu_\theta(x)\right|\right)^2.\label{eqn:e1q1}
\end{align}
Additionally,  by similar arguments as those shown in \Cref{remark:lebesgue} we have that for any $i\in \{1,\ldots,p\}$, it holds
\begin{align*}
\sup_{\theta\in\Theta}\sup_{x\in\X}\left|\frac{\partial}{\partial\theta_i}\mu_\theta(x)\right|=\sup_{\theta\in\Theta}\sup_{x\in\X}\left|\int_\X K(x,y)\frac{\partial}{\partial\theta_i}p_\theta(y)\lambda(dy)\right|<\infty.
\end{align*}

We finish by verifying item iii). Recall that $\nabla_\theta L_n(\theta_0)=-\frac{2}{n}\sum_{i=1}^n\phit(X_i,\theta_0)$, where $\phit(X_1,\theta_0),\ldots,\phit(X_n,\theta_0)$ are i.i.d. random variables. Moreover, under the null hypothesis, the items i) and ii) of this proposition deduce $\E(\phit(X_i,\theta_0))=0$ and $\E(\|\phit(X_i,\theta_0)\|^2)<\infty$.
Thus, $\E(\|\sqrt{n}\nabla_{\theta}L_n(\theta_0)\|^2)=4\E(\|\phit(X_1,\theta_0)\|^2)<\infty$, and we obtain
$\|\sqrt{n}\nabla_{\theta}L_n(\theta_0)\|=O_p(1)$.

By using the same arguments we deduce that for almost all sequences $(X_i)_{i\geq1}$, $$\E_D(\|\sqrt{n}\nabla_{\theta}L_n^\star(\thetahn)\|^2)= 4\E_D(\|\phit(X_1^\star,\thetahn)\|^2)<\infty.$$  Thus $\|\sqrt{n}\nabla_{\theta}L_n^\star(\thetahn)\|=O_p(1)$.
\end{proof}

\begin{proof}{\Cref{prop:eta}}
We start with i). Observe that
\begin{align*}
    \E_{X\sim \P_{\theta}}\left(\eta(\omega,X,\theta)\right)%
    &=\E_{X\sim \P_{\theta}}\left({\omega}(X)- \g{\theta}\right)-2\left\langle\nabla_{\theta}\g{\theta_0},\Hb^{-1}\E_{X\sim \P_{\theta}}\left(\phit(X,\theta)\right)\right\rangle=0,
\end{align*}
where the first equality follows from the linearity of expectations, and the last equality is due to $g(\omega,\theta)=\E_{X\sim \P_\theta}(\omega(X))$ and $\E_{X\sim \P_\theta}(\phit(X,\theta))=0$  from \Cref{prop:Normal}, item i).

We continue by verifying items ii) and iii). Observe that for any fixed $\omega\in\mathcal{H}$, $x\in\X$ and $\theta\in\Theta$, we have that
\begin{align}
|\eta(\omega,x,\theta)|
&=\left|{\omega}(x)-\g{\theta}-2\left\langle\nabla_{\theta}\g{\theta_0},\Hb^{-1}\phit(x,\theta)\right\rangle\right|\nonumber\\
&\leq |{\omega}(x)-\g{\theta}|+2\left\|\nabla_{\theta}\g{\theta_0}\right\|\|\Hb^{-1}\|_F\left\|\phit(x,\theta)\right\|\nonumber\\
&\leq|{\omega}(x)-\g{\theta}|+C'\left\|\nabla_{\theta}\g{\theta_0}\right\|\label{eqn:|eta|}
\end{align}
where $\|\cdot\|_F$ denotes the Frobenius norm, and $C'>0$ is a constant that does not depend on $\omega,$ $x$, nor $\theta$. In the previous set of equations, the first inequality follows from the triangle inequality and the Cauchy-Schwarz's inequality, and the second inequality holds by taking $C'=2\|\Hb^{-1}\|_F\sup_{\theta\in\Theta}\sup_{x\in\X}\left\|\phit(x,\theta)\right\|<\infty$  since $\|\Hb^{-1}\|_F<\infty$, and $\sup_{\theta\in\Theta}\sup_{x\in\X}\left\|\phit(x,\theta)\right\|<\infty$  by item ii) of \Cref{prop:Normal}.

We proceed to verify item ii). From \cref{eqn:|eta|}, we have that for any fixed $\omega\in\mathcal{H}$ and $\theta\in\Theta$,
\begin{align*}
  \E_{X\sim \P_\theta}\left(\eta^2(\omega,X,\theta)\right)&\leq 2\E_{X\sim \P_\theta}\left(\left|{\omega}(X)-\g{\theta}\right|^2\right)+2{C'}^2\left\|\nabla_{\theta}\g{\theta_0}\right\|^2\\
  &\leq C_1\E_{X\sim \P_\theta}\left(\omega^2(X)\right)+C_2\left\|\nabla_{\theta}\g{\theta_0}\right\|^2,
\end{align*}
where $C_1=2$ and $C_2=2C'^2$.  The first inequality holds from the fact that for any $a,b\in\Reals$ we have that $(a+b)^2\leq 2(a^2+b^2)$, and the second inequality follows from the fact that the variance of random variable is upper bounded by its second moment.

We continue with item  iii). Note that for each $\omega\in\mathcal{H}$, we have $\sup_{x\in\X}|\omega(x)|\leq \LH{\omega}\sqrt{C}$ (see \Cref{remakH}), and thus
\begin{align}
    \sup_{\theta\in\Theta} |g(\omega,\theta)|&\leq \sup_{\theta\in\Theta}\int_\X |\omega(x)|p_\theta(x)\lambda(dx)\leq \LH{\omega}\sqrt{C}.\label{eqn: 15iii1}
\end{align}
Additionally, note that (see \Cref{remark:lebesgue})
\begin{align} \sup_{\theta\in\Theta}\left\|\nabla_{\theta}\g{\theta}\right\|^2&= \sup_{\theta\in\Theta}\sum_{i=1}^p\left(\int_\X \omega(x)\frac{\partial}{\partial\theta_i}p_\theta(x)\lambda(dx)\right)^2\nonumber\\
  &\leq \LH{\omega}^2 C\sum_{i=1}^p\left(\int_\X\sup_{\theta\in\Theta}\left|\frac{\partial}{\partial\theta_i}p_{\theta}(y)\right|\lambda(dy)\right)^2<\infty,\label{eqn: 15iii2}
\end{align}
where the equality holds under \Cref{condition:model} by the Lebesgue's dominated convergence theorem (see \Cref{remark:lebesgue}), the first inequality is due to the fact that   $\sup_{x\in\X}|\omega(x)|\leq \LH{\omega}\sqrt{C}$, and the last inequality holds by \Cref{condition:model}. Then, by combining \cref{eqn: 15iii1,eqn:|eta|,eqn: 15iii2}, we get
\begin{align*}
\sup_{\theta\in\Theta}\sup_{x\in\X}|\eta(\omega,x,\theta)| &\leq   2\LH{\omega}\sqrt{C}+  \LH{\omega}^2 C\sum_{i=1}^p\left(\int_\X\sup_{\theta\in\Theta}\left|\frac{\partial}{\partial\theta_i}p_{\theta}(y)\right|\lambda(dy)\right)^2\\
&\leq   C_1\LH{\omega}+  C_2\LH{\omega}^2 <\infty,
\end{align*}
where $C_1>0$ and $C_2>0$ are constants that do not depend on $\omega$, $x$ or $\theta$.

\end{proof}

\begin{proof}{\Cref{prop:uniformconvergence}}
We begin by defining $Z_n(\theta)=|Y_n(\theta)-\gamma(\theta)|$ and $Z_n^\star(\theta)=|Y_n^\star(\theta)-\gamma(\theta)|$, where
\begin{align*}
    Y_n(\theta)&=\frac{1}{n}\sum_{i=1}^n\frac{\partial^2}{\partial\theta_\ell\partial\theta_j}\mu_{\theta}(X_i),\quad \gamma(\theta)=\E_{X\sim \P_{\theta_0}}\left(\frac{\partial^2}{\partial\theta_\ell\partial\theta_j}\mu_{\theta}(X)\right),\quad\text{and}\\
    Y_n^\star(\theta)&=\frac{1}{n}\sum_{i=1}^n\frac{\partial^2}{\partial\theta_\ell\partial\theta_j}\mu_{\theta}(X_i^\star).
\end{align*}
Observe  that \cref{eqn:unifconv1} is equivalent to $\sup_{\theta\in\Theta}Z_n(\theta)\overset{\Prob}{\to}0$ as n grows to infinity, and \cref{eqn:unifconv2} is equivalent to  $\Prob_D\left(\sup_{\theta\in\Theta}Z_n^\star(\theta)\geq \varepsilon\right)\overset{\Prob}{\to} 0$ holds for any $\varepsilon>0$.

To verify \cref{eqn:unifconv1} we use Theorem 21.9 of Davidson (1994), from which  we just need to check the following properties: (i) $Z_n(\theta)\overset{\Prob}{\to}0$ for each fixed $\theta\in\Theta$, and (ii) $Z_n(\cdot)$ is stochastically equicontinuous. Analogously, for \cref{eqn:unifconv2} we need to check (i') $\Prob_D\left(Z_n^\star(\theta)\geq \varepsilon\right)\overset{\Prob}{\to} 0$, and (ii') for almost all sequences $(X_i)_{i\geq 1}$, $Z_n^\star(\cdot)$ is stochastically equicontinuous.

We start proving (i) and (i'). Note that (i) follows from the law of large numbers since
 \begin{align*}
   \E_{X\sim \P_{\theta_0}}\left(\left|\frac{\partial^2}{\partial\theta_\ell\partial\theta_j}\mu_{\theta}(X)\right|\right)&=\E_{X\sim \P_{\theta_0}}\left(\left|\int_\X K(X,y)\frac{\partial^2}{\partial\theta_\ell\partial\theta_j}p_{\theta}(y)\lambda(dy)\right|\right)<\infty,
 \end{align*}
holds under \Cref{condition:model} (see \Cref{remark:lebesgue}). For item (i'), note that for any $\varepsilon>0$ we have
 \begin{align*}
   \Prob_D\left(Z_n^\star(\theta)\geq \varepsilon\right)&=  \Prob_D\left(|\gamma(\theta)-Y_n^\star(\theta)|\geq \varepsilon\right)\\
   &\leq \Prob_D\left(|\gamma_n(\theta)-Y_n^\star(\theta)|\geq \varepsilon/2\right)+\Prob_D\left(|\gamma(\theta)-\gamma_n(\theta)|\geq \varepsilon/2\right),
 \end{align*}
 where $\gamma_n(\theta)=\E_{X\sim \P_{\thetahn}}\left(\frac{\partial^2}{\partial\theta_\ell\partial\theta_j}\mu_{\theta}(X)\right)$.
The last two terms converge to zero. To see this observe that
\begin{align*}
    \Prob_D\left(\left|\gamma_n(\theta)-Y_n^\star(\theta)\right|\geq \frac{\varepsilon}{2}\right)&\leq 4\frac{\E_D(|\gamma_n(\theta)-Y_n^\star(\theta)|^2)}{\varepsilon^2}\leq\frac{4}{\varepsilon^{2}n}\E_D\left(\left(\frac{\partial^2}{\partial\theta_\ell\partial\theta_j}\mu_{\theta}(X^\star_1)\right)^2\right)\to0,
\end{align*}
where the first inequality is due to  Chebyshev's inequality, and the second inequality holds since $X_1^\star,\ldots,X_n^\star\overset{i.i.d.}{\sim}\P_{\thetahn}|\thetahn$. For the second term
\begin{align*}
    |\gamma_n(\theta)-\gamma(\theta)|&= \left|\int_\X \frac{\partial^2}{\partial\theta_\ell\partial\theta_j}\mu_{\theta}(x) (p_{\thetahn}(x)-p_{\theta_0}(x))\lambda(dx)\right|\\
    &\leq \sup_{x\in\X}\left|\frac{\partial^2}{\partial\theta_\ell\partial\theta_j}\mu_{\theta}(x)\right| \int_\X | (p_{\thetahn}(x)-p_{\theta_0}(x))|\lambda(dx)
\end{align*}
now, by Taylor's theorem, for each $x\in \X$, it exists $\theta_x$ in the line between $\theta_0$ and $ \thetahn$ such that $p_{\thetahn}(x)-p_{\theta_0}(x)= \nabla_\theta p_{\theta_x}(x) \cdot (\thetahn-\theta_0)$. By the previous equality and by Holder's inequality
\begin{align*}
    |\gamma_n(\theta)-\gamma(\theta)|&\leq \sup_{x\in\X}\left|\frac{\partial^2}{\partial\theta_\ell\partial\theta_j}\mu_{\theta}(x)\right| \left(\int_\X \left\|\nabla_\theta p_{\theta_x}(x)\right\|_1\lambda(dx)\right) \|\thetahn-\theta_0\|_\infty\\
    &\leq \sup_{x\in\X}\left|\frac{\partial^2}{\partial\theta_\ell\partial\theta_j}\mu_{\theta}(x)\right| \left(\int_\X \sup_{\theta\in\Theta}\left\|\nabla_\theta  p_{\theta}(x)\right\|_1\lambda(dx)\right) \|\thetahn-\theta_0\|_\infty\to 0,
\end{align*}
since $|\thetahn-\theta_0|=o_p(1)$ by \Cref{Prop:convergence}, and since the other terms are finite due to \Cref{condition:model,condition:mmd_kernel} and \Cref{remark:lebesgue}.%

We continue by checking (ii) and (ii'). To check ii) we use Theorem 21.10 of Davidson (1994) from which it is enough to verify that $|Z_n(\theta)-Z_n(\theta')|\leq B \|\theta-\theta'\|_{\infty}$ holds for all $\theta,\theta'\in\Theta$ and for all $n$, where $B$ is a constant which does not depend of $n$, $\theta$ and $\theta'$. To check condition ii') we will verify that there exists $B^\star>0$  such that $|Z_n^\star(\theta)-Z_n^\star(\theta')|\leq B^\star \|\theta-\theta'\|_\infty,$ holds for every sequence $(X_i)_{i\geq 1}$. We only verify that $|Z_n(\theta)-Z_n(\theta')|\leq B \|\theta-\theta'\|_{\infty}$ for some constant $B$ since the analogous result for parametric bootstrap follows from the same arguments.

By the triangle inequality, we obtain
\begin{align*}
  |Z_n(\theta)-Z_n(\theta')|&\leq  \left|\gamma(\theta)-\gamma(\theta')\right|+\left|Y_n(\theta)-Y_n(\theta')\right|.
\end{align*}
Since the kernel is bounded by a constant $C>0$, it holds
\begin{align}
 \left|\gamma(\theta)-\gamma(\theta')\right|&=\left|\E_{X\sim \P_{\theta_0}}\left(\frac{\partial^2}{\partial\theta_\ell\partial\theta_j}\mu_{\theta}(X)-\frac{\partial^2}{\partial\theta_\ell\partial\theta_j}\mu_{\theta'}(X)\right)\right|\nonumber\\
 &=\left|\E_{X\sim \P_{\theta_0}}\left(\int_\X K(X,y)\frac{\partial^2}{\partial\theta_\ell\partial\theta_j}(p_{\theta}(y)-p_{\theta'}(y))\lambda(dy)\right)\right|\nonumber\\
 &\leq C\int_\X\left|\frac{\partial^2}{\partial\theta_\ell\partial\theta_j}(p_{\theta}(y)-p_{\theta'}(y))\right|\lambda(dy),\label{eq:11}
\end{align}
and
\begin{align}
 \left|Y_n(\theta)-Y_n(\theta')\right|
 &=\left|\frac{1}{n}\sum_{i=1}^n\int_\X K(X_i,y)\frac{\partial^2}{\partial\theta_\ell\partial\theta_j}(p_\theta(y)-p_{\theta'}(y))\lambda(dy)\right|\nonumber\\
 &\leq C\int_\X\left|\frac{\partial^2}{\partial\theta_\ell\partial\theta_j}(p_{\theta}(y)-p_{\theta'}(y))\right|\lambda(dy),\label{eq:12}
\end{align}

By combining \cref{eq:11} and \cref{eq:12}, we get
\begin{align*}
    |Z_n(\theta)-Z_n(\theta')|\leq 2C\int_\X\left|\frac{\partial^2}{\partial\theta_\ell\partial\theta_j}(p_{\theta}(y)-p_{\theta'}(y))\right|\lambda(dy),
\end{align*}
By the mean value theorem, we have $\frac{\partial^2}{\partial\theta_\ell\partial\theta_j}(p_{\theta}(y)-p_{\theta'}(y))=\nabla_{\theta}\left(\frac{\partial^2}{\partial\theta_\ell\partial\theta_j}p_{\theta_y}(y)\right)\cdot (\theta-\theta')$, where $\theta_y$ is in the line between $\theta$ and $\theta'$ (which belongs to $\Theta$ by convexity of this set). By Holder's inequality, we obtain
\begin{align*}
    \left|\frac{\partial^2}{\partial\theta_\ell\partial\theta_j}(p_{\theta}(y)-p_{\theta'}(y))\right|&\leq \left\|\nabla_{\theta}\left(\frac{\partial^2}{\partial\theta_\ell\partial\theta_j}p_{\theta_y}(y)\right)\right\|_1\|\theta-\theta'\|_\infty\\&\leq\left( \sup_{\theta\in\Theta}\left\|\nabla_{\theta}\left(\frac{\partial^2}{\partial\theta_\ell\partial\theta_j}p_{\theta}(y)\right)\right\|_1\right)\|\theta-\theta'\|_\infty.
\end{align*}
Then, we conclude that
\begin{align*}
  |Z_n(\theta)-Z_n(\theta')|\leq \underbrace{2C\int_\X \sup_{\theta\in\Theta}\left\|\nabla_{\theta}\left(\frac{\partial^2}{\partial\theta_\ell\partial\theta_j}p_{\theta}(y)\right)\right\|_1\lambda(dy)}_{B_{\ell,j}}\|\theta-\theta'\|_\infty.
\end{align*}
Finally, choose $B=\max_{\ell,j\in[p]}B_{\ell,j}$ and note it is finite by \Cref{condition:model} together with \Cref{remark:lebesgue}.
\end{proof}

\begin{proof}{\Cref{prop:convergence of Hessian}}
We start by proving $\|\Hb_n-\Hb\|=o_p(1)$. To prove this result, it suffices to show that each component of $(\Hb_n)_{ij}$ converges in probability to $\Hb_{ij}$.
Observe that
\begin{align*}
(\Hb_n)_{ij}&=\frac{\partial^2}{\partial\theta_i\partial\theta_j}L_n(\tilde{\theta}^j)=\frac{\partial^2}{\partial\theta_i\partial\theta_j}(L_n(\tilde{\theta}^j)-L(\tilde{\theta}^j))+\frac{\partial^2}{\partial\theta_i\partial\theta_j}L(\tilde{\theta}^j).
\end{align*}

A simple computation shows that for each $\theta\in\Theta$ it holds
\begin{align*}
  \frac{\partial^2}{\partial\theta_i\partial\theta_j}(L_n(\theta)-L(\theta))=2\left(\E_{X\sim \P_{\theta_0}}\left(\frac{\partial^2}{\partial\theta_i\partial\theta_j}\mu_{\theta}(X)\right)-\frac{1}{n}\sum_{i=1}^n\frac{\partial^2}{\partial\theta_i\partial\theta_j}\mu_{\theta}(X_i)\right)=o_p(1).
\end{align*}
where the second equality holds by \Cref{prop:uniformconvergence}. Thus,
\begin{align*}
 (\Hb_n)_{ij}%
 &=o_p(1)+ \frac{\partial^2}{\partial\theta_i\partial\theta_j}L(\tilde{\theta}^j).
\end{align*}

Finally, note that $L(\theta)=\LH{\mu_\theta-\mu_{\theta_0}}^2\in\mathcal{C}^0(\Theta)$ since $\mu_\theta(x)\in\mathcal{C}^3(\Theta)$ for all $x\in\X$ (see \Cref{condition:model} and \Cref{remark:lebesgue}). Then, the  continuous mapping theorem, together with the fact that $\tilde\theta_n^j\overset{\Prob}{\to}\theta_0$ (by \Cref{Prop:convergence}), yields
\begin{align*}
   (\Hb_n)_{ij}= \frac{\partial^2}{\partial\theta_i\partial\theta_j}L({\theta}_0)+o_p(1)=\Hb_{ij}+o_p(1).
\end{align*}

We continue with the result for $\Hb_n^\star$.  Similarly to what we did before, we sum and subtract the term $\frac{\partial^2}{\partial\theta_i\partial\theta_j}L({\tilde{\theta}_\star^j})$ to $(\Hb_n)^\star_{ij}$ and obtain
\begin{align*}
 (\Hb_n)^\star_{ij}
 = \frac{\partial^2}{\partial\theta_i\partial\theta_j} \left(L_n^\star({\tilde{\theta}_\star^j})-L({\tilde{\theta}_\star^j})\right) + \frac{\partial^2}{\partial\theta_i\partial\theta_j}L({\tilde{\theta}_\star^j}).
\end{align*}
 Observe that for any $\theta\in\Theta$, under the null hypothesis, we have
\begin{align*}
  \frac{\partial^2}{\partial\theta_i\partial\theta_j} \left(L_n^\star(\theta)-L(\theta) \right)
  = 2\left(\E_{X\sim \P_{\theta_0}}\left(\frac{\partial^2}{\partial\theta_i\partial\theta_j}\mu_{\theta}(X)\right)-\frac{1}{n}\sum_{i=1}^n\frac{\partial^2}{\partial\theta_i\partial\theta_j}\mu_{\theta}(X_i^\star)\right).
\end{align*}
Then, the result follows from \Cref{prop:uniformconvergence} and from the fact that $\thetahn^\star-\theta_0=o_{p}(1)$ and the same arguments as before.
\end{proof}

\begin{proof}{\Cref{prop:normallimittheta}}
We start by proving item i). By \Cref{\assumptions}, we have that $L_n(\theta)=\LH{\mu_\theta-\mu_{n}}^2\in\mathcal{C}^3(\Theta)$ since $\mu_\theta(x)\in\mathcal{C}^3(\Theta)$ for all $x\in\X$ (see \cref{remark:lebesgue}). Then, for each $j\in \{1,\ldots,p\}$, a first order Taylor's expansion of $\frac{\partial}{\partial \theta_j}L_n(\theta)$ around $\thetahn$ exists and is given by
\begin{align*}
\frac{\partial}{\partial \theta_j}L_n(\theta)&=\frac{\partial}{\partial \theta_j}L_n(\thetahn)+\left\langle\nabla_\theta\left(\frac{\partial}{\partial \theta_j}L_n(\tilde\theta^j)\right) ,\theta-\thetahn\right\rangle=\left\langle\nabla_\theta\left(\frac{\partial}{\partial \theta_j}L_n(\tilde\theta^j)\right),\theta-\thetahn\right\rangle,
\end{align*}
where $\tilde\theta^j$ lies in the line segment between $\thetahn$ and $\theta$, and $\tilde\theta^j\in\Theta$ by convexity. The second equality holds since $\thetahn$ is the minimiser of $L_n$, and belongs to the interior of $\Theta$. By using the previous equation and by evaluating at $\theta_0$, we obtain $\nabla_\theta L_n(\theta_0)=\mathbf{H}_n^\intercal (\theta_0-\thetahn)$, where $(\mathbf{H}_n)_{ij}=\frac{\partial^2}{\partial\theta_i\partial\theta_j}L_n(\tilde{\theta}^j)$  for each $i,j\in[p]$.

By \Cref{prop:convergence of Hessian} we have that $\Hb_n$ converges in probability to the Hessian matrix $\Hb$. Recall that $\Hb$ is positive definite by \Cref{condition:invertible}.  Then, by the continuity of the determinant, we have that $\Hb_n$ is invertible in sets of arbitrarily large probability for all $n$ large enough, and thus
\begin{align*}
     \sqrt{n}(\theta_0-\thetahn)&=\sqrt{n}({\mathbf{H}_n^\intercal})^{-1} \nabla_\theta L_n(\theta_0)\\
     &=\sqrt{n}({\mathbf{H}^\intercal})^{-1} \nabla_\theta L_n(\theta_0)+ \sqrt{n}(({\mathbf{H}_n^\intercal})^{-1}-{\mathbf{H}}^{-1}) \nabla_\theta L_n(\theta_0).
\end{align*}
By \Cref{prop:Normal}, $\|\sqrt{n}\nabla_{\theta}L_n(\theta_0)\|=O_p(1)$, and since $\mathbf{H}_n \toprob \mathbf{H}$,  we have
\begin{align*}
    \sqrt{n}(\theta_0-\thetahn)&=\sqrt{n}({\mathbf{H}^\intercal})^{-1} \nabla_\theta L_n(\theta_0)+o_p(1).
\end{align*}

For item ii) we follow a very similar argument. Note that by \Cref{\assumptions}, $L_n^\star(\theta)\in\mathcal{C}^3(\Theta)$. Then, for each $j\in \{1,\ldots,p\}$, we perform a first order Taylor's approximation  of $\frac{\partial}{\partial \theta_j}L^\star_n(\theta)$ around $\theta_n^\star$ and we get
\begin{align*}
\frac{\partial}{\partial \theta_j}L_n^\star(\thetahn)&=\left\langle\nabla_\theta\left(\frac{\partial}{\partial \theta_j}L_n^\star(\tilde\theta^j_\star)\right),\thetahn-\theta_n^\star\right\rangle,
\end{align*}
where $\tilde\theta^j_\star$ lies in the line segment between $\theta_n^\star$ and $\thetahn$.

By using the previous equation, we obtain $\nabla_\theta L_n^\star(\thetahn)={\mathbf{H}_n^\star}^\intercal (\thetahn-\theta_n^\star),$ where $(\mathbf{H}_n^\star)_{ij}=\frac{\partial^2}{\partial\theta_i\partial\theta_j}L_n^\star(\tilde{\theta}^j_\star)$ for each $i,j\in[p]$, and all $\tilde\theta^1_\star,\ldots,\tilde\theta^p_\star$ lie in the line segment between $\thetahn$ and $\theta_n^\star$. By \Cref{prop:convergence of Hessian} we have that $\Hb_n^\star$ converges in probability to the Hessian matrix $\Hb$. Then, by the continuity of the determinant, we have that $\Hb_n^\star$ is invertible in sets of arbitrarily large probability for all $n$ large enough, and thus $\sqrt{n}(\thetahn-\theta_n^\star)= \sqrt{n}{({\mathbf{H}_n^\star}^\intercal)}^{-1}\nabla_\theta L_n^\star(\thetahn)$. We finish by noting that under \Cref{\assumptions}, \Cref{prop:Normal} deduces that $\|\sqrt{n}\nabla_\theta L_n^\star(\thetahn)\|=O_p(1)$ and by \Cref{prop:convergence of Hessian}, ${({\mathbf{H}_n^\star}^\intercal)}^{-1}$ converges to $\Hb^{-1}$.

\end{proof}

\begin{proof}{\Cref{Lemma:convergencePB1}}
We start by proving i). \Cref{prop:normallimittheta} yields  $\sqrt{n}(\thetahn-\theta_n^\star)=\sqrt{n}\Hb^{-1} \nabla_\theta L_n^\star(\thetahn)+o_p(1)$, and \Cref{prop:Normal} yields $\|\sqrt{n}\nabla_\theta L_n^\star(\thetahn)\|=O_p(1)$, which together give us that $\sqrt{n}(\thetahn-\theta_n^\star)=O_p(1)$.

We continue by proving ii). Observe that
\begin{align*}
    \Prob(|\theta_n^\star-\theta_0|\geq\epsilon)&\leq\Prob\left(|\theta_n^\star-\thetahn|\geq\epsilon/2\right)+\Prob\left(|\thetahn-\theta_0|\geq\epsilon/2\right)\to0,
\end{align*}
where the first probability tends to 0 by part i), and the second one by \Cref{Prop:convergence} (that can only be applied under the null hypothesis).
\end{proof}

\begin{proof}{\Cref{Lemma:1}}
Define $\gamma_n:\mathcal{H}\to\R$ by
\begin{align}
 \gamma_n(\omega)=\frac{1}{\sqrt{n}}\sum_{i=1}^n\left(\omega(X_i)-\g{\thetahn}\right)\label{eqn:eqProp1}.
\end{align}

By \Cref{\assumptions}, for each fixed $\omega\in\mathcal{H}$, we have  $\g{\theta}\in\mathcal{C}^3(\Theta)$ (see \Cref{remark:lebesgue}). Then, a first order Taylor's expansion of $g(\omega,\theta)$ around $\theta_0$ yields $\g{\thetahn}=\g{\theta_0}+\langle\nabla_{\theta}\g{\tilde{\theta}_n},\thetahn-\theta_0\rangle,$  where $\tilde{\theta}_n$ lies on the line segment between $\theta_0$ and $\thetahn$. Note that by the convexity of $\Theta$, it holds that $\tilde{\theta}_n\in\Theta$. Then, by replacing the previous expression in \cref{eqn:eqProp1} we obtain
\begin{align*}
     \gamma_n(\omega)&=\frac{1}{\sqrt{n}}\sum_{i=1}^n\left(\omega(X_i)-\g{\theta_0}\right)-\sqrt{n}\left\langle\nabla_{\theta}\g{\tilde{\theta}_n},\thetahn-\theta_0\right\rangle.
\end{align*}
By assuming the null hypothesis, \Cref{Prop:convergence} yields $\thetahn\overset{\Prob}{\to}\theta_0$, and \Cref{prop:Normal}.iii and \Cref{prop:normallimittheta}.i yield $\sqrt{n}(\thetahn-\theta_0)=O_p(1)$. Then, by the continuous mapping theorem we obtain
\begin{align}
    \gamma_n(\omega)
    &=\frac{1}{\sqrt{n}}\sum_{i=1}^n\left(\omega(X_i)-\g{\theta_0}\right)-\sqrt{n}\left\langle\nabla_{\theta}\g{\theta_0},\thetahn-\theta_0\right\rangle+o_p(1).\label{eqn:replace}
\end{align}

By combining \Cref{prop:Normal} and \Cref{prop:normallimittheta}.i, we have
\begin{align*}
\sqrt{n}(\thetahn-\theta_0)&=-\sqrt{n}\Hb^{-1} \nabla_\theta L_n(\theta_0)+o_p(1)=\frac{2}{\sqrt{n}}\sum_{i=1}^n\Hb^{-1}\phit(X_i,\theta_0)+o_p(1).
\end{align*}
By replacing the previous expression in \cref{eqn:replace} we obtain
\begin{align}
    \gamma_n(\omega)
    &=\frac{1}{\sqrt{n}}\sum_{i=1}^n\left({\omega}(X_i)-\g{\theta_0}-2\left\langle\nabla_{\theta}\g{\theta_0},\Hb^{-1}\phit(X_i,\theta_0)\right\rangle\right)+o_p(1)\nonumber\\
    &=\frac{1}{\sqrt{n}}\sum_{i=1}^n\eta(\omega,X_i,\theta_0)+o_p(1)\nonumber =S_n(\omega)+o_p(1).\nonumber
\end{align}
\end{proof}

\begin{proof}{\Cref{Prop:Normal1}}
First note that %
\begin{align*}
    \mathbf{S}_n=\frac{1}{\sqrt{n}}\sum_{i=1}^n(\eta(\omega_1,X_i,\theta_0),\ldots,\eta(\omega_m,X_i,\theta_0)),
\end{align*}
where $(\eta(\omega_1,X_i,\theta_0),\ldots,\eta(\omega_m,X_i,\theta_0))_{i=1}^n$ are a collection of i.i.d. random vectors.
By \Cref{prop:eta}.i, we have that for any fixed $\omega\in\mathcal{H}$ it holds $\E_{X\sim \P_{\theta_0}}(\eta(\omega,X,\theta_0))=0$, and thus  $\E_{X\sim \P_{\theta_0}}(\mathbf{S}_n)=\mathbf{0}$.
Additionally, by \Cref{prop:eta}.ii, for any $\omega\in \mathcal{H}$ we have
\begin{equation*}
 \E_{X\sim \P_{\theta_0}}(\eta^2(\omega,X,\theta_0))<\infty.
\end{equation*}
Thus, the Central Limit Theorem yields $\mathbf{S}_n \todist N_m(0,\mathbf{\Sigma})$, where for any $i,j\in[m]$ we have
\begin{align*}
    \mathbf{\Sigma}_{ij}=\sigma(\omega_i,\omega_j)=\E_{X\sim \P_{\theta_0}}(\eta(\omega_i,X,\theta_0)\eta(\omega_j,X,\theta_0)).
\end{align*}
\end{proof}

\begin{proof}{\Cref{Parametric approx}}
Define $\gamma^\star:\mathcal{H}\to\R$ by
\begin{align}
 \gamma_n^\star(\omega)=\frac{1}{\sqrt{n}}\sum_{i=1}^n\left(\omega(X_i^\star)-\g{\theta_n^\star}\right)\label{eqn:eqProp3}.
\end{align}
 By \Cref{\assumptions}, for each $\omega\in\mathcal{H}$, $g(\omega,\theta_n^\star)\in\mathcal{C}^3(\Theta)$ (see \Cref{remark:lebesgue}). By a first order Taylor's expansion of $\g{\theta}$ around $\thetahn$ we obtain $g(\omega,\theta^\star_n)=g(\omega,\thetahn)+\langle\nabla_\theta g(\omega,\tilde\theta_n^\star),\theta^\star_n-\thetahn\rangle$ where the $\tilde\theta_n^\star$ belongs to the line segment between $\thetahn$ and $\theta_n^\star$. Then, it follows that
\begin{align*}
    \gamma_n^\star(\omega)=\frac{1}{\sqrt{n}}\sum_{i=1}^n\left(\omega(X_i^\star)-g(\omega,\thetahn)-\left\langle\nabla_\theta g(\omega,\tilde\theta_n^\star),\theta^\star_n-\thetahn\right\rangle\right).
\end{align*}

 \Cref{Prop:convergence} yields that $\thetahn-\theta_0=o_p(1)$, and \Cref{Lemma:convergencePB1} yields $\theta^\star_n-\theta_0=o_{p}(1)$  and $\sqrt{n}(\theta^\star_n-\thetahn)=O_{p}(1)$. Thus, by the continuous mapping theorem it holds
\begin{align}
    \gamma_n^\star(\omega)&=\frac{1}{\sqrt{n}}\sum_{i=1}^n\left(\omega(X_i^\star)-g(\omega,\thetahn)-\left\langle\nabla_\theta g(\omega,\theta_0),\theta^\star_n-\thetahn\right\rangle\right)+o_{p}(1).\label{eqn:38}
\end{align}
Finally, by combining \Cref{prop:Normal,prop:normallimittheta}.ii we get
\begin{align*}
\sqrt{n}(\theta_n^\star-\thetahn)=-\sqrt{n}\Hb^{-1} \nabla_\theta L_n^\star(\thetahn)+o_p(1)=\frac{2}{\sqrt{n}}\sum_{i=1}^n\Hb^{-1}\phit(X_i^\star,\thetahn)+o_p(1),
\end{align*}
and thus by replacing in \cref{eqn:38},
\begin{align*}
 \gamma_n^\star(\omega)%
 &=\frac{1}{\sqrt{n}}\sum_{i=1}^n\left(\omega(X_i^\star)-g(\omega,\thetahn)-2\left\langle\nabla_\theta \g{\theta_0},\Hb^{-1}\phit(X_i^\star,\thetahn)\right\rangle\right)+o_{p}(1)\\
  &=\frac{1}{\sqrt{n}}\sum_{i=1}^n\eta(\omega,X_i^\star,\thetahn)+o_{p}(1).
\end{align*}
\end{proof}

\begin{proof}{\cref{Prop:Normal2}}
We will apply the Lidenberg-Feller Central Limit Theorem for multivariate triangular arrays. Define $Z_n=\frac{1}{n}\sum_{i=1}^n \bY_{n,i}$, where $\{\bY_{n,i}\}$ is a triangular array of random vectors $\bY_{n,i}=(\eta(\omega_1,X_{n,i}^\star,\thetahn),\ldots,\eta(\omega_m,X_{n,i}^\star,\thetahn))$, $\omega_1,\ldots,\omega_m\in\mathcal{H}$ and $m\in\mathbb{N}$. Observe that by \Cref{prop:eta}, we have that
\begin{align*}
  \E_D(\eta(\omega,X_{n,i}^\star,\thetahn))=\int_\X\eta(\omega,z,\thetahn))p_{\thetahn}(z)\lambda(dz)=0
\end{align*}
holds for any $\omega\in\mathcal{H}$, and thus $\E_D(\bY_{n,i})=0$. Consider  $\mathbf{V}_n=\frac{1}{n}{\sum_{i=1}^n}\Var_D(\bY_{n,i})$, and observe that for any $\ell,k\in[m]$, we have that
\begin{align}
(\mathbf{V}_n)_{\ell,k}=\Cov_D(\eta(\omega_\ell,X_{n,1}^\star,\thetahn),\eta(\omega_k,X_{n,1}^\star,\thetahn))&=\int_\X\eta(\omega_\ell,z,\thetahn)\eta(\omega_k,z,\thetahn))p_{{\thetahn}}(z)\lambda(dz)\nonumber\\
&\overset{\Prob}{\to} \int_\X\eta(\omega_\ell,z,\theta_0)\eta(\omega_k,z,\theta_0))p_{{\theta_0}}(z)\lambda(dz).\label{eqn:resultvar}
\end{align}
The convergence in probability holds since $\thetahn\overset{\Prob}{\to}\theta_0$ by \Cref{Prop:convergence}, and since the map $  \theta \to \int_\X \eta(\omega,z,\theta) \eta(\omega',z,\theta) p_\theta(z) \lambda(dz)
$ is continuous for any fixed $\omega,\omega'\in\mathcal{H}$.  Thus the continuous mapping theorem yields \cref{eqn:resultvar}.

 We proceed to check Linderberg's condition. Observe that for every $\varepsilon>0$
\begin{align*}
    \frac{1}{n}\sum_{i=1}^n\E_D\left(\|\bY_{n,i}\|^2\ind_{\{\|\bY_{n,i}\|\geq \varepsilon \sqrt{n}\}}\right)&=\frac{1}{n}\sum_{i=1}^n\sum_{j=1}^m\E_D\left(\eta^2(\omega_j,X_{n,i},\thetahn)\ind_{\{\|\bY_{n,i}\|\geq \varepsilon \sqrt{n}\}}\right)\\
    &\leq C_1\frac{1}{n}\sum_{i=1}^n\Prob_D\left(\|\bY_{n,i}\|\geq \varepsilon \sqrt{n}\right)\\
    &\leq C_1\frac{1}{n}\sum_{i=1}^n\frac{\E_D(\|\bY_{n,i}\|)}{\epsilon\sqrt{n}} \to 0,\quad a.s.
\end{align*}
The first inequality holds since $\sup_{\theta\in\Theta}\sup_{x\in\mathcal{X}}\eta^2(\omega_j,x,\theta)<\infty$ by \Cref{prop:eta}.iii.

\end{proof}

\begin{proof}{\Cref{prop:secondmomentbound}}
By \Cref{remark:lebesgue}
\begin{align}
\sum_{i=1}^\infty\|\nabla_{\theta}g(\psi_i,\theta_0)\|^2&=\sum_{i=1}^\infty\sum_{\ell=1}^p\left(\int\psi_i(y)\frac{\partial}{\partial\theta_\ell}p_{\theta_0}(y)\lambda(dy)\right)^2\label{eqn:comb2}.
\end{align}
Observe that for each $\ell\in \{1,\ldots,p\}$, it holds that
\begin{align}
&\sum_{i=1}^\infty\left(\int_\X\psi_i(y)\frac{\partial}{\partial\theta_\ell}p_{\theta_0}(y)\lambda(dy)\right)^2\nonumber\\
&\qquad=\sum_{i=1}^\infty\int_\X\int_\X\psi_i(y)\psi_i(x)\left(\frac{\partial}{\partial\theta_\ell}p_{\theta_0}(y)\right)\left(\frac{\partial}{\partial\theta_\ell}p_{\theta_0}(x)\right)\lambda(dy)\lambda(dx)\nonumber\\
 &\qquad\leq\int_\X\int_\X\sum_{i=1}^\infty|\psi_i(y)||\psi_i(x)|\left|\frac{\partial}{\partial\theta_\ell}p_{\theta_0}(y)\right|\left|\frac{\partial}{\partial\theta_\ell}p_{\theta_0}(x)\right|\lambda(dy)\lambda(dx)\nonumber\\
&\qquad\leq\int_\X\int_\X\sqrt{\sum_{i=1}^\infty|\psi_i(y)|^2}\sqrt{\sum_{j=1}^\infty|\psi_j(x)|^2}\left|\frac{\partial}{\partial\theta_\ell}p_{\theta_0}(y)\right|\left|\frac{\partial}{\partial\theta_\ell}p_{\theta_0}(x)\right|\lambda(dy)\lambda(dx)\nonumber\\
 &\qquad=\left(\int_\X\sqrt{\sum_{i=1}^\infty|\psi_i(y)|^2}\left|\frac{\partial}{\partial\theta_\ell}p_{\theta_0}(y)\right|\lambda(dy)\right)^2\nonumber\\
 &\qquad=\left(\int_\X\sqrt{K(y,y)}\left|\frac{\partial}{\partial\theta_\ell}p_{\theta_0}(y)\right|\lambda(dy)\right)^2 \leq\left(\sqrt{C}\int_\X\sup_{\theta\in\Theta}\left|\frac{\partial}{\partial\theta_\ell}p_{\theta}(y)\right|\lambda(dy)\right)^2<\infty\label{eqn:comb1}
\end{align}
where the second inequality is due to the Cauchy-Schwarz inequality, the second equality holds since $\sum_{i=1}^\infty\psi_i^2(y)=K(y,y)$, the third inequality follows from the fact that the kernel is bounded (\Cref{condition:mmd_kernel}), and the last inequality is due to  \Cref{condition:model}. By combining \cref{eqn:comb1,eqn:comb2} we conclude that $\sum_{i=1}^\infty\|\nabla_{\theta}g(\psi_i,\theta_0)\|^2<\infty$.

\end{proof}

\section{Illustration of Limitations}
\label{app:limitations}

\begin{figure}
  \centering
  \includegraphics{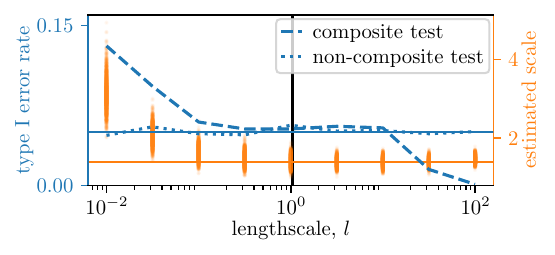}
  \caption{
    Comparison of the type I error rate of a composite and non-composite test, as we vary the lengthscale of the kernel.
    The left axis, blue, shows the type I error rate of the two tests.
    The right axis, orange, shows the estimates of the scale parameter of the model made by the composite test over $2000$ repeats.
    The solid blue horizontal line shows the level of the test, $0.05$.
    The solid orange horizontal line shows the true value of the standard deviation, $1.4$.
    The solid black vertical line shows the lengthscale selected by the median heuristic.
  }
  \label{fig:bad_lengthscale}
\end{figure}
We illustrate two of the limitations introduced in \Cref{sec:limitations}.
The first occurs when $\D(\P,\Q_n)$ is poor at distinguishing between probability distributions.
For example, this could occur when using a Gaussian kernel with a very small or large lengthscale.
\Cref{fig:bad_lengthscale} shows the type I error rate of the composite KSD test, using the parametric bootstrap, as the lengthscale of the Gaussian kernel varies.
We consider the null hypothesis case where $\nullc = $ the data is Gaussian and the data is sampled from $\Q = \Normal(0.4, 1.4^2)$.
The figure shows that the type I error rate of the composite test can exceed the level for very small lengthscales, and goes to zero for large lengthscales.
The figure also shows the type I error rate of a non-composite test with $\nullnc : \P = \Q$ as defined above, which reveals that the non-composite test does not suffer from this problem.
To investigate the issue the figure also shows the estimates of the scale parameter.
For very small lengthscales, the parameter estimates are not centered around the true parameter value, thus the test is comparing the data to a poor choice of model from $\{\P_\theta\}_\theta$, leading to type I errors.
For large lengthscales, we suggest that the type I error rate of the composite test goes to zero faster than that of the non-composite test because the small amount of variance in the estimated parameters makes distinguishing the null and alternative hypotheses more difficult.
Importantly, we note that the range of lengthscales for which the composite test has good performance is fairly wide (note the log x-axis scale), and the median heuristic selects a value in the middle of this range.
The same behaviour can also occur for the MMD test, although, in this case, we have the additional reassurance of theoretical results which show that the test will behave correctly as $n$ becomes large.

\begin{figure}
  \centering
  \includegraphics{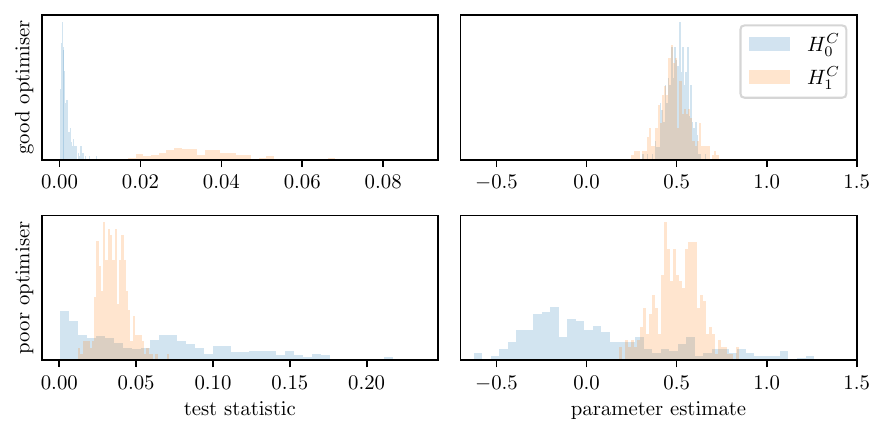}
  \caption{
    Demonstration of a failure mode where the parameter estimate has high variance due to a poorly configured optimiser.
    The first row shows the distribution of the test statistic (left) and that of the parameter estimate (right) for a well-configured optimiser, and the second row the distributions for a poorly-configured optimiser.
  }
  \label{fig:overlap_failure}
\end{figure}
The second limitation we will illustrate occurs when there is large variance in the estimate of the parameter, if either the minimum distance estimator has high variance or the optimisation process fails.
\Cref{fig:overlap_failure} demonstrates this limitation for the MMD test, testing $\nullc$ that $\Q = \Normal(\theta, 1^2)$ for $\theta \in \Reals$ against the alternative $\Normal(0.5, 1.5^2)$.
If the optimiser is configured correctly, the figure shows that the estimates of the parameter are centered around the true parameter value and have low variance, and there is clear separation between the distributions of the test statistic under $\nullc$ and $\altc$.
Thus, the test is able to distinguish $\nullc$ and $\altc$ and achieve a high power.
If the optimiser is configured poorly---we use too large a learning rate---we can see that the parameter estimates have higher variance, and the distributions of the test statistic under $\nullc$ and $\altc$ overlap completely.
Thus, no matter where the threshold is set it will not be possible to distinguish the two hypotheses, and the test will have very low power.

\section{Aggregated Composite Test}
\label{app:aggregated_test}

\begin{figure}
  \centering
  \includegraphics{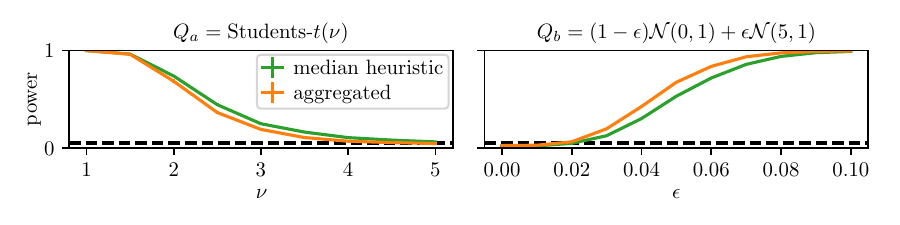}
  \caption[Power of an aggregated composite test]{
    Power of the KSD test with wild bootstrap, comparing setting the lengthscale set using the median heuristic and aggregating tests using $8$ different lengthscales.
    We set $\nullc : \P = \Normal(\mu, \sigma^2)$, with the left and right figure showing data generated from different distributions, $\Q_a$ and $\Q_b$.
    \tikz[baseline=-0.5ex]{\draw [line width=0.5mm, dashed, black] (0,0) -- (2ex,0);} shows the level.
  }
  \label{fig:aggregated_test}
\end{figure}
\Cref{fig:aggregated_test} shows an attempt at combining the aggregated test method introduced by \citet{schrabMMDAggregatedTwoSample,schrabKSDAggregatedGoodnessoffit} with our composite KSD test, using the wild bootstrap.
The aggregated test is the combination of $8$ composite tests using the Gaussian kernel, where the lengthscales are set as the $\{25, 32, 39, 46, 54, 61, 68, 75\}$th percentiles of the pairwise distances between the observations.
The non-aggregated test uses a single lengthscale set using the median heuristic, i.e. at the $50$th percentile of the pairwise distances.
The null hypothesis of the test is $\nullc : \P = \Normal(\mu, \sigma^2)$, for $\mu \in \Reals$ and $\sigma \in \Reals^+$.
To evaluate the performance of the test, we consider the two alternatives specified in the figure.
Against a Student's $t$-distribution, we find that the median heuristic has slightly better performance than the aggregated test, while, against a mixture of Gaussians, the aggregated test has a small advantage.
We would expect the aggregated test to have a bigger advantage in the case of the mixture alternative, as here the spread of lengthscales will bring a bigger benefit than in the unimodal case of the Student's $t$ alternative.
It is somewhat unexpected that the aggregated test has slightly lower performance than the median heuristic against the Student's $t$ alternative.
Future work could investigate why this happens, and whether any modifications can be made to the aggregated testing procedure to improve the performance in the composite case.

\section{Experiment Details}
\label{appendix:experiment_details}

For all experiments we use a test level of $\alpha = 0.05$.
The experiments are implemented in JAX, and executed on an Nvidia RTX 3090 GPU.

\subsection{Closed-Form KSD Estimator} \label{app:closed-form-ksd-estimator}
For any $\P_\theta$ in the exponential family, we can calculate $\thetahn = \argmin_\theta \KSD^2(\P_\theta, \Q_n)$ in closed-form.
Here we state the expression for this estimator, which we use in our Gaussian and kernel exponential family experiments.
The detailed derivation of this result can be found in  \citet[Appendix~D3]{barpMinimumSteinDiscrepancy} or \citet{Matsubara2021}.

Let the density of a model in the exponential family be $p_\theta(x) = \exp( \eta(\theta) \cdot t(x) - a(\theta) + b(x) )$, with $\eta: \Theta \to \Reals^k$ an invertible map, $t: \Reals^d \to \Reals^k$ any sufficient statistic for some $k \in \mathbb{N}_1$, $a: \Theta \to \Reals$ and $b: \Reals^d \to \Reals$.
Then the \gls{ksd} estimator is given by $\hat{\theta}_n
  := \eta^{-1} \left( -\frac{1}{2} \Lambda_n^{-1} v_n \right)$, where $\Lambda_{n} \in \Reals^{k \times k}$ and $\nu_{n} \in \Reals^{k}$ are defined as
$\Lambda_{n} := \frac{1}{n^2} \sum_{i=1}^{n} \sum_{j=1}^{n} \Lambda(x_i, x_j)$ and
$\nu_n := \frac{1}{n^2} \sum_{i=1}^{n} \sum_{j=1}^{n} \nu(x_i, x_j)$.
These are based on functions $\Lambda: \Reals^d \times \Reals^d \to \Reals^{k \times k}$ and
$\nu: \Reals^d \times \Reals^d \to \Reals^k$
which depend on the specific model, defined as $\Lambda(x, x') : = K(x, x') \nabla_{x}t(x)  \nabla_{x}t(x')^\top$ and
\begin{align*}
  \nu(x, x')     : ={} & K(x, x') \nabla_{x}b(x) \nabla_{x}t(x')^\top
  + \nabla_{x}t(x) \nabla_{x'} K(x, x')                                 \\
                       & + K(x, x') \nabla_{x}b(x') \nabla_{x}t(x)^\top
  + \nabla_{x}t(x') \nabla_{x} K(x', x) \eqstop
\end{align*}

\subsection{Kernels}
\label{app:kernels}
We define the Gaussian kernel as $K(x_1, x_2) = \exp( - \norm{x_1 - x_2}_2^2/2 l^2 )$, and the IMQ kernel as $  K(x_1, x_2) = (1 + \norm{x_1 - x_2}_2^2/2 l^2)^{-\half}$, where $l \in \Reals^+$ is the lengthscale, and the exponent of $-\half$ follows \citet{Matsubara2021}.
Where we set the lengthscale using the median heuristic, this is defined as $\lmed = \sqrt{\operatorname{median}(\norm{y_i - y_j}_2^2 / 2)}$, where the median is taken over all pairs of observations, $y_i$ and $y_j$.

Where we use a sum kernel, this is defined as $K(x_1, x_2) = \frac{1}{L} \sum_{i=1}^{L} K_{l_i}(x_1,x_2)$, where $K_{l_1},\ldots,K_{l_L}$ are a set of $L$ Gaussian or IMQ kernels with lengthscales $l_1,\ldots,l_L$.
Prior work has found that using a sum of kernels with a range of lengthscales, rather than attempting to choose the lengthscale of a single kernel,  produces good empirical results \citep{liGenerativeMomentMatching,renConditionalGenerativeMomentMatching,sutherlandGenerativeModelsModel}.
Note that a sum of characteristic kernels is a characteristic kernel \citep{Sriperumbudur2009}, so this choice satisfies the assumptions we make in our theoretical results.

\subsection{Details of Specific Figures}
\subsubsection[Distribution of Wild Bootstrap]{
  Distribution of Wild Bootstrap Statistic (\cref{fig:wild_bootstrap_distribution})
}
$\nullc : \P = \Normal(\mu, \sigma^2)$ for $\mu \in \Reals$ and $\sigma \in \Reals^+$, with the estimator configured the same as in the Gaussian experiments below.
We sample observations from $\Q = \Normal(0.3, 1.0)$.
We use a Gaussian kernel with lengthscale $1.0$.
To compute the distribution of the test statistic, we take $1000$ sets of $n=1000$ samples from $\Q$, and for each estimate $\P_{\thetahn}$ and compute $\MMD^2(\P_{\thetahn,n}, \Q_n)$.
To compute the distribution of the wild bootstrap statistic, we take $n=1000$ samples from $\Q$, estimate $\P_{\thetahn}$, sample $1000$ sets of Rademacher variables, and thus compute $1000$ samples of $\MMD^2_W(\P_{\thetahn,n}, \Q_n)$.

\subsubsection[Gaussian Model]{Gaussian Model
  (\Cref{fig:gpc_increasing_n,fig:gpc_increasing_d,fig:gpc_ours_vs_variance_aware,fig:robustness,fig:gpc_other_tests})
}
\begin{center}
  \begin{tabular}{r l}
    \toprule
    repeats to compute power (per seed) & $1000$              \\
    $K$                                 & Gaussian, $l=\lmed$ \\
    wild bootstrap samples              & $500$               \\
    parametric bootstrap samples        & $300$               \\
    \bottomrule
  \end{tabular}
\end{center}

\paragraph{MMD test details}
For the experiments using the MMD, we minimize the minimum distance estimator using the Adam optimiser \citep{kingma2015adam}.
We use a learning rate of $0.05$ and $100$ iterations.
When $\theta = \mu_d$ (\cref{fig:gpc_increasing_d,fig:gpc_ours_vs_variance_aware,fig:robustness}) we sample the initial value of $\mu_d$ from $\Normal(\bm{0}_d, I_d)$.
When $\theta = \{\mu,\sigma^2\}$ (\cref{fig:gpc_increasing_n,fig:gpc_other_tests}) sample the initial $\mu$ from $\Normal(0, 1)$, and sample $\sigma$ from a standard normal distribution truncated between $0.5$ and $1.5$.

\paragraph{Timing}
We estimate the time taken by each bootstrap as follows.
We run $3$ tests as a warmup, then we run $2000$ tests and calculate the mean time taken.
The full configuration is as follows:

\begin{center}
  \begin{tabular}{r l @{\hskip 1cm} r l}
    \toprule
    $\nullc$ & $\P_{\mu,\sigma} = \Normal(\mu, \sigma^2)$
    for $\mu \in \Reals$, $\sigma \in \Reals^+$
             &
    $n$      & $100$                                                                  \\
    $\D$     & KSD                                        & bootstrap samples & $300$ \\
    $K$      & Gaussian, $l=1.0$                                                      \\
    \bottomrule
  \end{tabular}
\end{center}

\subsubsection{Toggle Switch}
\label{app:toggle_switch_details}
\paragraph{\Cref{fig:toggle_switch_time_evo,fig:toggle_switch_increasing_T,fig:toggle_switch_fits}}
The true model parameters, $\theta_0$, are
\begin{center}
  \begin{tabular}{r r r r r r r r}
    \toprule
    parameter  & $\alpha_1$ & $\alpha_2$ & $\beta_1$ & $\beta_2$
               & $\mu$      & $\sigma$   & $\gamma$              \\
    \midrule
    true value & $22.0$     & $12.0$     & $4.0$     & $4.5$
               & $325.0$    & $0.25$     & $0.15$                \\
    \bottomrule
  \end{tabular}
\end{center}
We use stochastic gradient descent with random restarts to estimate the parameters.
The algorithm is:
\begin{enumerate}
  \item Sample $500$ initial parameter values, $\theta_{\text{init}}^1,\ldots,\theta_{\text{init}}^{500}$.
  \item Select the $15$ $\theta_{\text{init}}^i$ for which $\MMD^2(\P_{\theta_{\text{init}}^i},\Q_n)$ is smallest.
  \item Run SGD $15$ times, once for each of the selected $\theta_{\text{init}}^i$, to find $\thetahn^1,\ldots,\thetahn^{15}$.
        Use learning rate $0.04$ and perform $300$ iterations.
  \item Return $\thetahn^i$ for which $\MMD^2(\P_{\thetahn^i},\Q_n)$ is smallest.
\end{enumerate}
The initial parameters are sampled from a uniform distribution with the following ranges:
\begin{center}
  \begin{tabular}{r r r r r r r r}
    \toprule
    parameter                      & $\alpha_1$ & $\alpha_2$ & $\beta_1$ & $\beta_2$
                                   & $\mu$      & $\sigma$   & $\gamma$              \\
    \midrule
    \multirow{2}{*}{initial range} & $0.01$     & $0.01$     & $0.01$    & $0.01$
                                   & $250.0$    & $0.01$     & $0.01$                \\
                                   & $50.00$    & $50.00$    & $5.00$    & $5.00$
                                   & $450.0$    & $0.50$     & $0.40$                \\
    \bottomrule
  \end{tabular}
\end{center}
To run the test, we use the following configuration:
\begin{center}
  \begin{tabular}{r p{9cm}}
    \toprule
    $K$               & unweighted mixture of Gaussian kernels
    with lengthscales $20.0$, $40.0$, $80.0$, $100.0$, $130.0$, $200.0$,
    $400.0$, $800.0$, $1000.0$                                 \\
    $n$               & $400$                                  \\
    bootstrap         & parametric                             \\
    bootstrap samples & $200$                                  \\
    \bottomrule
  \end{tabular}
\end{center}
To create \Cref{fig:toggle_switch_fits} we apply kernel density estimation to samples from the model.
We use a Gaussian kernel, with the bandwidth set to $0.1$.

\subsubsection{Nonparametric Density Estimation}
\label{app:kef_details}
\paragraph{\Cref{fig:kvf_galaxies}}
We are testing whether the family of models considered by \citet{Matsubara2021} fits the data, thus our choices of $\qkef$ and $\lkef$ match the original paper.
\begin{center}
  \begin{tabular}{r p{8cm}}
    \toprule
    $K$                    &
    unweighted sum of IMQ kernels with lengthscales $0.6$, $1.0$, $1.2$ \\
    $n$                    & $82$                                       \\
    wild bootstrap samples & $500$                                      \\
    $\qkef$                & $\Normal(0.0, 3.0^2)$                      \\
    $\lkef$                & $\sqrt{2}$                                 \\
    \bottomrule
  \end{tabular}
\end{center}
Following \citet{Matsubara2021}, we normalise each observation in the data set as follows: $
  y_i^{\text{norm}} = \frac{y_i - \mu}{0.5 * \sigma}$,
where $\mu$ and $\sigma$ are the mean and standard deviation of the data set.

To plot the density of the model, we estimate the normalising constant of the model using the importance sampling approach suggested by \citet[Section~3.2]{wenliangLearningDeepKernels}.
As a proposal we use $\Normal(0.0, 3.5^2)$, and we take $2000$ samples.

\bibliography{references_converted}

\end{document}